\documentclass{article}

\PassOptionsToPackage{numbers, compress, sort}{natbib}

 \usepackage[main, final]{neurips_2026}

\usepackage[utf8]{inputenc} %
\usepackage[T1]{fontenc}    %
\usepackage{hyperref}       %
\usepackage{url}            %
\usepackage{booktabs}       %
\usepackage{amsfonts}       %
\usepackage{nicefrac}       %
\usepackage{microtype}      %
\usepackage{xcolor}         %

\usepackage[utf8]{inputenc} %
\usepackage[T1]{fontenc}    %
\usepackage{microtype}
\usepackage{graphicx}
\usepackage{amsmath}
\usepackage{amssymb}
\usepackage{amsthm}
\usepackage{enumitem}
\usepackage{xfrac}
\usepackage{numprint}
\usepackage{booktabs} %
\usepackage[textsize=scriptsize,textwidth=1.9cm]{todonotes}
\usepackage{xspace}
\usepackage{subcaption}
\usepackage{etoolbox}
\usepackage[fixed]{fontawesome5}
\usepackage{etoc}
\usepackage{wrapfig}
\usepackage{placeins}
\usepackage{makecell}
\usepackage{etoc}
\usepackage{amsfonts}       %
\usepackage{nicefrac}       %
\usepackage{mathtools}
\usepackage{multirow}
\usepackage[table,svgnames]{xcolor}
\usepackage{babel}
\usepackage{todonotes}
\usepackage{caption,tikz}
\usepackage{tcolorbox}
\usepackage{titletoc} %
\usepackage{longtable} %

\usepackage{minitoc}
\noptcrule

\newcommand{\customfootnotetext}[2]{{%
  \renewcommand{\thefootnote}{#1}%
  \footnotetext[0]{#2}}}%

\usepackage{hyperref}
\hypersetup{
		colorlinks=true,
		linkcolor=blue!70!black,
		citecolor=blue!70!black,
		urlcolor=blue!70!black}
\usepackage[hyperpageref]{backref}
\renewcommand*{\backref}[1]{}
\renewcommand*{\backrefalt}[4]{%
  \ifcase #1 %
    \\\textit{\footnotesize Not cited.}%
  \or
    \\\textit{\footnotesize Cited on page #2.}%
  \else
    \\\textit{\footnotesize Cited on pages #2.}%
  \fi
}
\usepackage{adjustbox}
\usepackage{color, colortbl}
\definecolor{Gray}{gray}{0.9}
\definecolor{textcolor}{HTML}{DCDCDC}

\usepackage{nicematrix}
\usepackage{siunitx}

\hypersetup{
  colorlinks   = true, %
  urlcolor     = RoyalBlue, %
  linkcolor    = RoyalBlue, %
  citecolor   = RoyalBlue %
}

\usepackage{rotating}
\usepackage{placeins}
\usepackage{enumitem}
\usepackage{wrapfig}
\usepackage{xspace}
\usepackage{fvextra}
\usepackage{tablefootnote}
\usepackage{arydshln}
\usepackage{tabularx}
\usepackage{multicol}
\usepackage{wrapfig}
\usepackage{xcolor}
\usepackage{colortbl}
\usepackage{adjustbox}
\usepackage{subcaption}

\usepackage{fontawesome5}

\newcommand{\huggingface}{\raisebox{-1.5pt}{\includegraphics[height=1.05em]{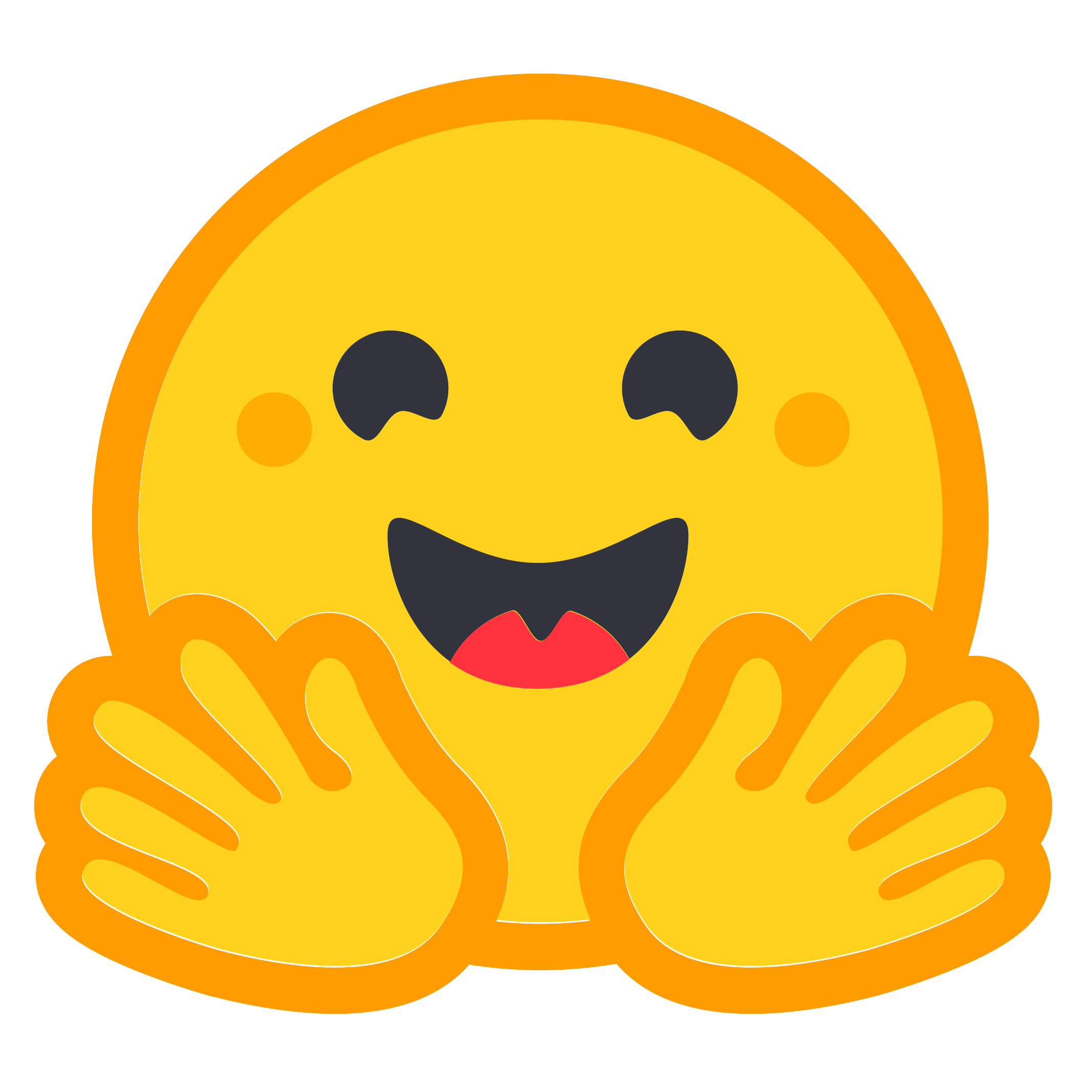}}\xspace}

\definecolor{nicergreen}{RGB}{34,139,34}
\definecolor{light-light-gray}{gray}{0.92}

\definecolor{light}{rgb}{0.68, 0.90, 0.77}
\definecolor{orange}{rgb}{0.93, 0.74, 0.60}
\definecolor{lightorange}{rgb}{1, 0.87, 0.68}
\definecolor{lightgreen}{rgb}{0.76, 0.88, 0.76}
\definecolor{lightgray}{rgb}{0.92, 0.92, 0.92}
\definecolor{lightred}{rgb}{0.92, 0.29, 0.36}
\definecolor{lightcyan}{rgb}{0.424, 0.651, 0.804}

\newcommand{\smallsec}[1]{\vspace{1pt}\noindent\textbf{#1}}

\usepackage{pifont}%
\newcommand{\cmark}{\ding{51}}
\newcommand{\xmark}{\ding{55}}

\definecolor{lightgray}{rgb}{0.95, 0.95, 0.97}

\usepackage{cleveref}
\crefformat{footnote}{#2\footnotemark[#1]#3}
\Crefname{table}{Table}{Tables}
\crefname{table}{Tab.}{Tabs.}
\Crefname{figure}{Figure}{Figure}
\crefname{figure}{Fig.}{Figs.}
\Crefname{appendix}{Appendix}{Appendix}
\crefname{appendix}{Appx.}{Apps.}
\Crefname{algorithm}{Algorithm}{Algorithm}
\crefname{algorithm}{Alg.}{Algs.}
\Crefname{section}{Section}{Section}
\crefname{section}{Sec.}{Secs.}

\newcommand{\dcvlm}{\textsc{DCVLM}\xspace}
\newcommand{\fv}{\textsc{FineVision}\xspace}

\newcommand{\torun}[1]{\textcolor{red}{[Experiment ongoing]}}
\newcommand{\minortorun}[1]{\textcolor{red}{[Minor reruns required]}}

\definecolor{mm-green}{HTML}{8ce5a1}

\newcommand{\fighero}{%
\begin{figure*}[!t]
\centering
\includegraphics[width=0.95\linewidth]{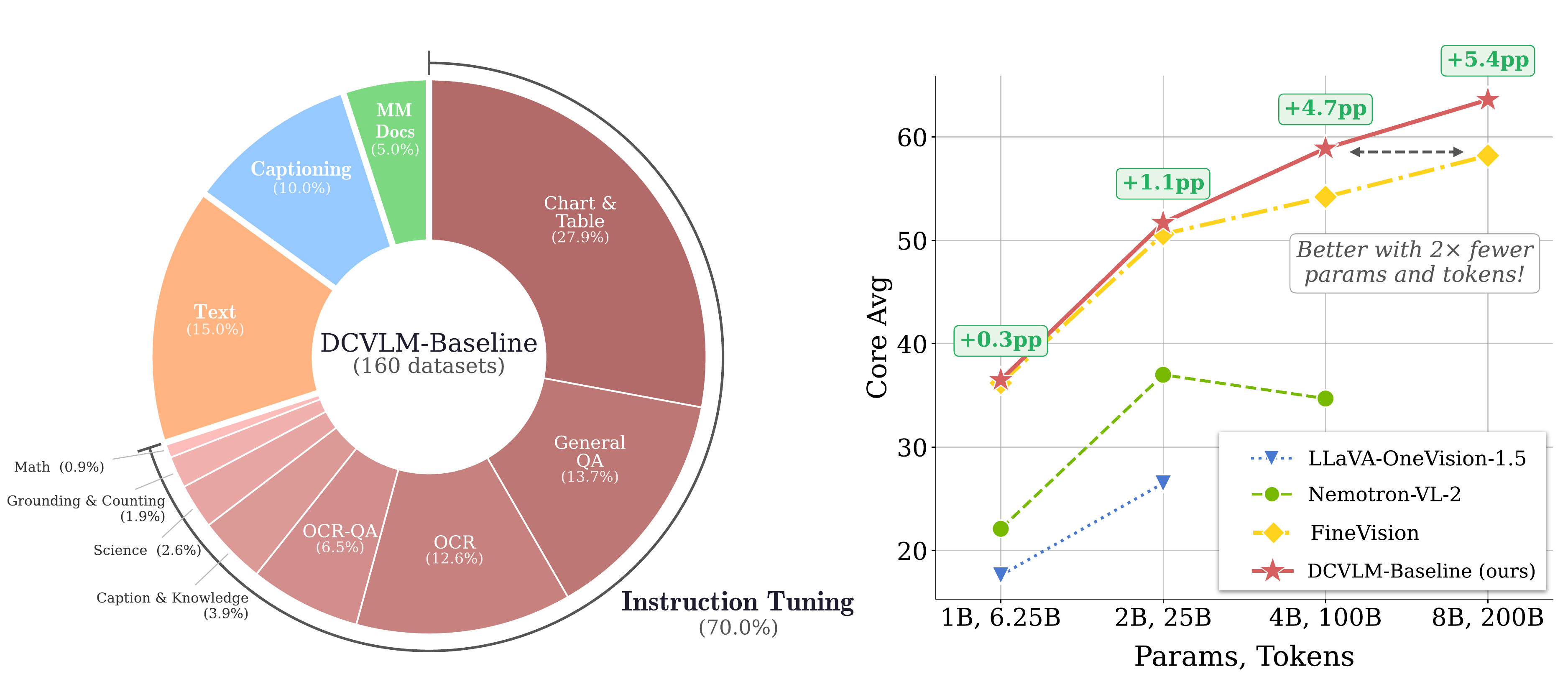}
\caption{\textbf{\textsc{\dcvlm-Baseline} outperforms open VLM training datasets.} \textsc{\dcvlm-Baseline} \textit{(left)} combines 160 sources as $10$\% image-caption pairs, $5$\% multimodal documents, $15$\% text-only, and $70$\% multimodal instruction-tuning data. On our 33-evaluation Core set \textit{(right)}, it outperforms existing datasets~\cite{deshmukh2025nvidia, an2025llava, wiedmann2025finevision} across all scales. Notably, a 4B model trained on \textsc{\dcvlm-Baseline} for 100B tokens beats an 8B model trained on \textsc{FineVision} for 200B tokens, a ${4}{\times}$ compute reduction.}
\label{fig:hero}
\vspace{-1.5em}
\end{figure*}
}

\newcommand{\figpoolcomposition}{%
\begin{figure}[!h]
\centering
\includegraphics[width=\linewidth]{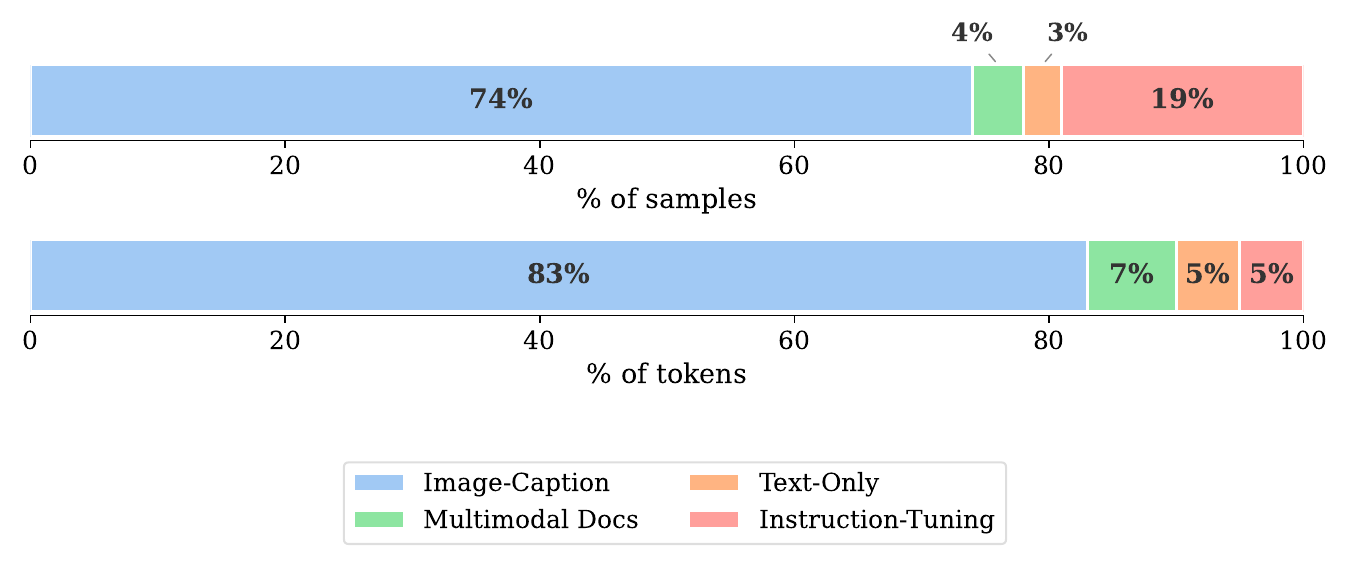}
\caption{\textbf{DCVLM pool composition by data type.} Share of total samples (\textit{top}) vs.\ share of total multimodal tokens (\textit{bottom}) for each of the four data types. 
The pool is dominated by image-caption pairs on both axes  (83\% tokens vs\ 74\% samples). Text-only data exhibits the opposite asymmetry, with 19\% of samples but only 5\% of tokens. 
Instruction-tuning data and Multimodal documents are token-dense, i.e, their overall token proportion is much larger than their overall sample proportion, thanks to the presence of potentially many multi-image examples contributing to visual tokens.}
\label{fig:pool-composition}
\end{figure}
}

\newcommand{\figpoolscatter}{%
\begin{figure}[!t]
\centering
\includegraphics[width=0.78\linewidth]{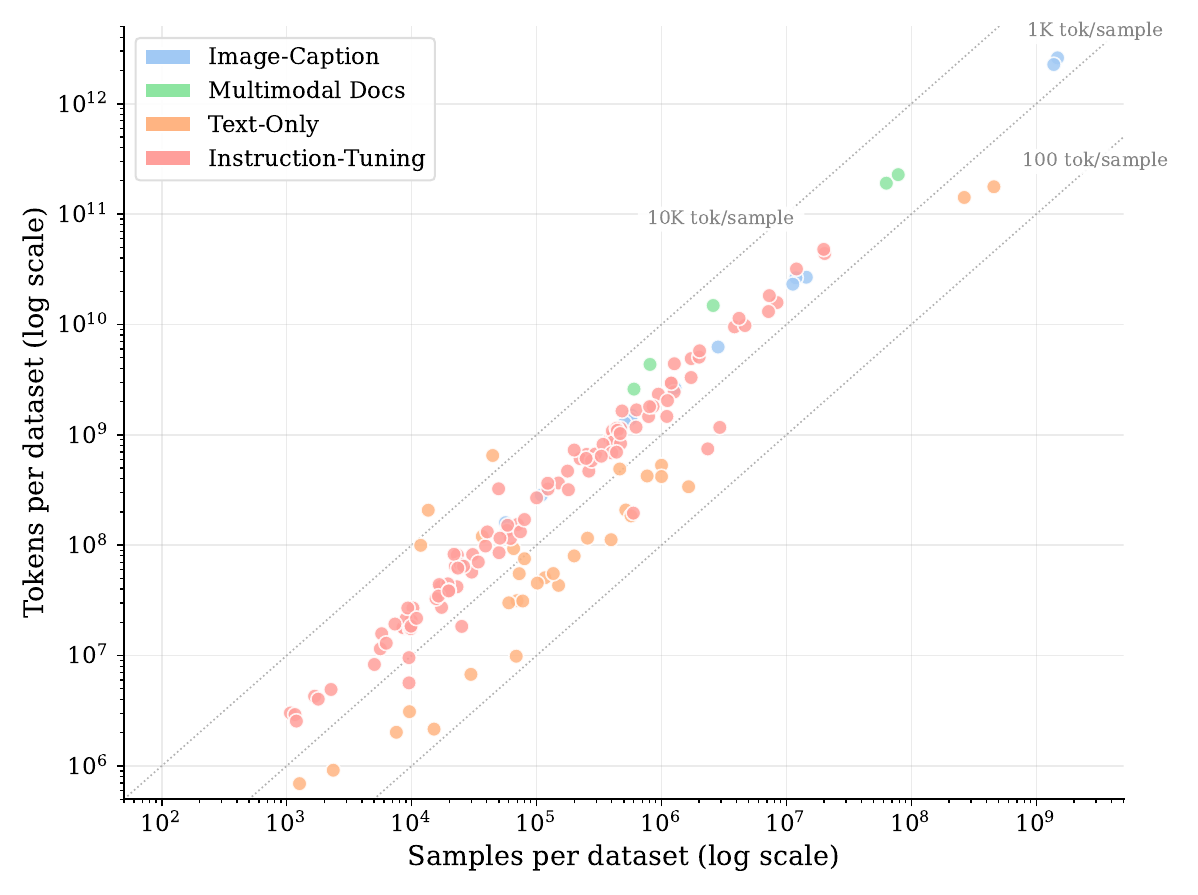}
\caption{\textbf{Samples vs.\ multimodal tokens (log--log).}
Each marker is one of 160 datasets in our pool, colored by data type.
Diagonal reference lines mark constant tokens-per-sample regimes (100, 1K, 10K).
Image-caption datasets cluster tightly along the 1--2K tok/sample diagonal driven by the visual-token contribution per image; multimodal documents sit one decade higher (multi-image samples); text-only datasets occupy a much wider band (100--15K tok/sample) reflecting the diversity of short instruction data and long-context corpora; instruction-tuning datasets span the largest dynamic range in size (10$^3$--10$^7$ samples) but a relatively narrow tokens-per-sample band.}
\label{fig:pool-scatter}
\end{figure}
}

\newcommand{\figlangdistribution}{%
\begin{figure}[!t]
\centering
\includegraphics[width=0.95\linewidth]{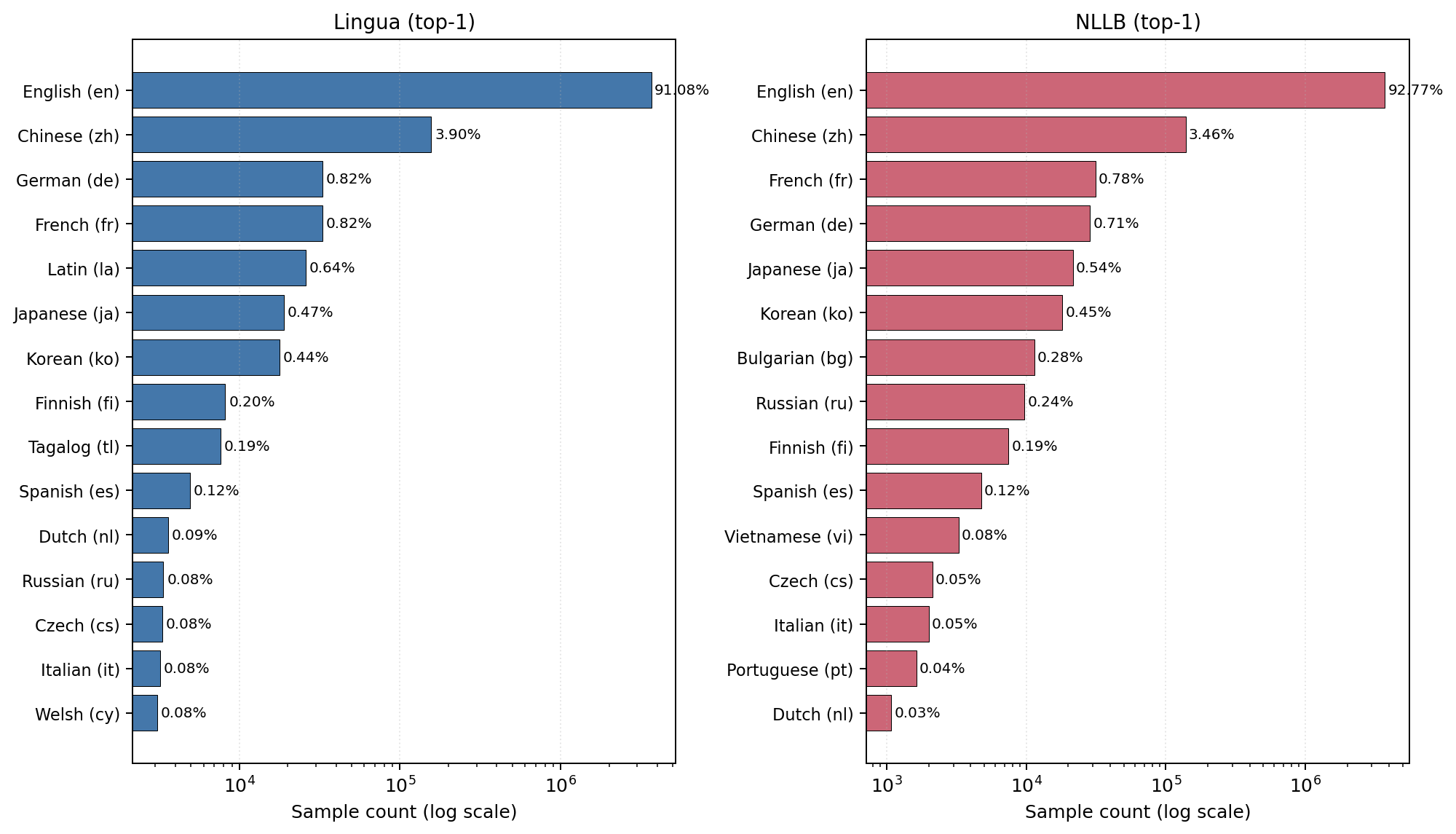}
\caption{\textbf{Language distribution of \dcvlm{} pool}.
Top-15 languages by per-sample top-1 prediction from \textbf{Lingua} (left, blue) and \textbf{NLLB} (right, red), shown on a log-scaled x-axis.
The pool is dominated by English ($91.1\%$ Lingua / $92.8\%$ NLLB), followed by Chinese ($3.9\%$ / $3.5\%$); the remaining $\sim$5\% is spread across 73 languages (Lingua) or 114 languages (NLLB), with French, German, Japanese, Korean, and Russian forming the next tier.}
\label{fig:lang-distribution}
\end{figure}
}

\title{DataComp-VLM:\\Improved Open Datasets for Vision-Language Models}

\author{%
\begin{tabular}{c}
\hspace*{-1.4cm}
\textbf{Matteo Farina}$^{*1,2}$\quad \textbf{Vishaal Udandarao}$^{*2,3}$\quad \textbf{Thao Nguyen}$^{*15}$\quad \textbf{Selim Kuzucu}$^{\dagger,5}$\quad \textbf{Maximilian Böther}$^{\dagger,7}$\\
\textbf{Andreas Hochlehnert}$^{\dagger,2}$\quad \textbf{Adhiraj Ghosh}$^{\dagger,2}$\quad \textbf{Marianna Nezhurina}$^{\dagger,8,9}$\quad \textbf{Karsten Roth}$^{\dagger,10}$\\
\textbf{Joschka Struber}$^{2}$, \textbf{Yuhui Zhang}$^{4}$, \textbf{Sebastian Dziadzio}$^{2}$, \textbf{Elaine Sui}$^{4}$, \textbf{Soumya Jahagirdar}$^{2}$,\\
\textbf{Dhruba Ghosh}$^{4}$, \textbf{Hasan Hammoud}$^{11}$, \textbf{Thomas De Min}$^{1}$, \textbf{Simone Caldarella}$^{1}$, \textbf{Jehanzeb Mirza}$^{12}$,\\
\textbf{Sedrick Keh}$^{13}$, \textbf{Mehdi Cherti}$^{8,9}$, \textbf{Hilde Kuehne}$^{2}$, \textbf{Bernt Schiele}$^{5,6}$, \textbf{Serena Yeung-Levy}$^{4}$,\\
\textbf{Muhammad Ferjad Naeem}$^{6}$, \textbf{Federico Tombari}$^{6}$, \textbf{Ana Klimovic}$^{7}$, \textbf{Elisa Ricci}$^{1,14}$, \textbf{Matthias Bethge}$^{2}$,\\
\textbf{Sewoong Oh}$^{15}$, \textbf{Ameya Prabhu}$^{2}$, \textbf{Alessio Tonioni}$^{6}$, \textbf{Jenia Jitsev}$^{8,9}$, \textbf{Massimiliano Mancini}$^{1}$\\
\textbf{Ludwig Schmidt}$^{\ddagger,4}$\quad \textbf{Nikhil Parthasarathy}$^{\ddagger,10}$\\[4pt]
{\small $^{*}$Project leads\quad $^{\dagger}$Core contributors\quad $^{\ddagger}$Equal supervision}
\end{tabular}%
}

\begin{document}

\maketitle
\customfootnotetext{}{%
\noindent\hspace*{-1.8em}%
$^{1}$University of Trento,
$^{2}$T\"ubingen AI Center, University of T\"ubingen,
$^{3}$University of Cambridge,
$^{4}$Stanford University,
$^{5}$Max Planck Institute for Informatics,
$^{6}$Google,
$^{7}$ETH Z\"urich,
$^{8}$LAION,
$^{9}$Juelich Supercomputing Center (JSC), Research Center Juelich (FZJ),
$^{10}$Google DeepMind,
$^{11}$KAUST,
$^{12}$MIT,
$^{13}$Toyota Research Institute,
$^{14}$Fondazione Bruno Kessler,
$^{15}$University of Washington%
}

\begin{center}
    \vspace{-2em}
    \huggingface{} \href{https://huggingface.co/collections/mlfoundations/dcvlm}{Models \& Data} \quad \faGithub \href{https://github.com/mlfoundations/dcvlm}{Code} \quad
    \faGlobe \href{https://www.datacomp.ai/dcvlm}{Website} \quad \\
    \vspace{3pt} Contact: \texttt{m.farina@unitn.it}, \texttt{vishaal.udandarao@bethgelab.org}
    
\end{center}

\begin{abstract}
Building performant Vision-Language Models (VLMs) requires carefully curating large-scale training datasets, yet the community lacks systematic benchmarks for evaluating such curation strategies.
We introduce \underline{D}ata\underline{C}omp for \underline{VLM}s (\dcvlm), a benchmark for controlled data-centric experiments to improve VLM training. 
As part of \dcvlm, we collect 160 datasets spanning four data types---image-caption pairs, multimodal interleaved documents, text-only, and instruction-tuning data---into a corpus of 6T multimodal tokens.
\dcvlm allows participants to test curation strategies (filtering, mixing, formatting, sampling) across 1B--8B models and 6.25B--200B token budgets. Models are then evaluated on a carefully selected suite of up to 52 downstream benchmarks across 9 domains.
We conduct extensive experiments on \dcvlm and find that data mixing, \emph{not} filtering, is key to a high-quality training dataset: instruction-heavy mixtures \emph{scale better} than caption-heavy ones, with gains widening at larger scales. 
The resulting dataset, \textsc{\dcvlm-Baseline}, enables training an 8B VLM to $63.6\%$ accuracy on our 33-task core suite with 200B training tokens. 
Compared to \textsc{FineVision}, the state-of-the-art open VLM training dataset, this represents an improvement of %
$+5.4$pp.
\dcvlm and all accompanying artifacts will be made publicly available \href{https://github.com/mlfoundations/dcvlm}{here}.%

\end{abstract}

\section{Introduction}\label{sec:intro}

The performance of foundation models is fundamentally shaped by the composition and quality of their pretraining\footnote{The VLM literature is highly fragmented in its terminology, with stages variously called ``pretraining,'' ``alignment,'' or ``instruction tuning'' depending on the data type used. We refer to VLM ``pretraining'' as in \cite{chen2024expanding,zhu2025internvl3,wang2025internvl3}, i.e., the first or only multimodal training stage of a VLM, starting from an independently pretrained vision encoder and language model, bridged by a randomly initialized connector~\citep{liu2023visual}. When a VLM is trained in multiple stages, ``pretraining'' denotes the first stage; otherwise, it denotes the sole training cycle.}data~\citep{gadre2023datacomp,li2024datacomp,grattafiori2024llama,penedo2024fineweb,nguyen2022quality,fang2022data,udandarao2025data,ghosh2025concept,penedo2023refinedweb,schuhmann2022laion}.
This has led to a rise of systematic studies of pretraining data curation, including 
DataComp~\citep{gadre2023datacomp} for contrastive vision-language models, DCLM~\citep{li2024datacomp}, Nemotron-CC~\citep{su2025nemotron}, and FineWeb~\citep{penedo2024fineweb} for language models, and OlmoASR~\citep{ngo2025olmoasr} in the speech domain.
The core design principle of these works is to fix model architecture and pretraining procedure while varying \emph{only} the data, enabling isolated measurement of data-centric interventions. 
However, progress in \emph{autoregressive} vision-language models (VLMs) has mostly focused on novel  architectures~\citep{deitke2025molmo,xu2025qwen3,hong2025glm,gemmateam2025gemma3technicalreport,coreteam2025mimovltechnicalreport,deshmukh2025nvidia,yang2025kwai,wu2024deepseek,abouelenin2025phi,bevli2026falcon,beyer2024paligemma}, training recipes~\citep{tong2024cambrian,zhu2025internvl3,wang2025internvl3,an2025llava,cho2025perceptionlm,liu2023visual,liu2024improved,liu2024llavanext,li2025xiaomi,marafioti2025smolvlm,mckinzie2024mm1,guo2025seed1,lu2024deepseek,steiner2024paligemma}, or evaluation protocols~\citep{duan2024vlmevalkit,zhang2025lmms,joshi2026datbench,ghosh2025onebench}, treating data as a second-class citizen. 
The data curation strategies behind their success (which datasets to include, how to filter them, what ratios to mix them in) remain poorly understood and largely irreproducible~\citep{gemmateam2025gemma3technicalreport,mckinzie2024mm1, tong2024cambrian,zhu2025internvl3,wang2025internvl3,an2025llava,cho2025perceptionlm,joshi202620}.
Our goal is precisely to fill this gap and enable \emph{open} data curation research for the latest class of modern autoregressive VLMs.

\fighero
Several factors make VLM data curation more challenging compared to other domains.
\textbf{First}, unlike early text or vision-language models that often train directly on raw web crawls (e.g., CommonCrawl), modern VLMs are typically trained by aggregating existing datasets from a wide variety of data types---web-crawled image-caption pairs, interleaved multimodal documents, text-only corpora, and multimodal instruction-tuning data---that differ in quality and downstream utility. 
Because these datasets have already undergone varying degrees of upstream curation, what actually drives quality under this aggregation-based regime---filtering, mixing ratios, or something else---remains an open question. Indeed, existing models largely sidestep it, drawing on a single data type~\citep{liu2023visual} or, at most, an ad-hoc subset~\citep{alayrac2022flamingo,bai2025qwen25vltechnicalreport}.
\textbf{Second}, existing open training datasets~\citep{wiedmann2025finevision,an2025llava,deshmukh2025nvidia,zhao2023svit,deitke2025molmo,tong2024cambrian} operate at the scale of millions of samples, far below the trillions of tokens used by state-of-the-art (SoTA) models~\citep{team2025kimi,wang2025internvl3,bai2025qwen3,coreteam2025mimovltechnicalreport}. This limits the scope of curation experiments that can be conducted. 
\textbf{Third}, the interaction between data types, model scale, and 
training budget creates a 
design space that is too large for exhaustive experimentation.
\textbf{Fourth}, VLM evaluation lacks standardization~\citep{joshi2026datbench}: different papers use different benchmark suites, making fair comparisons across datasets difficult.

To mitigate these challenges and enable controlled comparisons, we introduce \underline{D}ata\underline{C}omp for \underline{VLM}s (\dcvlm), the first benchmark designed to systematically study data curation strategies within a realistic VLM-practitioner's paradigm.
\dcvlm provides the following:
\begin{enumerate}[leftmargin=*, itemsep=0pt, topsep=0pt, parsep=0pt, partopsep=0pt]
    \item A \textbf{standardized data pool} of 160 existing datasets spanning four data types: image-caption pairs, multimodal documents, text-only data, and (multimodal) instruction-tuning data. 
    Our pool contains 6T multimodal tokens, enabling a diverse range of data-centric experiments.
    \item A \textbf{principled scaling ladder} spanning 1B--8B model parameters and 6.25B--200B training tokens. This enables researchers to test curation strategies across a wide range of compute scales.
    \item A \textbf{comprehensive evaluation protocol} with 52 downstream benchmarks organized across 9 domains, split into validation, core, and extended tiers, filtered for stability and reliability. 
\end{enumerate}
 
Using our benchmark, we conduct more than 1,000 experiments yielding multiple findings, including:

\smallsec{Mixing, \emph{not} filtering, is the dominant lever.}
Recent VLM technical reports often apply additional downstream quality filters (e.g., CLIP-score or image quality) on top of existing public datasets~\citep{zhang2026penguin,team2025kimi,yu2026minicpm}. Yet, through controlled experiments with common quality filters, we find that such downstream filtering provides diminishing, and sometimes negative, returns (\cref{sec:filtering}). We trace this to the modern data landscape: unlike models trained directly on raw web crawls (e.g., CommonCrawl), today's VLM datasets have already undergone moderate to significant upstream curation. While curating VLM training data directly from raw pools remains an important direction for future work, our results show that applying additional filters to already-curated data is largely ineffective. In contrast, optimizing mixture ratios, which specifically interpolate instruction-tuning and image-caption proportions, yields significant, scale-dependent gains: instruction-heavy mixtures scale better than caption-heavy ones, and this gap widens with model size and token budget (\cref{sec:mixing}).
 
\smallsec{Pretraining decisions reliably transfer after supervised fine-tuning and across backbones.}
We show that pretraining performance predicts post-SFT performance with near-perfect fidelity (Pearson $r = 0.99$ across 54 SFT runs), and that our findings are robust to the choice of LLM initialization, i.e., initializing from Qwen2.5-Base or Qwen2.5-Instruct \cite{qwen25}
produces similar data rankings.
This validates the use of pretraining-only metrics for data curation research with DCVLM (\cref{sec:controls}). 
 
Our controlled experiments yield \textsc{\dcvlm-Baseline}, a new state-of-the-art open VLM training dataset (\cref{fig:hero}). 
At the \texttt{x-large} scale (8B model, 200B tokens), a \textsc{\dcvlm-Baseline}-trained model achieves $63.6\%$ on our core set, outperforming \fv~\cite{wiedmann2025finevision}, the previous best open dataset, by ${+5.4}$ pp (\cref{sec:results}).
We release the full data pool, evaluation suite, model checkpoints at four scales, and all experimental infrastructure to serve as a reproducible testbed for future research.

\section{Related Work}\label{sec:related}

\smallsec{Vision-Language Pretraining.}
Modern VLMs adopt a modular architecture consisting of a pretrained vision encoder, a language model backbone, and a connector~\citep{liu2023visual,li2024llava,an2025llava,zhu2025internvl3,wang2025internvl3,bai2025qwen25vltechnicalreport,bai2025qwen3,shapourian2026zaya1}.
Originally, ``pretraining'' involved training the connector on large-scale data, predominantly image-caption pairs~\citep{an2025llava}. %
In contrast, recent SoTA models train all parameters~\cite{zhu2025internvl3, wang2025internvl3} and incorporate diverse data types. %
However, the precise mixture ratios, filtering criteria, and formatting choices largely remain proprietary and poorly documented across leading VLMs, motivating our %
benchmark.

\smallsec{Benchmarking Data Curation.}
DataComp~\citep{gadre2023datacomp} and DataPerf~\citep{mazumder2023dataperf} established the paradigm of fixing model architecture and training procedure while \textit{varying only data}.
DCLM~\citep{li2024datacomp} extended this paradigm to language models, demonstrating that a fasttext classifier trained on high-quality samples can substantially boost performance.
FineWeb~\citep{penedo2024fineweb} and its educational-quality variant, FineWeb-Edu, showed similar gains through filtering. 
Generally, quality-based filtering has shown strong results for text~\citep{penedo2024fineweb,li2024datacomp} and image-text pairs~\citep{gadre2023datacomp,wang2025unifilter}.
Common approaches include CLIP-score filtering~\citep{hessel2021clipscore}, image quality assessment~\citep{mao2026wan,wu2025qwen}, text quality classifiers~\citep{su2025nemotron,deshmukh2025nvidia}, and multimodal quality estimators~\citep{wang2025unifilter}. 
Beyond filtering, prior works have also explored data mixing approaches such as domain weighting~\citep{xie2023doremi,olmo20242}, mixture optimization~\citep{chen2026olmix,berasi2026linear,diao2025nemotron,kang2024autoscale,ye2024data,liu2024regmix}, and temperature-sampling~\citep{deitke2025molmo,clark2026molmo2}. %
Despite recent released datasets (e.g., \fv~\citep{wiedmann2025finevision}, \textsc{Cauldron}~\citep{laurenccon2024matters}), there exists no systematic study on filtering and mixing strategies in the VLM setting.  %
Our work fills this gap by 
providing the first \emph{scale-aware} study of data curation for VLMs.

\section{The \dcvlm Benchmark}\label{sec:methodology}

\begin{figure}[t!]
  \centering
  \includegraphics[width=0.9\linewidth]{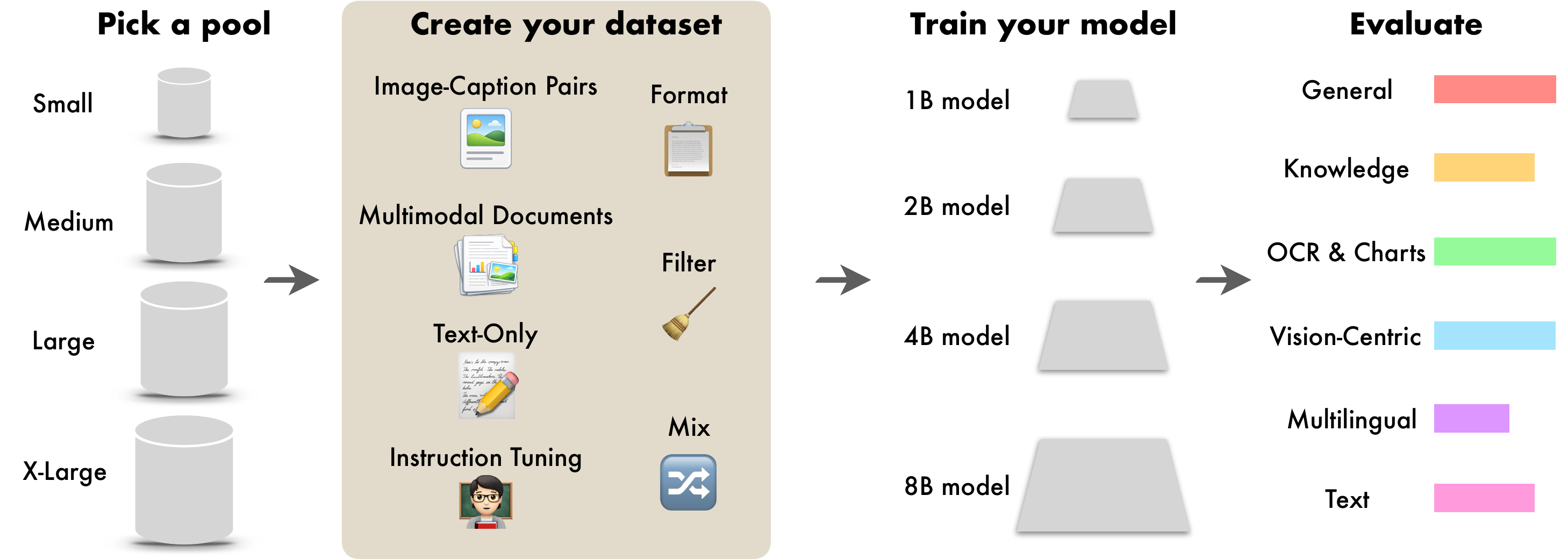}
  \caption{\textbf{\dcvlm allows researchers to construct effective multimodal datasets.} Participants can choose one of four scales (\texttt{small}, \texttt{medium}, \texttt{large}, and \texttt{x-large}) according to their compute availability. 
  We provide tools to format, filter, and mix the data pool so that participants can create their own datasets. 
  The resulting datasets are then used to train an autoregressive VLM using a fixed training recipe. 
  Models are comprehensively evaluated across a broad spectrum of capabilities.}
  \label{fig:dcvlm-schema}
  \vspace{-0.5cm}
\end{figure}

\dcvlm provides a controlled framework (\cref{fig:dcvlm-schema}) for constructing VLM training sets. 
We fix the model and training recipe, and participants propose ways to filter and mix data from our pool. 
We next describe the pool (\cref{sec:pool}), training recipe (\cref{modeltrainingrecipe}), scales (\cref{sec:scales}), and evaluation (\cref{sec:eval-protocol}).

\subsection{Data Pool Construction}\label{sec:pool}

Our data pool aggregates 160 publicly available datasets organized into four data types. For the full list of source datasets, pool composition, visualizations, and sample and token counts, see\textcolor{black}{~\cref{app:dataset_pool}}.
 
\ding{172} \textbf{Image-caption pairs} form the largest component, with great variability in their constituent datasets.
At one end, sources like DataComp-1B~\citep{gadre2023datacomp} and ReLAION-2B~\citep{schuhmann2022laion} provide abundant, CLIP-score-filtered image-alt-text pairs from web crawls. 
At the other, datasets like the synthetic ShareGPT-4o~\citep{cui2025comprehensive} and the human-annotated Pixmo-Cap~\citep{deitke2025molmo} offer fewer yet higher quality samples.

\ding{173} \textbf{Multimodal interleaved documents} consist of web-crawled interleaved image-text sequences as they appear on websites, PDF documents, and academic papers.
Sources include MINT-1T-HTML~\citep{awadalla2024mint}, MINT-1T-PDF~\citep{awadalla2024mint}, WanJuan~\citep{he2023wanjuan}, OmniCorpus~\citep{li2024omnicorpus}, and Multimodal-Textbook~\citep{zhang2025textbook}.
These are the least curated sources in our pool. 
Most are scraped directly from the web with minimal URL-based and heuristic filtering, and thus tend to have lower quality scores on average. 

\ding{174} \textbf{Text-only data} preserve the language model's capabilities during multimodal training, following recent VLMs~\citep{zhu2025internvl3,bai2025qwen3}. Examples include FLAN~\citep{longpre2023flan}, SlimOrca~\citep{lian2023slimorca}, and Dolly~\citep{conover2023dolly}, alongside image-free science and knowledge sources such as Numina-Math-1.5~\cite{numina_math_15} and xCoder80k~\cite{wang2024xcoder}.

\ding{175} \textbf{Multimodal instruction-tuning data} comprise single or multi-turn instruction-tuning datasets, typically with human-written or model-generated question-answer pairs grounded in one or multiple images.
We manually categorize these into eight capabilities following \cite{wiedmann2025finevision}: knowledge, chart \& table understanding, general-QA, grounding \& counting, math, naive OCR, OCR-QA, and science.
For a complete breakdown of the capability distribution of instruction-tuning data, see\textcolor{black}{~\cref{app:pool-composition}}.

Our \dcvlm{} pool contains 6T multimodal tokens (measured using the InternVL-2.5~\citep{chen2024expanding} tokenizer). It is highly heterogeneous in source quality, data types, instruction-tuning capabilities (e.g., grounding, OCR, chart and table understanding, captioning), visual and textual domains (e.g., natural, synthetic, tabular images), and languages (over $20$ including English and Chinese, see \textcolor{black}{\cref{app:multilingual}}). This heterogeneity is \emph{deliberate}: it lets participants study curation recipes in a realistic setup with several confounders to control for.
To avoid train-test overlap, we decontaminate our entire pool against our \textbf{Extended} eval suite of 52 benchmarks (\cref{sec:eval-protocol}): multimodal samples are filtered with a ResNet-50 SSCD embedding model~\citep{pizzi2022self} (cosine-sim $>0.75$ to any test image) and text-only samples with MinHash~\citep{broder1997resemblance} Jaccard similarity ($>0.55$). 
The exact details of decontamination are in \cref{app:decontamination}.

\subsection{Model Architecture and Training Recipe}\label{modeltrainingrecipe}
To ensure \dcvlm employs a state-of-the-art training recipe, we use an architecture that mimics InternVL3 models \cite{zhu2025internvl3}: an InternViT-300M vision encoder \cite{chen2024expanding}, a 2-layer MLP projector, and a Qwen2.5-Base language model~\citep{qwen25} (we show that our central findings transfer to Instruct backbones as well in~\cref{sec:controls}).
We adopt \textit{AnyRes}~\citep{liu2024llavanext} tiling, where images are dynamically split into $448{\times}448$ tiles, each encoded into 256 visual tokens after pixel shuffling \cite{chen2024expanding,shi2016real}. 
We use the AdamW \cite{loshchilov2017decoupled} optimizer, a linear ${3}{\%}$ warmup, and a cosine decay with peak learning rate of $2\times10^{-5}$, identified after an initial sweep to ensure optimal hyperparameters. 
For more details, refer to \cref{app:architecture,app:hparams}.

\subsection{Competition Scales and Design Principles}\label{sec:scales}

\begin{table}[!t]
\centering
\caption{\textbf{\dcvlm scales.} Each scale specifies model size ($N$), number of training tokens ($D$), and token size of the original pool to be used for curation (`Pool'). We also present the vision encoder and language model that we initialize training runs from, along with compute estimates (`H100 hrs').}
\label{tab:scales}
\begin{adjustbox}{width=.9\linewidth}
\footnotesize
\begin{tabular}{lccccccc}
\toprule
\textbf{Scale} & $\boldsymbol{N}$ & $\boldsymbol{D}$ & \textbf{Vision init.} & \textbf{LLM init.} & \textbf{H100 hrs} & \textbf{Pool} \\
\midrule
\texttt{small}   & 1B & 6.25B & InternViT-300M & Qwen2.5-0.5B & 80    & 187.5B \\
\texttt{medium}  & 2B & 25B   & InternViT-300M & Qwen2.5-1.5B   & 640  & 750B \\
\texttt{large}   & 4B & 100B  & InternViT-300M & Qwen2.5-3B   & 5,120  & 3T \\
\texttt{x-large} & 8B & 200B  & InternViT-300M & Qwen2.5-7B   & 20,480 & 6T \\
\bottomrule
\end{tabular}
\end{adjustbox}
\end{table}

A key principle of \dcvlm is to evaluate data curation strategies \emph{across scales}, because findings at small scales may not transfer to larger ones~\citep{goyal2024scaling,mizrahi2025language,nezhurina2025scaling,maithinking1}.
To simultaneously (\textit{i}) approach the scale of foundation models like InternVL-3~\citep{zhu2025internvl3} and (\textit{ii)} ensure accessibility for researchers with fewer resources, we define four scales: \texttt{small}, \texttt{medium}, \texttt{large}, and \texttt{x-large}.
Model sizes and token budgets are illustrated in \cref{tab:scales}. %
We design the \texttt{small}, \texttt{medium}, and \texttt{large} scales such that a step corresponds to an $8\times$ compute increase: models become $2\times$ larger and tokens increase by $4\times$.
At the \texttt{x-large} scale, our entire pool of 6T tokens is the candidate for dataset construction.
We design all scales to fix pool-to-training token ratio at ${30}{\times}$,
i.e, the pool always contains $30\times$ more tokens than the training budget.
The primary reason for keeping this ratio constant is to enable participants to experiment with aggressive filtering at all scales while hitting a constant number of data repetitions.

\subsection{Evaluation Protocol}\label{sec:eval-protocol}
Participants in \dcvlm can evaluate models on up to 52 benchmarks. 
To get reliable signal, we
start from a candidate set of 65 benchmarks, which we categorize across 9 domains based on the majority consensus of prior work \cite{zhu2025internvl3, chen2024expanding, wang2024qwen2,mckinzie2024mm1,tong2024cambrian}: General Understanding, Knowledge-Centric, OCR \& Charts, Vision-Centric, Multilingual, Text-Only, Safety, Hallucination, and Reasoning benchmarks. 
We then filter them for (\textit{i}) \textit{stability}, removing those with high seed variance~\citep{udandarao2025active,madaan2024quantifying}, and (\textit{ii}) \textit{monotonicity}, removing those that do not improve from \texttt{small} to \texttt{medium} scales~\citep{heineman2025signal,penedo2024fineweb}.
We organize benchmarks into three nested tiers, each a superset of the previous: a \textbf{Validation set}, used for rapid iteration (13 benchmarks), a \textbf{Core set}, the primary tier used for main results (33 benchmarks), and an \textbf{Extended set} (52 benchmarks), including all benchmarks for comprehensive analysis. 
Safety, Hallucination, and Reasoning are deferred to the extended tier, as they are typically targeted by (and thus most relevant for) post-training methods. 
Unless otherwise specified, we report the average accuracy across all benchmarks in a given tier. %
For full details of benchmark selection, see \cref{app:eval_details}.

\section{Towards a Strong Baseline on \dcvlm}\label{sec:experiments}

We now present a suite of controlled experiments showing how to obtain a strong baseline dataset on \dcvlm{}, along two primary axes: 
data filtering (\cref{sec:filtering}) and data mixing (\cref{sec:mixing}). 
For additional axes (including data formatting, synthetic captions, and temperature sampling), refer to~\cref{app:filtering-details}.
We also run control experiments validating the generality of our results (\cref{sec:controls}). 
Unless specified, all filtering experiments use a base mixture of $75$\% image-caption, $18$\% text-only, $4$\% multimodal documents, and $3$\% instruction-tuning data, derived by length-proportional sampling across the pool. 
In this section, we always report results on our 33-task Core evaluation suite.

\subsection{Data Filtering}\label{sec:filtering}
Quality-based filtering has been central to pretraining strong language~\citep{li2024datacomp,penedo2024fineweb,olmo20242,olmo2025olmo} and CLIP models~\citep{gadre2023datacomp,fang2023data,udandarao2025active,evans2024data}, hence a natural question is whether these gains transfer to VLMs as well.
We answer this question in the \emph{negative} 
by testing \textbf{more than 60 filter configurations} at both \texttt{small} and \texttt{medium} scales (for an exhaustive report, refer to {\cref{app:filtering-dont-help-sec}}). 
To illustrate our findings, here we report and discuss \texttt{medium} scale results for filters shown to be successful in prior work (in {\cref{app:filtering-dont-help-sec}}, we describe several other variants across scales, yielding the same conclusions):
\begin{itemize}[leftmargin=*, itemsep=2pt, topsep=0pt, parsep=0pt, partopsep=0pt]
    \item \textbf{CLIP-score}. We experiment with filtering image-caption pairs according to three different CLIP models: OpenAI's CLIP ViT-L/14~\citep{radford2021learning}, DFN-CLIP~\citep{fang2023data}, and SigLIP-2-B/16@384~\citep{tschannen2025siglip}. 
    \item \textbf{Text quality classifiers}. We experiment with filtering samples according to the quality of their constituent text snippet(s), as judged by three classifiers: DCLM's fasttext classifier~\cite{li2024datacomp}, as well as NVIDIA's Nemotron and Mixtral educational-quality classifiers~\cite{su2025nemotron}. 
    \item \textbf{Multimodal filters}. We additionally experiment with (i) filtering with two UniFilter models \cite{wang2025unifilter} (Qwen2.5-1.5B and Qwen3-0.6B), and (ii) filters grounded in perplexity~\citep{ankner2024perplexed}: text-only perplexity (computed on text tokens by \textit{excluding} image tokens), multimodal perplexity (computed on text tokens by \textit{including} image tokens), and Conditional Mutual Information \cite{lee2026selective}, which measures their difference (i.e., the reduction in perplexity with and without image tokens).
\end{itemize}

\begin{figure}[t!]
    \centering
    \includegraphics[width=\linewidth]{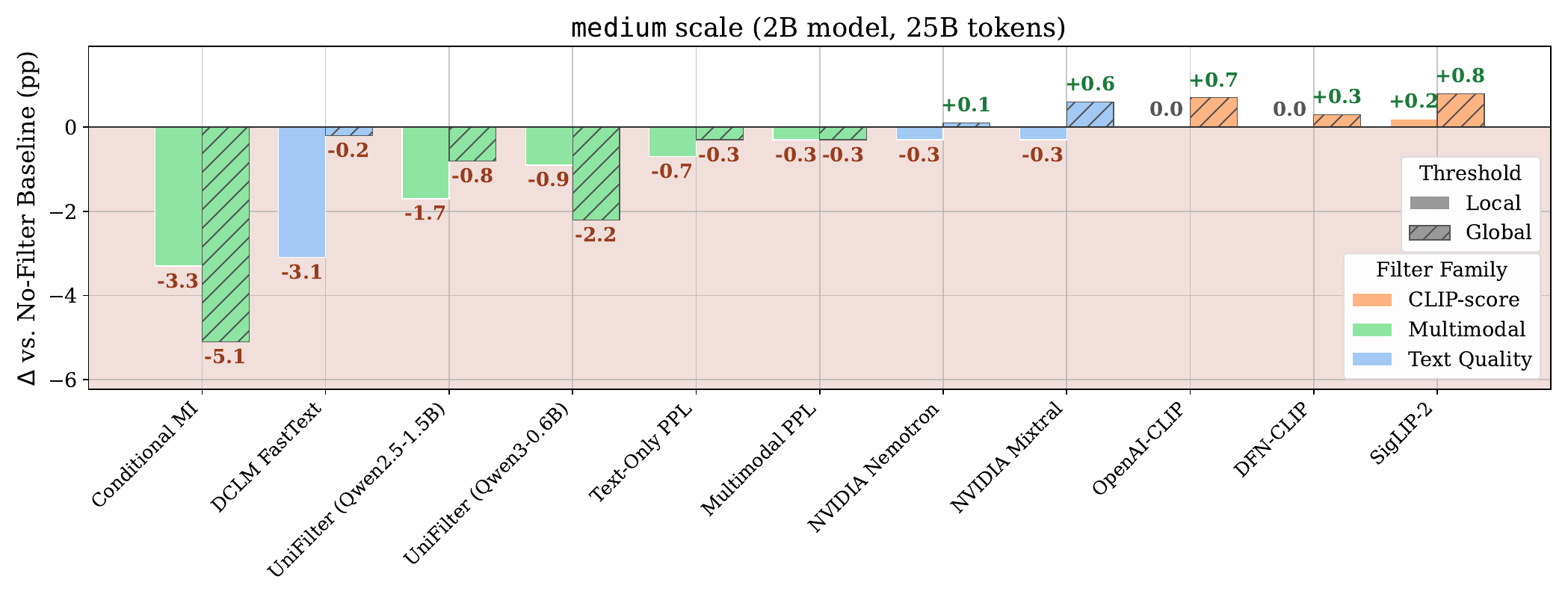}

    \caption{\textbf{Filtering rarely helps, but changing the data composition does move performance substantially.}
     Established data filtering techniques do \emph{not} significantly outperform a no-filter baseline. 
     This observation holds consistently at both the \texttt{small} (\cref{fig:filtering-small}) and \texttt{medium} scales of our benchmark.
    At the same time, inducing a different data mixture via global filtering (hatched bars) leads to significant performance variations compared to locally filtered datasets (solid bars).}
    \label{fig:filtering}
\vspace{-0.5em}
\end{figure}
 
Importantly, we study two different filtering paradigms to isolate the impact of data filtering and that of implicit data mixing:
\ding{172} \textbf{Local filtering}, which computes filtering percentile thresholds independently within each source dataset. 
This preserves the global mixture by construction: every dataset loses the same sample fraction, and 
\ding{173} \textbf{Global filtering}, which computes a single filtering threshold across the entire pool of samples to which the filter can be applied. 
Because different data sources have systematically different score distributions, 
a global cut implicitly \emph{reshapes the data mixture}. 
Following prior evidence that smaller models benefit from more aggressive filtering~\citep{allal2025smollm2,mizrahi2025language}, we retain the top-10\% of samples at the \texttt{small} scale, and the top-40\% at the \texttt{medium} scale. 

\cref{fig:filtering} illustrates the results. 
We make two key observations:
(\textit{i}) 
regardless of whether the mixture is held fixed, no quality filter we tested produces a robust and significant improvement over a no-filter baseline; and (\textit{ii}) local and global filtering yield notably different results. We expand on each below.

\textbf{Filtering rarely helps, but why?} The best filtering outcome is given by SigLIP-2 when globally filtering image-text pairs (rightmost bar in \cref{fig:filtering}), 
yet this result is defined by a marginal $+0.8$pp improvement, far below the gains one would apriori expect from quality-based filtering~\citep{schuhmann2022laion,li2024datacomp,gadre2023datacomp,su2025nemotron,udandarao2025active,evans2024data}.
Other filters 
either leave the baseline mostly unchanged or actively hurt performance. 
This observation holds across both \texttt{small} and \texttt{medium} scales (see \cref{fig:filtering-small} for the \texttt{small} figure).

\begin{wrapfigure}[17]{r}{0.4\textwidth}
    \centering
    \includegraphics[width=\linewidth]{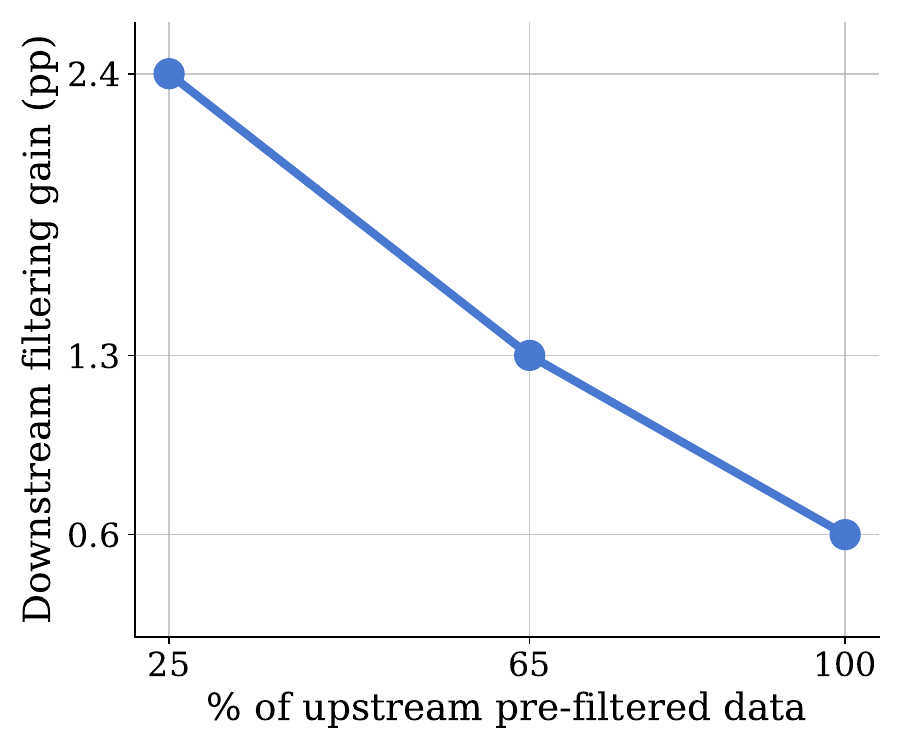}
    \caption{Upstream filtering leads to diminishing returns from additional (i.e., ``downstream'') filtering.}
    \label{fig:raw-vs-filtered}
\end{wrapfigure}

This failure is surprising, especially in light of strong results from prior works.
We hypothesize this is because \emph{there is no significant noise to remove from our base pool}: unlike raw Common Crawl (used in DCLM~\cite{li2024datacomp}) or raw web-crawled image-text pairs (used in DataComp-CLIP~\citep{gadre2023datacomp}), existing VLM training sets 
aggregate datasets that have already undergone a level of upstream filtering (e.g. CLIP-score filtering) by their original creators, and our pool follows this data collection process. 
To validate our hypothesis, we create three sub-pools from our original pool, varying the effective percentage (25\%, 65\%, and 100\%) of ``pre-filtered'' data samples in the mixture (see\textcolor{black}{~\cref{app:filtering-exp}} for more details). 
From each of these sub-pools, we create three more training sets by applying further CLIP-score filtering to the image-caption data. 
For each pair of datasets, we then train \texttt{small} scale models and measure the performance gain due to downstream filtering (\cref{fig:raw-vs-filtered}). For 25\% pre-filtering (i.e., when the sub-pool is dominated by unfiltered data), the gain is significant ($+2.4$pp). However, this decreases the more the 
sub-pool is pre-filtered (dropping to $+1.3$pp at 65\% and $+0.6$pp at 100\%).
In other words, \textit{additional filtering on top of already-curated data operates in a regime of diminishing returns}.

\textbf{Interaction between filtering and implicit mixing.} The second takeaway from \cref{fig:filtering} is \emph{local and global filtering produce very different results}. 
The inconsistent trends suggests that global filtering is not a reliable strategy. 
However, given significant performance fluctuations between local and global filtering, we hypothesize the underlying mixture distribution is the lever that dictates performance. %

\subsection{Data Mixing}\label{sec:mixing}

Having established that filtering over the base mixture provides negligible gains on our pool, we turn to data mixing, i.e., the allocation of training samples across data types, as our primary curation lever.

\textbf{Setup.}
We optimize the mix along an important axis based on prior work~\citep{liu2023visual,karamcheti2024prismatic,mckinzie2024mm1,zhang2024mm1,shukor2025scaling,chen2024internvl,tong2024cambrian}:
the ratio of \emph{image-caption pairs} to \emph{instruction-tuning data}. 
Text-only samples and multimodal documents are fixed at $15$\% and $5$\%, respectively, as supporting components. 
Here, we study three ratios along the image-caption $\leftrightarrow$ instruction-tuning axis: (\textit{i}) a \textbf{Caption-heavy} mixture with $65$\% image-caption pairs and $15$\% instruction-tuning data; a (\textit{ii}) \textbf{Balanced} mixture with $40$\% image-caption and $40$\% instruction; and
(\textit{iii}) an \textbf{Instruction-heavy} mixture with $10$\% image-caption and $70$\% instruction-tuning data (see \cref{app:data-mixing-fine-sweep} for finer sweeps across scales).
Each mixture is evaluated across a scaling grid of 3 model sizes (1B, 2B, 4B) $\times$ 3 token budgets (6.25B, 12.5B, 25B). %

\begin{figure}[!t]
    \centering
    \includegraphics[width=\linewidth]{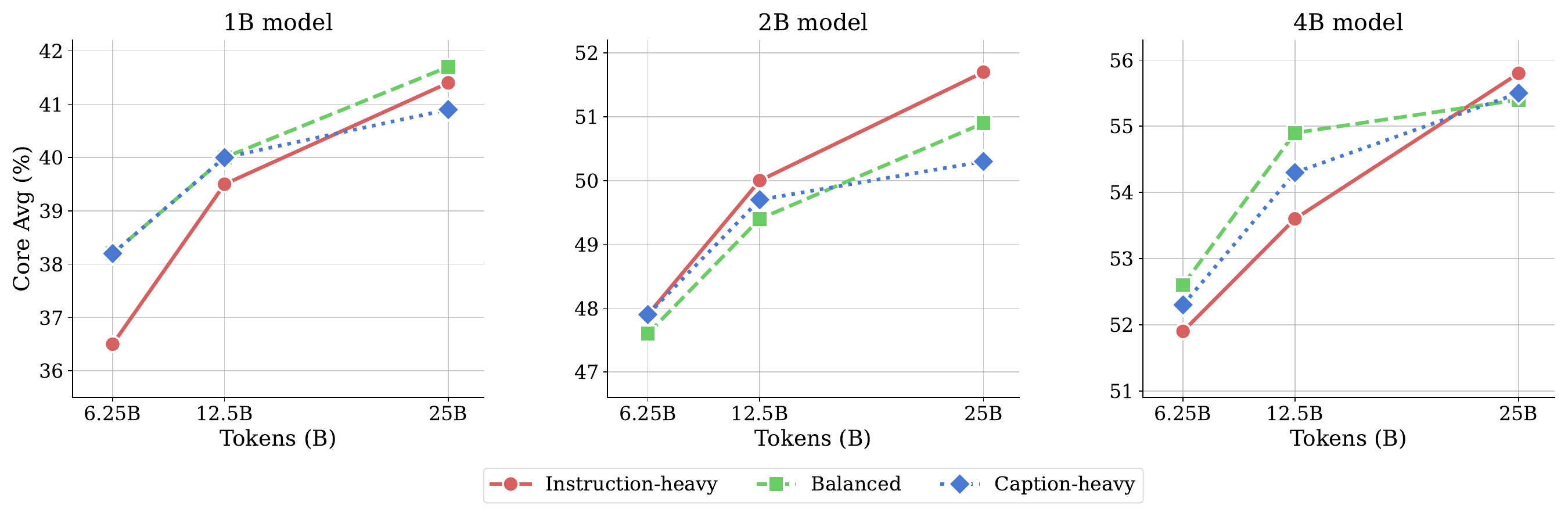}
    \caption{\textbf{Instruction-heavy mixtures scale better with compute.}
    For the 1B model (\textit{left}), the Instruction-heavy mix (red) starts as the worst mixture with 6.25B training tokens, but recovers quickly up to becoming the second-best with 25B training tokens. 
    For the 2B model (\textit{middle}), all mixtures have comparable performance with 6.25B tokens, but performance gains consistently grow in favor of the Instruction-heavy mixture as training tokens grow.
    For the 4B model (\textit{right}), yet again, the Instruction-heavy mix starts as the worst with 6.25B tokens, and becomes the best at 25B tokens.}
    \label{fig:mixing-panels}
    \vspace{-1.75em}
\end{figure}

\begin{wraptable}{r}{0.4\linewidth}
    \centering
    \caption{Instruction-heavy mixes are robust to moderate data repetitions.}
    \label{tab:it-repeats}
    \small
    \begin{tabular}{lc}
        \toprule
        \textbf{Configuration} & \textbf{Core Avg} \\
        \midrule
        Instruction-heavy, unique    & 51.7 \\
        Instruction-heavy, $\sim$2$\times$  & 50.2 \\
        Instruction-heavy, $\sim$4$\times$  & 49.8 \\
        Instruction-heavy, $\sim$8$\times$  & 48.6 \\
        \midrule
        \multicolumn{2}{l}{\textit{Other mixes (unique data)}} \\
        \quad Balanced & 50.9 \\
        \quad Caption-heavy & 50.3 \\
        \quad Base mix           & 48.8 \\
        \bottomrule
    \end{tabular}
\end{wraptable}

\textbf{Data mixing cannot be scale agnostic.}\label{sec:scale-mixing} \cref{fig:mixing-panels} reveals a striking interaction between data mixture and compute scale:
as both model size and token budget increase, the Instruction-heavy mix exhibits a markedly steeper scaling slope.
It starts as the \emph{worst} mixture at 1B$\times$6.25B (\texttt{small} scale) but becomes the \emph{best} at 2B$\times$25B (\texttt{medium} scale), and remains so at 4B$\times$25B. 
This crossover pattern has an important practical implication: \emph{mixture rankings established at small scale do not transfer reliably to larger scales}.
In our setting, optimizing the data mix at the \texttt{small} scale (1B$\times$6.25B) would select the Caption-heavy mix and miss the Instruction-heavy configuration that ultimately performs best.
This underscores the need for \textit{scale-aware data curation} that validates mixture choices across multiple points on the scaling ladder, rather than at a single small-scale proxy~\citep{nezhurina2025scaling,nguyen2025mixturevitae,mizrahi2025language,goyal2024scaling,shukor2025scaling,mohri2026bitter,gao2021empirical}.

\textbf{Repeatability of instruction-tuning data.}\label{sec:it-repeats} Given our previous finding that Instruction-heavy mixes scale better,
a natural concern about \emph{scalability} arises: instruction-tuning datasets are typically orders of magnitude smaller in size than web-crawled image-caption pairs.
A 70\% allocation might require extreme data repetitions to fill the token budget, a known cause of performance degradation~\citep{muennighoff2023scaling,hernandez2022scaling,fang2025datasets,carlini2022quantifying,liu2026infolaw}.
We test this effect by holding all non-instruction data sources fixed and randomly subsampling instruction-tuning data to induce up to 2$\times$, 4$\times$, and 8$\times$ repetitions at the \texttt{medium} scale of our benchmark. 
From~\cref{tab:it-repeats}, we find that performance degrades gracefully: each doubling of repetition factor costs roughly 0.5--1.0\% in performance.
Notably, the Instruction-heavy mix with 2$\times$ repetitions (50.2\%) still matches the Caption-heavy mix with fully unique data (50.3\%), and at 4$\times$ repetitions it remains above the base mix (49.8\% vs 48.8\%).
The mix ultimately degrades at ${\sim}8\times$ repetitions.
This result has a practical takeaway: 
\textit{the benefits of a good mixture outweigh 
the costs of moderate data repetition.}
Our results corroborate similar findings from the language domain regarding the benefits of including instruction-like data during pretraining~\citep{baek2026finetuner,akter2025front,washbourne2026zaya1,zeng2025glm,li2026legalone,feng2026early,allen2023physics}.

\subsection{Control Experiments}\label{sec:controls}
We now verify the generality of our findings before scaling up. 
Specifically, we ask:
\textit{(i)} Does the effectiveness of pretraining data curation hold after supervised fine-tuning (SFT)?, and \textit{(ii)} Are our findings tied to the LM backbone (Qwen2.5-Base) used for initialization? We provide answers next.

\begin{figure}[t!]
    \centering
    \includegraphics[width=\linewidth]{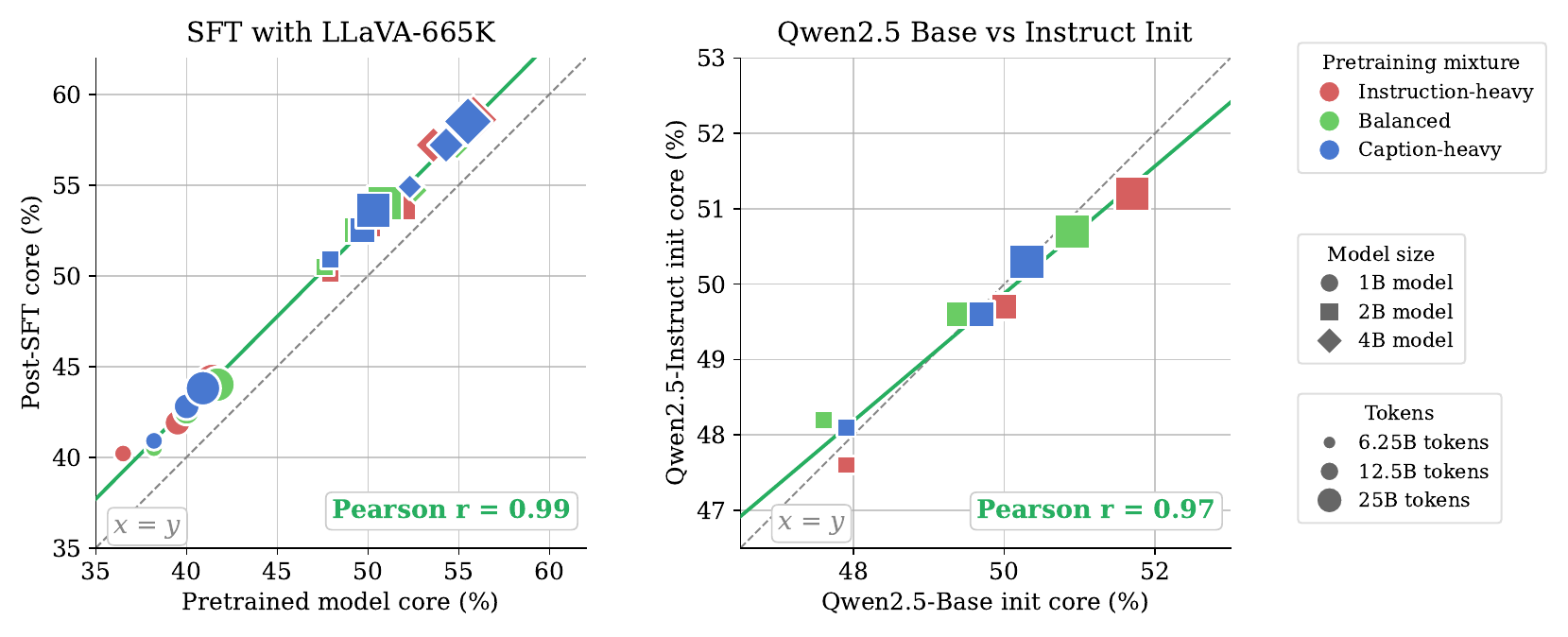}
    \caption{\textbf{Control experiments.}
    \textit{(Left)} Pretraining performance predicts post-SFT performance with near-perfect fidelity.
    \textit{(Right)} Data mixture rankings are preserved when switching the LM backbone from Qwen2.5-Base to Qwen2.5-Instruct, verifying robustness of our results to choice of backbone.}
    \label{fig:sft-correlation}
    \vspace{-0.5em}
\end{figure}

\textbf{Pretraining results transfer reliably post-SFT.} A common concern with pretraining-only evaluations is the worry that SFT will overwrite differences induced by pretraining data choices~\citep{kumar2022fine}. 
In particular, given the findings in \cref{sec:mixing}, it is natural to hypothesize that SFT (which also uses instruction-tuning data by definition) may interfere with or diminish the effect of using an Instruction-heavy pretraining mixture. %
We study this by SFT-ing all 27 pretrained checkpoints from our previous scaling grid (3 mixes $\times$ 3 model sizes $\times$ 3 token budgets) using two different SFT datasets: LLaVA-665K~\citep{liu2023visual} and Mammoth-VL-12M~\citep{guo2025mammoth}, for a total of 54 SFT runs.
We set the total SFT tokens to $0.29\times$ the pretraining tokens by estimating InternVL3's~\citep{zhu2025internvl3} SFT-to-pretraining token ratio.
 \cref{fig:sft-correlation} (left) shows the results with LLaVA-665K (refer to \textcolor{black}{\cref{app:sft}} for identical results with Mammoth-VL-12M).
 We observe that pretraining and post-SFT scores are near-perfectly correlated (Pearson ${r}{=}{0.99}$; Spearman ${\rho}{=}{0.99}$), 
and \textit{the pretraining ordering is preserved across all runs}.

\textbf{Our findings are robust to LM initialization.}
So far, we've used Qwen2.5-Base as the language model backbone. 
To verify that our findings are not specific to this particular choice, we repeat the full 2B-model sweep (3 mixes $\times$ 3 token budgets) with Qwen2.5-Instruct-2B as the LM.
This allows us to verify whether instruction-heavy mixes are better even when the LM has been ``unimodally'' instruction-tuned already. 
As shown in \cref{fig:sft-correlation} (right), it produces nearly identical mixture rankings to Qwen2.5-Base (Pearson $r{=}0.97$), especially at larger token budgets (denoted by larger markers). 
These results provide some evidence of generality of our results---particularly, the advantage of instruction-heavy mixes after training at scale may be agnostic to the LM initialization choice.

\begin{table}[t]
    \centering
    \caption{\textbf{DCVLM results across scales.} We compare our \textsc{\dcvlm-Baseline} against the best open pretraining datasets (\textsc{LLaVA-OneVision-1.5}, \textsc{Nemotron-VL-2}, \textsc{FineVision}) on the core evaluation suite, across all four scales. 
    We also report pretrained InternVL models for reference. Benchmark categories: \textbf{Gen} = General Understanding -- \textbf{Know} = Knowledge-Centric -- \textbf{OCR} = OCR \& Charts -- \textbf{Vision} = Vision-Centric -- \textbf{MTL} = Multilingual -- \textbf{Text} = Text-Only Understanding.}
    \label{tab:main-results}
    \resizebox{\textwidth}{!}{
    \begin{tabular}{lccccccccc}
        \toprule
        \textbf{Method} & \textbf{Model} & \textbf{Tokens} & \rotatebox{0}{\textbf{Gen}} & \rotatebox{0}{\textbf{Know}} & \rotatebox{0}{\textbf{OCR}} & \rotatebox{0}{\textbf{Vision}} & \rotatebox{0}{\textbf{MTL}} & \rotatebox{0}{\textbf{Text}} & \textbf{Core Avg} \\
        \midrule
        \rowcolor{gray!8} \multicolumn{10}{l}{\texttt{small} scale} \\
        LLaVA-OneVision-1.5          & 1B & 6.25B & 22.4 & 34.8 & 8.2 & 27.8 & 13.5 & 6.9 & 17.6 \\
        Nemotron-VL-2                & 1B & 6.25B & 20.0 & 39.7 & 7.9 & 33.5 & 16.1 & 20.7 & 22.1 \\
        FineVision                   & 1B & 6.25B & 40.1 & 45.6 & 35.0 & 41.0 & 28.2 & 28.9 & 36.2 \\
        \dcvlm-Baseline (ours) & 1B & 6.25B & 40.5 & 43.6 & 33.0 & 39.1 & 25.4 & 34.7 & \textbf{36.5} \\

        \midrule
        \rowcolor{gray!8} \multicolumn{10}{l}{\texttt{medium} scale} \\
        LLaVA-OneVision-1.5          & 2B & 25B   & 33.3 & 43.0 & 21.0 & 30.4 & 21.5 & 16.0 & 26.5 \\
        Nemotron-VL-2                & 2B & 25B   & 48.6 & 54.6 & 19.9 & 41.1 & 36.7 & 28.6 & 37.0 \\
        FineVision                   & 2B & 25B   & 55.3 & 62.6 & 51.9 & 45.8 & 40.6 & 46.3 & 50.6 \\
        \dcvlm-Baseline (ours)       & 2B & 25B   & 62.3 & 60.5 & 45.8 & 47.3 & 44.2 & 47.8 & \textbf{51.7} \\
        \midrule
        \rowcolor{gray!8} \multicolumn{10}{l}{\texttt{large} scale} \\
        Nemotron-VL-2                & 4B & 100B  & 31.5 & 53.8 & 23.6 & 38.6 & 27.5 & 36.4 & 34.7 \\
        FineVision                   & 4B & 100B  & 59.0 & 70.7 & 58.9 & 39.1 & 45.1 & 51.2 & 54.2 \\
        \dcvlm-Baseline (ours)       & 4B & 100B  & 68.4 & 67.6 & 54.1 & 57.2 & 50.9 & 53.8 & \textbf{58.9} \\
        \midrule
        \rowcolor{gray!8} \multicolumn{10}{l}{\texttt{x-large} scale} \\
        FineVision                   & 8B & 200B  &  63.5 &  72.8 &  57.5 &  49.6 &  48.4 &  55.7 &  58.2 \\
        \dcvlm-Baseline (ours)       & 8B & 200B  & 73.0 & 73.0 & 53.4  & 63.5 & 56.1 & 61.1 & \textbf{63.6} \\
        \midrule
        \rowcolor{gray!8} \multicolumn{10}{l}{{\textit{open-weight, closed-data reference models}}} \\
        InternVL-2.5-8B              & 8B & $\sim$98B    & 68.2 & 70.7 & 52.2 & 52.3 & 45.6 & 63.3 & 60.0 \\
        InternVL-3-8B                & 8B & $\sim$200B    & 78.8 & 81.1 & 64.1 & 65.1 & 60.7 & 61.4 & 68.5 \\
        InternVL-3.5-8B              & 8B &$\sim$250B    & 77.2 & 80.2 & 63.6 & 63.7 & 59.7 & 63.1 & 68.1 \\
        \bottomrule
    \end{tabular}
    }
\vspace{-0.25em}
\end{table}

\section{Scaling Up Our Findings}\label{sec:results}
Building on our findings, we propose \textsc{\dcvlm-Baseline}---a simple data recipe that forgoes filtering and instead focuses on carefully tuned data mixtures. 
\textcolor{black}{Accordingly, we use the Instruction-heavy mix: 10\% Image-Caption data, 5\% Multimodal Documents, 15\% Text-Only data, and 70\% Instruction-Tuning data (see the full mix in~\cref{fig:hero}), which was found to be optimal for \texttt{medium} and \texttt{large} scales (\cref{sec:mixing}). For simplicity, we use this as well for the \texttt{small} scale---however, we reiterate the \textit{scale-aware} nature of data curation and the fact that the ``optimal'' mixture for the \texttt{small} scale was in fact the Caption-heavy one (\cref{fig:mixing-panels}).}
For each data type, we fill the token budget by drawing samples from its constituent sources via simple length-proportional sampling.

We compare \textsc{\dcvlm-Baseline} to three open VLM pretraining datasets: \ding{172} LLaVA-OneVision-1.5-Midtraining-85M, used to pretrain the \textsc{LLaVA-OneVision-1.5} family~\cite{an2025llava}; \ding{173} the public data released with \textsc{Nemotron-VL-2}~\cite{deshmukh2025nvidia}; and \ding{174} \textsc{FineVision}~\cite{wiedmann2025finevision}, the prior largest effort to unify existing sources into a single open dataset. As upper-bound reference points, we also report results from the pretrained InternVL 8B models (\textsc{InternVL-2.5}, \textsc{InternVL-3}, \textsc{InternVL-3.5}).

We train models on \textsc{\dcvlm-Baseline} and the \textsc{FineVision} baseline at all compute scales of our benchmark (\texttt{small}, \texttt{medium}, \texttt{large}, and \texttt{x-large}). For the other baselines (\textsc{LLaVA-OneVision-1.5} and \textsc{Nemotron-VL-2}), we observe that performance is quite poor at smaller scales and hence did not use those datasets for training at the larger scales. We report all results in \cref{tab:main-results}.
These results, across all scales, confirm that \textbf{\textsc{\dcvlm-Baseline} outperforms open pretraining dataset for VLMs.}
Specifically, compared with \textsc{FineVision} (the previous best open pretraining dataset), we observe consistent gains that \emph{increase with scale}: \textsc{\dcvlm-Baseline} achieves progressive gains of $+0.3$pp (\texttt{small}), $+1.1$pp (\texttt{medium}), $+4.7$pp (\texttt{large}), and $+5.4$pp (\texttt{x-large}) on our 33-task Core evaluations.
Remarkably, a 4B model trained for 100B tokens on \textsc{\dcvlm-Baseline} (our \texttt{large} scale) \emph{outperforms} an 8B model trained for 200B tokens on \textsc{FineVision} (\texttt{x-large}).

\textbf{52-task Extended results.} These trends are further confirmed by our 52-task Extended evaluation suite (\cref{app:extended-evals}), where a \dcvlm-Baseline-trained model at the \texttt{x-large} scale, scores $60.5\%$ vs. $56.6\%$ for the corresponding \texttt{x-large} scale \textsc{FineVision}-trained model (an absolute improvement of $+3.9$pp). 
In fact, with a score of $56.0$, the \dcvlm-Baseline-trained model at the \texttt{large} scale nearly achieves the same performance as the FineVision \texttt{x-large} model.

\section{Conclusion}\label{sec:conclusion}

We introduced \dcvlm, a systematic benchmark for studying data curation strategies for VLM pretraining.
Through extensive experimentation across a principled scaling ladder, we established two central findings:
(i) individual quality filters provide negligible benefits when the source pools are pre-filtered, which is typical for VLMs, and
(ii) data mixture optimization (specifically, instruction-heavy mixtures) is the most effective curation lever, providing gains that scale reliably with model size and compute;
at our largest scale (8B model, 200B tokens), \textsc{\dcvlm-Baseline} outperforms \fv by $+5.4$pp on our comprehensive 33-benchmark core set.
We release the full data pool (160 datasets), evaluation suite (52 benchmarks in total), model checkpoints at 4 scales, and all experimental infrastructure
to serve as a reproducible testbed for future data research.

\section*{Acknowledgements}
The authors would like to thank (in no particular order): Jeffrey Li, Etash Guha, Alex Fang, Pratyush Maini, Hritik Bansal, Moreno D'Incá, Songlong Xing, Olivier Henaff, Matthew Leavitt, Siddharth Joshi, Wieland Brendel, Samuel Albanie, Francesco Tonini, and Evgenia Rusak, for thoughtful feedback and comments throughout the project.

VU, AH, AG, SD and JS thank the
International Max Planck Research School for Intelligent Systems (IMPRS-IS). VU, SK and SD also thank
the European Laboratory for Learning and Intelligent Systems (ELLIS) PhD program for support.
AH acknowledges funding by the Federal Ministry of Research, Technology and Space (BMFTR), FKZ: 16IS24079A.
SD acknowledges support by the Tübingen AI Center.
AP acknowledges funding by the Federal Ministry of Research, Technology and Space (BMFTR), FKZ: 16IS24085B.
VU was supported by a Google PhD Fellowship in Machine Intelligence.
This work was supported by the German Research Foundation (DFG): SFB 1233, Robust
Vision: Inference Principles and Neural Mechanisms, TP4, project number: 276693517 and the UKRI
grant: Turing AI Fellowship EP/W002981/1. MB is a member of the Machine Learning Cluster of
Excellence, funded by the Deutsche Forschungsgemeinschaft (DFG, German Research Foundation)
under Germany’s Excellence Strategy – EXC number 2064/1 – Project number 390727645.
The authors gratefully acknowledge LAION and the Gauss Centre for Supercomputing e.V. for funding this work
by providing computing time on the JUWELS Booster at Jülich Supercomputing Centre (JSC). AG, MM, ER, JJ and HK receive funding
from the European Union’s Horizon Europe research and innovation program under ELLIOT - Grant Agreement No
101214398.
MN and JJ acknowledge co-funding by EU from Digital Europe Programme under grant no. 101195233 (openEuroLLM), co-funding from EU under Digital Europe Programme under grant no. 101198470 (LLMs4EU) and from EuroHPC Joint Undertaking programme under grant no. 101182737 (MINERVA), funding by the German Federal Ministry of Research, Technology and Space (BMFTR) under grant no. 01IS24085C (OPENHAFM), under the grant 01IS22094B (WestAI -- AI Service Center West), and under the grant 16HPC117K (MINERVA).
SJ is funded by the European Research Council (ERC) under the Starting Grant GraViLa (101117556).
MF acknowledges travel support
from ELIAS (GA no 101120237). MB acknowledge financial support by the Federal Ministry of Education and
Research (BMBF), FKZ: 011524085B and Open Philanthropy Foundation funded by the Good Ventures Foundation.
The authors acknowledge the CINECA award under the ISCRA initiative for the availability of high-performance computing resources and support, and the projects EU Horizon projects ELIAS (No. 101120237) and ELLIOT (No. 101214398). The authors also acknowledge the Leonardo supercomputing hours awarded by through the project EHPC-AIF-2025SC03-174.
TN and SO acknowledge NSF grants 2505865, 2229876, 2229876, and 2502281. 
The entire team wishes to acknowledge the Gauss Centre for Supercomputing e.V. (GCS) for funding this work by providing computing time through the John von Neumann Institute for Computing (NIC) on the supercomputer JUWELS Booster and JUPITER at J"ulich Supercomputing Centre (JSC), storage resources on JUST granted and operated by JSC and supported by Helmholtz Data Federation (HDF), computing time granted by the JARA and JSC on the supercomputer JURECA at JSC, computing time granted on prototype JEDI via JUREAP (JUPITER Early Access Program) grant at JSC and computing time granted via Gauss AI Competition (reformo) on JUPITER through GCS and German Federal Ministry of Research, Technology and Space (BMFTR). 
LAION further acknowledges public storage grant by HuggingFace that allows us to provide convenient access to the output of the open-source research to broad community via HF repository. 
Further thanks go for support provided by supercomputing facilities and their teams, especially to Bjoern Hagemeier and Mathis Bode from Juelich Supercomputer Center (JSC, Germany).
SK is supported by the CS at Max Planck Doctoral Program, VIA Center and Saarland Informatics Campus. The authors acknowledge the GCP Credit Award Program by Google with award GCP444206605 for supporting the project with computational credits on GCP.
MBö is supported by the Swiss National Science Foundation (project number 200021\_204620).

\newpage

\bibliographystyle{abbrvnat}
\bibliography{bibliography}

\newpage

\addcontentsline{toc}{section}{Appendix} %
\part{Appendix}
\parttoc

\appendix
\startcontents[sections]
\printcontents[sections]{l}{1}{\setcounter{tocdepth}{2}}
\clearpage

\section{Contributions}\label{app:contributions}

This was a large collaborative effort, and the work spanned data curation, infrastructure, experimentation, and analysis. Below we summarize the main contribution areas. Within each area, contributors are listed in a random shuffled order. An author may appear under multiple areas.

\smallsec{Project coordination.} Matteo Farina, Vishaal Udandarao, Thao Nguyen, Nikhil Parthasarathy, Ludwig Schmidt

\smallsec{Data pool construction.} Thao Nguyen, Vishaal Udandarao, Matteo Farina, Selim Kuzucu, Andreas Hochlehnert, Adhiraj Ghosh, Mehdi Cherti, Karsten Roth, Joschka Struber, Yuhui Zhang, Sebastian Dziadzio, Elaine Sui, Dhruba Ghosh, Hasan Hammoud, Thomas De Min, Simone Caldarella, Sedrick Keh 

\smallsec{Data filtering.} Vishaal Udandarao, Matteo Farina, Selim Kuzucu, Thao Nguyen, Maximilan Böther, Andreas Hochlehnert, Marianna Nezhurina, Adhiraj Ghosh, Soumya Jahagirdar, Elaine Sui, Jehanzeb Mirza

\smallsec{Train--test decontamination.} Matteo Farina, Vishaal Udandarao, Maximilian Böther, Adhiraj Ghosh, Marianna Nezhurina

\smallsec{Annotation infrastructure.} Maximilian Böther, Matteo Farina, Marianna Nezhurina

\smallsec{Data mixing.} Vishaal Udandarao, Matteo Farina

\smallsec{Training infrastructure and scaling.} Matteo Farina, Marianna Nezhurina, Vishaal Udandarao

\smallsec{Evaluation suite.} Matteo Farina, Sebastian Dziadzio, Karsten Roth

\smallsec{Transfer and controlled experiments.} Vishaal Udandarao, Matteo Farina, Thao Nguyen, Andreas Hochlehnert, Adhiraj Ghosh

\smallsec{Writing.} Vishaal Udandarao, Matteo Farina, Adhiraj Ghosh, Soumya Jahagirdar, Massimiliano Mancini, Nikhil Parthasarathy, Ludwig Schmidt

\smallsec{Advising and supervision.} Nikhil Parthasarathy, Ludwig Schmidt, Massimiliano Mancini, Jenia Jitsev, Alessio Tonioni, Ameya Prabhu, Sewoong Oh, Matthias Bethge, Elisa Ricci, Ana Klimovic, Federico Tombari, Muhammad Ferjad Naeem, Serena Yeung-Levy, Bernt Schiele, Hilde Kuehne

\clearpage

\section{Extended Related Work}\label{app:related-work}

In the main paper, we provided a brief overview of the most relevant recent papers for our work. Here, we provide a deeper dive into these related papers.

\smallsec{Vision-Language Model Training Regimes.}
The development of modern autoregressive VLMs has converged on a modular architecture, consisting of a pretrained vision encoder, a language model backbone, and a lightweight connector between the two. Early methods differed in how this connection was implemented.  Notable works include BLIP-2~\citep{li2023blip} which used a Q-former to compress visual tokens and Flamingo~\citep{alayrac2022flamingo}, which inserted cross-attention layers between frozen vision and language features. The dominant blueprint can be attributed to  LLaVA~\citep{liu2023visual,liu2024improved,liu2024llavanext} which popularized the simpler recipe of ``pretraining" the connector on predominantly image-text pairs ~\citep{an2025llava}, before conducting supervised fine-tuning (SFT) on curated instruction data. 

In contrast, recent works have considerably relaxed these constraints. First, frontier works like InternVL3\cite{zhu2025internvl3}, InternVL3.5\cite{wang2025internvl3} and LLaVA-OneVision-1.5~\citep{an2025llava} fine-tune all model parameters from scratch. The relationship between these training choices and model scale, image resolution and data composition have been studied too~\citep{mckinzie2024mm1,zhang2024mm1}. Concurrently, the focus has also shifted to making the data composition more heterogeneous while training VLMs, moving away from an over-reliance on image-text pairs. Idefics~\citep{laurenccon2023obelics,laurenccon2024idefics2} trains on interleaved image-text sequences, UReader uses multimodal documents~\citep{ye2023ureader}, PaliGemma~\citep{beyer2024paligemma} combines image-text pairs with generated VQA, multi-object detection and OCR,  Cambrian~\citep{tong2024cambrian} includes text-only corpora for preserving language capabilities, etc. Most notably, ~\citep{zhu2025internvl3,wiedmann2025finevision} advocate for using instruction-tuning data during pretraining itself. However, the precise mixture ratios, filtering criteria, and formatting choices that drive these systems remain proprietary or only coarsely documented, motivating our systematic benchmark.

\smallsec{Benchmarking Data Curation.}
Controlled data-curation benchmarks keep model architectures and training pipelines fixed and only vary the data distribution fed to the model.  DataPerf~\citep{mazumder2023dataperf} established this paradigm, while
DataComp~\citep{gadre2023datacomp} brought it to CLIP pretraining, enabling principled comparison of curation strategies at scale. DCLM~\citep{li2024datacomp} extended this to language model pretraining, demonstrating that a simple fasttext classifier trained on high-quality text can substantially improve downstream performance.
FineWeb~\citep{penedo2024fineweb} and its educational-quality variant, FineWeb-Edu, showed similar gains through quality-based filtering of Common Crawl. In general, quality-based data filtering has shown strong results for text~\citep{penedo2024fineweb,li2024datacomp} and image-text pairs~\citep{gadre2023datacomp,wang2025unifilter}.

Existing data curation methods can be categorized into two groups:   filtering and mixing. Common filtering approaches include CLIP-score filtering~\citep{hessel2021clipscore}, image quality assessment~\citep{agnolucci2024arniqa}, text quality classifiers~\citep{su2025nemotron,deshmukh2025nvidia}, and learned multimodal quality estimators~\citep{wang2025unifilter}. These filtering methods have proved to be quite effective in driving downstream performance for single data-type (image-text or text-only) datasets, training better CLIP models being an example \citep{gadre2023datacomp}. Using pretrained data-selector models or multimodal quality scores 
are more recent approaches to quantify whether a data sample is likely to improve pretraining~\citep{mahmoud2024sieve,wang2024cliploss,wang2024variance,chen2024internvl,kim2024hype}.

Beyond model-based filters, offline curation also comprises deduplication, recaptioning and concept-aware selection. Deduplication ranges from general pruning~\citep{sorscher2022beyond} to semantic deduplication~\citep{abbas2023semdedup,abbas2024effective}. Recaptioning methods aims to replace weak web-scale alt-text with synthetic or fused captions using VLMs or caption augmentation ~\citep{li2024if,yu2024capsfusion,fan2024improving,nguyen2024improving,nguyen2024multilingual,ghosh2025concept}. Concept-aware methods control the training distribution through concept filtering or balancing  ~\citep{ghosh2025concept,xu2023demystifying,oquab2023dinov2}. Put together, it has specifically been shown that the offline curation of noisy web-scale data results in large pretraining efficiency gains~\citep{jia2021scaling,changpinyo2021conceptual,gadre2023datacomp,fang2023data,vo2024automatic,maini2023t}.

Prior works have also explored data mixing instead of filtering, with standard approaches relying on strategies such as domain weighting~\citep{xie2023doremi,olmo20242}, mixture optimization~\citep{chen2026olmix,berasi2026linear,diao2025nemotron,kang2024autoscale,ye2024data,liu2024regmix} and temperature-scaled sampling~\citep{deitke2025molmo,clark2026molmo2}. %
Despite the efforts in releasing curated datasets (e.g., \fv~\citep{wiedmann2025finevision}), there exists no systematic study ablating filtering or mixing strategies in the VLM setting.  Our work fills this gap by providing a controlled testbed for multimodal data curation, providing the first \emph{scale-aware} study of data-type mixing for VLMs.

\smallsec{Train-Test (De)contamination.}
Train-test overlap (contamination) is a well-documented concern, especially in language model evaluation, where several works have demonstrated how benchmark scores can be inflated when test-set or their near-duplicate samples appear in the pretraining data pool ~\citep{jiang2024investigating,magar2022data,sainz2024data,bordt2024much,schaeffer2026quantifying}.
This is a problem for VLM training as well as the contamination can stem from many sources: text, duplicate or near-duplicate images or documents, etc. To mitigate such concerns, and also to ensure that models do not degrade to rote memorization of the training sets, several works conduct robust decontamination procedures on their training sets~\citep{beyer2024paligemma,zhai2022scaling,oquab2023dinov2,trinh2018simple,gao2020pile,mizrahi2025language,allal2025smollm2}, i.e, they attempt to remove training examples too similar (or, at worst, identical) to evaluation examples. 
Some canonical methods include embedding-based similarity search~\citep{oquab2023dinov2,wiedmann2025finevision}, MinHash signatures for approximate text-matching ~\citep{penedo2024fineweb,li2024datacomp,broder1997resemblance} and direct string-search using suffix arrays~\citep{lee2022deduplicating}. 
In our work, we employ two-way decontamination: a form of embedding-based decontamination for multimodal samples and MinHash signatures for text-only samples. 

\smallsec{Scaling Laws and Scale-Aware Curation.}
An important consequence of scaling-law studies is that a data curation strategy chosen at one scale may not remain optimal at others. A growing body of evidence suggests that the effectiveness of these filters is \emph{scale-dependent}: \citet{goyal2024scaling} and \citet{mizrahi2025language} show that optimal filtering aggressiveness decreases with compute budget. Our work successfully extends this finding to the multimodal setting.
showing that \emph{at sufficient scale and with optimized mixtures, no individual quality filter provides reliable and consistent gains}.

\clearpage

\section{Model Architecture Details}\label{app:architecture}

All our experiments use a single VLM architecture template, parameterised across the four scales of the scaling ladder (\cref{tab:scales}). The template follows the InternVL-3~\citep{zhu2025internvl3} family: a vision encoder $\rightarrow$ a randomly-initialised MLP projector $\rightarrow$ an autoregressive language-model backbone, with all three components trained jointly from the start (single-stage pretraining, no frozen components). Across our four scales, only the language-model backbone changes; the vision encoder and the projector recipe are held fixed. We document each component in turn.

\subsection{Vision Encoder}\label{app:arch-vision}

We use \textbf{InternViT-300M-448px-V2.5}~\citep{chen2024expanding} for all experiments. It is a Vision Transformer~\citep{dosovitskiy2020image} with the modifications introduced in the InternVL series~\citep{chen2024internvl,chen2024expanding,zhu2025internvl3}, kept identical across our four scales. \cref{tab:arch-vision} reports its key structural choices.

\begin{table}[!h]
\centering
\caption{\textbf{Vision encoder architecture (InternViT-300M-448px-V2.5).} The same vision encoder is used across all four scales. 
The encoder is fully unfrozen and updated jointly with the projector and the LM backbone.}
\label{tab:arch-vision}
\small
\begin{tabular}{ll}
\toprule
\textbf{Component} & \textbf{Value} \\
\midrule
\rowcolor{textcolor}\multicolumn{2}{l}{\textit{Input}} \\
Image resolution & $448 \times 448$ (one tile; dynamic high-res tiling, see \cref{sec:methodology}) \\
Patch size & $14 \times 14$ \\
Tokens per tile (pre-pixel-shuffle) & $32 \times 32 = 1024$ \\
Tokens per tile (post-pixel-shuffle, fed to LM) & $16 \times 16 = 256$ \\
Channels & 3 (RGB) \\
\midrule
\rowcolor{textcolor}\multicolumn{2}{l}{\textit{Transformer trunk}} \\
Depth (layers) & 24 \\
Hidden size & 1024 \\
Attention heads & 16 \\
Head dim & 64 \\
FFN intermediate size & 4096 \\
FFN activation & GELU \\
FFN style & 2-layer MLP (Linear $\rightarrow$ GELU $\rightarrow$ Linear) \\
Attention style & Standard multi-head; QKV bias \cmark, O-proj bias \xmark \\
QK normalisation & \xmark \\
Normalisation & Pre-LayerNorm, $\varepsilon = 10^{-6}$ \\
Positional embeddings & Learned absolute (interpolated to 448px) \\
Flash attention & \cmark (FA-2~\citep{dao2023flashattention}) \\
\midrule
\rowcolor{textcolor}\multicolumn{2}{l}{\textit{Total}} \\
Parameters & $\sim$304M \\
\bottomrule
\end{tabular}
\end{table}

The tile-based tokenisation produces $32 \times 32 = 1024$ patches per $448 \times 448$ tile. After the projector's pixel-shuffle reduction (\cref{app:arch-projector}), this becomes $16 \times 16 = 256$ visual tokens per tile ($4{\times}$ reduction) that are fed to the language model. 
Multiple tiles and a thumbnail image are concatenated into the LM input, following the dynamic high-resolution scheme of InternVL-2.5~\citep{chen2024expanding}.

\subsection{Projector}\label{app:arch-projector}

The vision encoder and the language model are bridged by a small randomly-initialised MLP-style projector (often called the ``connector'' in VLM literature~\citep{mckinzie2024mm1,liu2023visual}). It is the only module that is randomly initialised at the start of training, everything else is loaded from pretrained checkpoints. \cref{tab:arch-projector} gives the exact structure.
\begin{table}[!h]
\centering
\caption{\textbf{Projector architecture.} The projector is a fixed-depth, fixed-activation 2-layer MLP whose width is the only quantity that varies across scales (it tracks $D_{\mathrm{LM}}$, the language-model hidden size).}
\label{tab:arch-projector}
\small
\begin{tabular}{ll}
\toprule
\textbf{Stage} & \textbf{Operation} \\
\midrule
0. Pre-projection & Pixel shuffle, factor $0.5$: $1024$ tokens of dim $D_{V} \rightarrow 256$ tokens of dim $4 D_{V}$ \\
1. Norm & LayerNorm$(4 D_{V})$, $\varepsilon = 10^{-5}$ \\
2. Linear-1 & Linear$(4 D_{V} \rightarrow D_{\mathrm{LM}})$, bias \cmark \\
3. Activation & GELU \\
4. Linear-2 & Linear$(D_{\mathrm{LM}} \rightarrow D_{\mathrm{LM}})$, bias \cmark \\
\midrule
\rowcolor{textcolor}\multicolumn{2}{l}{\textit{Per-scale projector parameter count}} \\
Small (1B; $D_{\mathrm{LM}}{=}896$)   & $\sim$4.5M \\
Medium (2B; $D_{\mathrm{LM}}{=}1536$) & $\sim$8.6M \\
Large (4B; $D_{\mathrm{LM}}{=}2048$)  & $\sim$12.6M \\
X-Large (8B; $D_{\mathrm{LM}}{=}3584$) & $\sim$27.5M \\
\bottomrule
\end{tabular}
\end{table}
Following InternVL-2.5/3, we keep depth and activation fixed across scales; only the projector's hidden width tracks the LM. 

\subsection{Language Model Backbones}\label{app:arch-lm}

For the four points on our scaling ladder we use four different sizes from the \textbf{Qwen2.5} family~\citep{qwen25}: $0.5$B, $1.5$B, $3$B, and $7$B parameters. All four share the Qwen2 transformer architecture~\citep{yang2024qwen2technicalreport}---SwiGLU FFN~\citep{shazeer2020glu}, RMSNorm~\citep{zhang2019root}, RoPE position embeddings~\citep{su2024roformer}, grouped-query attention (GQA)~\citep{ainslie2023gqa}, no QK-normalisation. 
They differ only in their depth/width/head budget and in two minor configuration knobs (max position length and embedding tying), summarised in \cref{tab:arch-lm}. 
Unless specified, we always initialise from the \textit{Base} (non-Instruct) checkpoints by default. 

\begin{table}[!h]
\centering
\caption{\textbf{Language-model backbone architecture across the four scaling-ladder scales.} All are Qwen2.5 base checkpoints~\citep{qwen25}. ``Head dim'' is hidden size divided by query head count. Vocabularies are the standard Qwen2.5 tokenizer.}
\label{tab:arch-lm}
\small
\setlength{\tabcolsep}{4pt}
\begin{tabular}{lcccc}
\toprule
\textbf{Scale} & \textbf{Small (1B)} & \textbf{Medium (2B)} & \textbf{Large (4B)} & \textbf{X-Large (8B)} \\
\textbf{Qwen2.5 size} & 0.5B & 1.5B & 3B & 7B \\
\midrule
\rowcolor{textcolor}\multicolumn{5}{l}{\textit{Shared structural choices (all scales)}} \\
Architecture family & \multicolumn{4}{c}{Qwen2 transformer~\citep{yang2024qwen2technicalreport}} \\
Normalisation & \multicolumn{4}{c}{Pre-RMSNorm, $\varepsilon = 10^{-6}$} \\
QK-normalisation & \multicolumn{4}{c}{\xmark} \\
FFN style & \multicolumn{4}{c}{SwiGLU (gated MLP, SiLU activation)} \\
Attention style & \multicolumn{4}{c}{Grouped-query attention (GQA), QKV bias \cmark, O-proj bias \xmark} \\
Positional embeddings & \multicolumn{4}{c}{RoPE, base $\theta = 10^{6}$, no scaling} \\
\midrule
\rowcolor{textcolor}\multicolumn{5}{l}{\textit{Per-scale dimensions}} \\
Layers & 24 & 28 & 36 & 28 \\
Hidden size & 896 & 1536 & 2048 & 3584 \\
Query heads & 14 & 12 & 16 & 28 \\
KV heads (GQA) & 2 & 2 & 2 & 4 \\
Head dim & 64 & 128 & 128 & 128 \\
FFN intermediate & 4{,}864 & 8{,}960 & 11{,}008 & 18{,}944 \\
\midrule
\rowcolor{textcolor}\multicolumn{5}{l}{\textit{Per-scale config knobs}} \\
Max position embedding & 32{,}768 & 131{,}072 & 32{,}768 & 131{,}072 \\
Tied input/output embed. & \cmark & \cmark & \cmark & \xmark \\
Vocabulary size & 151{,}936 & 151{,}936 & 151{,}936 & 152{,}064 \\
LM parameters & $\sim$494M & $\sim$1.54B & $\sim$3.09B & $\sim$7.62B \\
\bottomrule
\end{tabular}
\end{table}

A few cross-scale observations are worth flagging because they surface in our scaling experiments:

\begin{itemize}[leftmargin=*,itemsep=2pt,topsep=2pt]
\item \textbf{Head dimension is not constant.} The $0.5$B model uses $64$-dim heads, while $1.5$B/$3$B/$7$B all use $128$-dim heads. Practitioners scaling pretraining recipes should be aware that the small scale therefore has a slightly different attention behaviour than the rest of the ladder, even though the rest of the architecture is uniform.
\item \textbf{KV-head count is heavily compressed.} GQA ratios are $14{:}2$, $12{:}2$, $16{:}2$, and $28{:}4$ from Small to X-Large---all sub-7B models share the same minimal $2$ KV heads.
\item \textbf{Tied embeddings only at sub-7B.} The $7$B model is the only scale where input/output embeddings are not tied. This costs $\sim$50M extra LM parameters at the X-Large scale.
\item \textbf{Layer count is non-monotonic.} The $3$B model is deeper (36 layers) than the $7$B model (28 layers); $7$B grows primarily by widening (hidden size $2048 \rightarrow 3584$) rather than deepening.
\end{itemize}

These idiosyncrasies are inherited from the Qwen2.5 release and we deliberately do \emph{not} smooth them out, since our purpose is to produce a benchmark whose scaling axis can be reproduced from publicly-released checkpoints rather than to study clean architectural scaling.

\subsection{End-to-end Parameter Accounting}\label{app:arch-totals}

\cref{tab:arch-totals} sums the three components per scale, giving the total trainable-parameter count behind the ``1B / 2B / 4B / 8B'' labels used throughout the paper. The vision encoder and projector together contribute roughly $5$--$60$\% of parameters at the small scale and $\sim$5\% at the X-Large scale.

\begin{table}[!h]
\centering
\caption{\textbf{Total parameter count per scale.} Vision encoder (InternViT-300M-448px-V2.5) is fixed; LM backbone is the corresponding Qwen2.5 size.}
\label{tab:arch-totals}
\small
\begin{tabular}{lcccc}
\toprule
& \textbf{Small} & \textbf{Medium} & \textbf{Large} & \textbf{X-Large} \\
\midrule
Vision encoder & 304M & 304M & 304M & 304M \\
Projector & 4.5M & 8.6M & 12.6M & 27.5M \\
LM backbone & 494M & 1.54B & 3.09B & 7.62B \\
\midrule
\textbf{Total trainable} & \textbf{$\sim$0.80B} & \textbf{$\sim$1.85B} & \textbf{$\sim$3.40B} & \textbf{$\sim$7.95B} \\
Paper label & 1B & 2B & 4B & 8B \\
\bottomrule
\end{tabular}
\end{table}

All parameters are trained jointly---there is no frozen-encoder pretraining stage, no LoRA~\citep{hu2022lora} adapter, and no separate connector-warmup phase. We refer the reader to \cref{tab:hparams} for the optimizer, schedule, and packing settings used during this joint training.

\clearpage

\section{Training and hyperparameter details}\label{app:hparams}

We provide the exact hyperparameters we use for all our training runs in~\cref{tab:hparams}. For the most part, these were derived from the InternVL-2.5~\citep{chen2024expanding} and InternVL-3~\citep{zhu2025internvl3} configurations. However, we did run a small learning rate (LR) sweep of our own to confirm that these were indeed the best performing on a subset of downstream evaluations. 

\begin{table}[!h]
    \centering
    \caption{\textbf{Pretraining hyperparameters.} All values are fixed across scales unless noted in \cref{tab:scales}.} 
    \label{tab:hparams}
    \small
    \begin{tabular}{ll}
        \toprule
        \textbf{Hyperparameter} & \textbf{Value} \\
        \midrule
        \rowcolor{textcolor}\multicolumn{2}{l}{\textit{Optimization}} \\
        Optimizer & AdamW~\citep{loshchilov2017decoupled} ($\beta_{1}{=}{0.9}$, $\beta_{2}{=}{0.999}$, ${\epsilon}{=}{1}{e}{-}{8}$) \\
        Learning rate (pretraining) & $2 \times 10^{-5}$ \\
        LR scheduler & Cosine decay~\citep{loshchilov2016sgdr} \\
        Warmup ratio & 0.03 \\
        Weight decay & 0.01 \\
        Precision & BF16~\citep{dean2012large} \\
        Global batch size & 1024 \\
        Per-device batch size & 1 \\
        Gradient checkpointing & \cmark \\
        Parallelism & DeepSpeed ZeRO-1~\citep{rajbhandari2020zero} \\
        \midrule
        \rowcolor{textcolor}\multicolumn{2}{l}{\textit{Sequence packing}} \\
        Max sequence length & 8192 tokens \\
        Max packed tokens & 8192 tokens \\
        Max packed images & 24 \\
        Sampling & With replacement \\
        \midrule
        \rowcolor{textcolor}\multicolumn{2}{l}{\textit{Loss}} \\
        Loss reduction & Square-averaging~\citep{chen2024expanding} \\
        \midrule
        \rowcolor{textcolor}\multicolumn{2}{l}{\textit{Architecture}} \\
        Vision encoder & InternViT-300M-448px-V2.5 \\
        Image resolution & $448 \times 448$ (dynamic tiling) \\
        Pixel shuffle & down-sample ratio 0.5 \\
        Include thumbnail image & \cmark \\
        Vision layer for features & last \\
        Drop path rate & 0.0 \\
        Connector & 2-layer MLP \\
        \bottomrule
    \end{tabular}
\end{table}

\subsection{Learning Rate Selection}\label{app:lr-sweep}

To ensure that the InternVL LR configurations were optimal, we conducted a small LR-sweep ourselves\footnote{We did this LR-sweep quite early on in the project when we had a slightly different experimental setup: (i) we were using the Qwen-Instruct backbones instead of the Qwen-Base LM, and (ii) we had not yet finalized the validation or core evaluation sets, and hence used 12 randomly selected benchmarks for tracking our average-metric.}. We select the learning rate by sweeping five values ($2 \times 10^{-4}$, $4 \times 10^{-5}$, $2 \times 10^{-5}$, $8.91 \times 10^{-6}$, $2 \times 10^{-6}$) at each model scale using 10B training tokens with the base mixture.
All other hyperparameters are held fixed (global batch size 1024, cosine schedule, 3\% warmup) according to those specified in~\cref{tab:hparams}.
As shown in \cref{tab:lr-sweep}, $\text{lr} = 2 \times 10^{-5}$ achieves the best or second-best performance at every model scale and LM backbone setting.
Learning rates above $4 \times 10^{-5}$ cause training instability---particularly at the 1B scale, where $\text{lr} = 2 \times 10^{-4}$ collapses to near-chance performance. This behaviour has also been observed in prior works~\citep{wortsman2023small,zhang2026empirical,roth2024practitioner}.
Learning rates below $10^{-5}$ underfit, with the gap widening at larger model sizes.
We therefore adopt $\text{lr} = 2 \times 10^{-5}$ for all our experiments. This hence also corroborates the LR order-of-magnitude used in InternVL-2.5 and InternVL-3.

\begin{table}[!h]
    \centering
    \caption{\textbf{Learning rate sweep across model scales.} All runs use 10B training tokens with the base mixture. Bold indicates the best average per model-size and LLM-backbone group. $\text{lr} = 2 \times 10^{-5}$ is consistently optimal across scales and backbones.}
    \label{tab:lr-sweep}
    \resizebox{\textwidth}{!}{
    \begin{tabular}{llccccccccccccc}
        \toprule
        \textbf{LR} & \textbf{LLM} & \rotatebox{70}{\textbf{MMMU}} & \rotatebox{70}{\textbf{3DSRBench}} & \rotatebox{70}{\textbf{AI2D}} & \rotatebox{70}{\textbf{BLINK}} & \rotatebox{70}{\textbf{COCO}} & \rotatebox{70}{\textbf{Hall.Bench}} & \rotatebox{70}{\textbf{MMB-CN}} & \rotatebox{70}{\textbf{MMB-EN}} & \rotatebox{70}{\textbf{MMStar}} & \rotatebox{70}{\textbf{Mantis}} & \rotatebox{70}{\textbf{TextVQA}} & \rotatebox{70}{\textbf{SEED}} & \rotatebox{70}{\textbf{Average}} \\
        \midrule
        \rowcolor{gray!8} \multicolumn{15}{l}{\textit{1B model}} \\
        $2 \times 10^{-4}$    & Qwen-Inst.  & 21.0 & 44.1 & 24.6 & 38.6 & 13.4 & 29.9 &  2.2 &  1.1 & 25.2 & 30.4 & 42.1 & 26.0 & 24.9 \\
        $4 \times 10^{-5}$    & Qwen-Inst.  & 30.2 & 45.3 & 35.4 & 37.6 & 15.1 & 28.4 & 35.6 & 42.4 & 33.5 & 35.0 & 49.6 & 44.6 & 36.0 \\
        $2 \times 10^{-5}$    & Qwen-Inst.  & 30.2 & 45.3 & 36.7 & 37.8 & 15.2 & 24.7 & 43.1 & 42.9 & 34.7 & 35.0 & 44.5 & 46.0 & \textbf{36.3} \\
        $8.91 \times 10^{-6}$ & Qwen-Inst.  & 30.4 & 45.0 & 36.3 & 35.9 & 14.8 & 31.9 & 35.5 & 38.7 & 34.4 & 36.9 & 42.2 & 44.6 & 35.5 \\
        $2 \times 10^{-6}$    & Qwen-Inst.  & 30.8 & 45.0 & 39.6 & 36.4 & 12.7 & 26.8 & 32.0 & 35.5 & 34.1 & 40.6 & 13.4 & 43.2 & 32.5 \\
        \midrule
        \rowcolor{gray!8} \multicolumn{15}{l}{\textit{2B model}} \\
        \cdashline{1-15}
        $4 \times 10^{-5}$    & Qwen-Inst.  & 38.0 & 45.1 & 53.6 & 39.0 & 16.6 & 37.6 & 57.2 & 57.5 & 35.8 & 45.2 & 53.1 & 59.2 & 44.8 \\
        $2 \times 10^{-5}$    & Qwen-Inst.  & 38.0 & 45.8 & 53.6 & 39.2 & 18.1 & 36.6 & 57.9 & 59.7 & 35.8 & 44.7 & 53.8 & 58.2 & \textbf{45.1} \\
        $8.91 \times 10^{-6}$ & Qwen-Inst.  & 40.2 & 45.1 & 54.7 & 38.8 & 13.7 & 37.3 & 57.5 & 59.3 & 35.5 & 47.0 & 51.7 & 56.4 & 44.8 \\
        \midrule
        \rowcolor{gray!8} \multicolumn{15}{l}{\textit{4B model}} \\
        $2 \times 10^{-4}$    & Qwen-Inst.  & 31.2 & 44.6 & 44.2 & 38.4 & 21.9 & 34.2 & 51.9 & 51.3 & 38.0 & 40.6 & 53.7 & 54.3 & 42.0 \\
        $4 \times 10^{-5}$    & Qwen-Inst.  & 40.3 & 47.4 & 59.3 & 39.0 & 21.1 & 39.7 & 63.3 & 63.1 & 39.1 & 51.6 & 58.8 & 64.4 & 48.9 \\
        $2 \times 10^{-5}$    & Qwen-Inst.  & 42.2 & 46.1 & 61.4 & 39.5 & 20.5 & 37.5 & 63.6 & 64.0 & 40.5 & 54.8 & 59.1 & 63.7 & \textbf{49.4} \\
        $8.91 \times 10^{-6}$ & Qwen-Inst.  & 42.0 & 46.4 & 57.0 & 38.0 & 15.7 & 35.8 & 58.5 & 60.3 & 40.5 & 49.3 & 52.1 & 61.1 & 46.4 \\
        $2 \times 10^{-6}$    & Qwen-Inst.  & 40.6 & 45.4 & 54.8 & 37.4 & 17.0 & 31.3 & 50.3 & 51.0 & 35.2 & 43.3 & 19.9 & 53.4 & 40.0 \\
        \bottomrule
    \end{tabular}
    }
\end{table}
\clearpage

\section{DCVLM Pool Details}\label{app:dataset_pool}

Our DCVLM pool aggregates 160 publicly available datasets across four data types:
\textbf{image-caption pairs} (13 datasets), \textbf{multimodal interleaved documents} (5),
\textbf{text-only} (33), and \textbf{multimodal instruction-tuning} (109,
spanning 8 capability categories following \cite{wiedmann2025finevision}: Captioning \& Knowledge, Chart \& Table,
General QA, Grounding \& Counting, Math, Naive OCR, OCR QA, and Science).
Our full pool contains 3.9B samples and 6.0T
multimodal tokens, averaging 1.5K tokens per sample. All token counts
are measured using the InternVL-2.5~\citep{chen2024expanding} tokenizer
over the full pool.
The complete per-dataset breakdown of our DCVLM pool is given in \cref{tab:dcvlm-datamix-samples} (showing number of \textit{samples} per dataset) and \cref{tab:dcvlm-datamix-tokens} (showing number of multimodal \textit{tokens} per dataset).

\subsection{Pool Composition}\label{app:pool-composition}

\figpoolcomposition

\definecolor{capcolor}{HTML}{BFDEFF}      %
\definecolor{textcolor}{HTML}{FFCBA9}     %
\definecolor{mmdoccolor}{HTML}{96E19B}    %
\definecolor{instructcolor}{HTML}{FFC6C4} %

\begin{table}[!h]
\centering
\caption{\textbf{DCVLM pool per-dataset sample counts.} The mix combines \colorbox{capcolor}{captioning data}, \colorbox{mmdoccolor}{multimodal documents}, \colorbox{instructcolor}{visual instruction-tuning data} (organised by capability), and \colorbox{textcolor}{text-only data}. Across all 160 datasets, our pool contains \textbf{3.9B} samples.}
\label{tab:dcvlm-datamix-samples}
\scriptsize
\setlength{\tabcolsep}{1.2pt}
\renewcommand{\arraystretch}{1.05}
\begin{minipage}[t]{0.245\textwidth}
\centering
\begin{tabular}{lr}
\toprule
\rowcolor{capcolor}\multicolumn{2}{l}{\textit{Captioning}} \\
Dataset & Size \\
\midrule
ReLAION-2B-en~\cite{schuhmann2022laion} & 1.5B \\
DataComp-1B~\cite{gadre2023datacomp} & 1.4B \\
AS-100M~\cite{wang2024allseeing} & 2.8M \\
GRIT (Cap.)~\cite{peng2023kosmos2} & 14.4M \\
InternVL-SA1B~\cite{chen2024internvl} & 11.9M \\
FaceCaption-15M~\cite{dai2024facecaption} & 11.2M \\
PixMo-Cap~\cite{deitke2025molmo} & 575K \\
ShareGPT-4o~\cite{chen2024sharegpt4v} & 56K \\
TextOCR-GPT4V~\cite{carter2024textocrgpt4v} & 25K \\
TextCaps~\cite{sidorov2020textcaps} & 109K \\
COCO (Cap.)~\cite{chen2015cococaptions} & 569K \\
OpenImages (Cap.)~\cite{kuznetsova2020openimages} & 508K \\
SEA-VL~\cite{cahyawijaya2025seavl} & 1.3M \\
\midrule
\textbf{Total} & \textbf{2.9B} \\
\midrule
\rowcolor{mmdoccolor}\multicolumn{2}{l}{\textit{Multimodal Docs}} \\
Dataset & Size \\
\midrule
MINT-HTML~\cite{awadalla2024mint} & 63.0M \\
MINT-PDF~\cite{awadalla2024mint} & 2.6M \\
OmniCC~\cite{li2024omnicorpus} & 78.4M \\
Multimodal Textbook~\cite{zhang2025textbook} & 602K \\
WanJuan~\cite{he2023wanjuan} & 809K \\
\midrule
\textbf{Total} & \textbf{145M} \\
\midrule
\rowcolor{instructcolor}\multicolumn{2}{l}{\textit{Cap. \& Know.}} \\
Dataset & Size \\
\midrule
Art500K~\cite{mao2017art500k} & 470K \\
LLaVA-595K~\cite{liu2024improved} & 595K \\
MMInstruct~\cite{liu2024mminstruct} & 386K \\
ShareGPT4V~\cite{chen2024sharegpt4v} & 1.2M \\
SVIT~\cite{zhao2023svit} & 3.8M \\
\midrule
\textbf{Total} & \textbf{6.5M} \\
\midrule
\rowcolor{instructcolor}\multicolumn{2}{l}{\textit{Chart \& Table}} \\
Dataset & Size \\
\midrule
BigDocsBench~\cite{rodriguez2025bigdocs} & 406K \\
Chart2Text~\cite{kantharaj2022chart2text} & 8.7K \\
ChartGemma~\cite{masry2024chartgemma} & 150K \\
ChartLlama~\cite{han2023chartllama} & 1.1K \\
ChartQA~\cite{masry2022chartqa} & 30K \\
ChartX~\cite{xia2024chartx} & 17K \\
CoSyn-400K~\cite{yang2025scaling} & 404K \\
DocStruct4M~\cite{hu2024docowl15} & 4.7M \\
DVQA~\cite{kafle2018dvqa} & 2.3M \\
FigureQA~\cite{kahou2017figureqa} & 1.3M \\
FinTabNet~\cite{zheng2021globaltable} & 8.4M \\
MMC-Instruct~\cite{liu2024mmc} & 408K \\
PixMo-Docs~\cite{deitke2025molmo} & 252K \\
PlotQA~\cite{methani2020plotqa} & 20.2M \\
PosterSum~\cite{saxena2025postersum} & 10K \\
SBSFigures~\cite{shinoda2024sbsfigures} & 4.2M \\
SciGraphQA~\cite{li2023scigraphqa} & 296K \\
SimChart9K~\cite{xia2023structchart} & 70K \\
SPIQA~\cite{pramanick2024spiqa} & 262K \\
TabMWP~\cite{lu2023tabmwp} & 23K \\
UniChart~\cite{masry2023unichart} & 7.2M \\
VisText~\cite{tang2023vistext} & 9.9K \\
\midrule
\textbf{Total} & \textbf{50.6M} \\
\bottomrule
\end{tabular}
\end{minipage}
\hfill
\begin{minipage}[t]{0.245\textwidth}
\centering
\begin{tabular}{lr}
\toprule
\rowcolor{instructcolor}\multicolumn{2}{l}{\textit{General QA}} \\
Dataset & Size \\
\midrule
AlgoPuzzleVQA~\cite{ghosal2024algopuzzlevqa} & 1.7K \\
ALLaVA~\cite{chen2024allava} & 1.7M \\
A-OKVQA~\cite{schwenk2022aokvqa} & 17K \\
Cambrian-GPT4o~\cite{tong2024cambrian} & 58K \\
EST-VQA~\cite{wang2020estvqa} & 20K \\
GQA~\cite{hudson2019gqa} & 944K \\
Hateful Memes~\cite{kiela2020hateful} & 8.5K \\
IconQA~\cite{lu2021iconqa} & 62K \\
iNaturalist-2018~\cite{vanhorn2018inaturalist} & 438K \\
LVIS-Instruct4V~\cite{wang2023lvisinstruct4v} & 223K \\
MMDU~\cite{liu2024mmdu} & 50K \\
OK-VQA~\cite{marino2019okvqa} & 9.0K \\
ProVision-10M~\cite{zhang2024provision} & 19.9M \\
SoM-LLaVA~\cite{yan2024listitems} & 631K \\
Spot-the-Diff~\cite{jhamtani2018spotthediff} & 9.5K \\
ViQuAE~\cite{lerner2022viquae} & 1.2K \\
VisDial~\cite{das2017visdial} & 124K \\
Visual7W~\cite{zhu2016visual7w} & 31K \\
VQAv2~\cite{goyal2017making} & 444K \\
VSR~\cite{liu2023vsr} & 7.4K \\
\midrule
\textbf{Total} & \textbf{24.7M} \\
\midrule
\rowcolor{instructcolor}\multicolumn{2}{l}{\textit{Grounding \& Counting}} \\
Dataset & Size \\
\midrule
All-Seeing-V2~\cite{wang2024allseeingv2} & 123K \\
LRV-Instruction~\cite{liu2024mitigating} & 341K \\
Objects365~\cite{shao2019objects365} & 1.7M \\
PixMo-Points~\cite{deitke2025molmo} & 276K \\
RefCOCO/+/g~\cite{kazemzadeh2014referitgame,yu2016modeling} & 59K \\
TallyQA~\cite{acharya2019tallyqa} & 249K \\
TolokaVQA~\cite{ustalov2023tolokavqa} & 39K \\
V3Det~\cite{wang2023v3det} & 177K \\
\midrule
\textbf{Total} & \textbf{3.0M} \\
\midrule
\rowcolor{instructcolor}\multicolumn{2}{l}{\textit{Math}} \\
Dataset & Size \\
\midrule
CLEVR-Math~\cite{lindstrom2022clevrmath} & 788K \\
Geometry3K~\cite{lu2021intergps} & 9.6K \\
GeomVerse~\cite{kazemi2023geomverse} & 9.3K \\
GeoQA+~\cite{cao2022geoqa} & 17K \\
MAVIS-Function~\cite{zhang2024mavis} & 200K \\
MAVIS-Geometry~\cite{zhang2024mavis} & 1.2M \\
UniGeo (Calc.)~\cite{chen2022unigeo} & 5.0K \\
UniGeo (Proof)~\cite{chen2022unigeo} & 9.8K \\
\midrule
\textbf{Total} & \textbf{2.2M} \\
\bottomrule
\end{tabular}
\end{minipage}
\hfill
\begin{minipage}[t]{0.245\textwidth}
\centering
\begin{tabular}{lr}
\toprule
\rowcolor{instructcolor}\multicolumn{2}{l}{\textit{Naive OCR}} \\
Dataset & Size \\
\midrule
AnyWord-3M~\cite{tuo2023anytext} & 2.9M \\
ArT~\cite{chng2019icdar} & 50K \\
CASIA~\cite{liu2011casia} & 1.1M \\
Chinese-OCR~\cite{xu2018chineseocr} & 5.8K \\
COCO-Text~\cite{veit2016cocotext} & 17K \\
CTW~\cite{yuan2019ctw} & 23K \\
EATEN~\cite{guo2019eaten} & 470K \\
HME-100K~\cite{yuan2022hme100k} & 74K \\
IAM~\cite{marti2002iam} & 5.6K \\
LSVT~\cite{sun2019icdar} & 400K \\
MTWI~\cite{he2018mtwi} & 9.9K \\
ParSynth-OCR-200K~\cite{singh2021parsynth} & 180K \\
POIE~\cite{kuang2023poie} & 2.3K \\
ReCTS~\cite{zhang2019icdar} & 20K \\
RenderedText~\cite{wendlerf2023renderedtext} & 12.0M \\
SROIE-2019~\cite{huang2019icdar} & 34K \\
SynthDoG~\cite{kim2022donut} & 2.0M \\
SynthText~\cite{gupta2016synthtext} & 856K \\
\midrule
\textbf{Total} & \textbf{20.2M} \\
\midrule
\rowcolor{instructcolor}\multicolumn{2}{l}{\textit{OCR QA}} \\
Dataset & Size \\
\midrule
ArXivQA~\cite{li2024multimodalarxiv} & 100K \\
Docmatix~\cite{laurencon2024building} & 1.3M \\
DocReason25K~\cite{hu2024docowl15} & 22K \\
DocVQA~\cite{mathew2021docvqa} & 40K \\
InfoVQA~\cite{mathew2022infographicvqa} & 1.2K \\
KVQA~\cite{shah2019kvqa} & 25K \\
LLaVAR~\cite{zhang2023llavar} & 437K \\
MapQA~\cite{chang2022mapqa} & 483K \\
MathWriting~\cite{gervais2024mathwriting} & 625K \\
MultiUI~\cite{liu2024harnessing} & 7.3M \\
OCR-VQA~\cite{mishra2019ocrvqa} & 803K \\
Screen2Words~\cite{wang2021screen2words} & 16K \\
ST-VQA~\cite{biten2019stvqa} & 26K \\
TextOCR~\cite{singh2021textocr} & 22K \\
TextVQA~\cite{singh2019towards} & 23K \\
VisualMRC~\cite{tanaka2021visualmrc} & 11K \\
\midrule
\textbf{Total} & \textbf{11.2M} \\
\bottomrule
\end{tabular}
\end{minipage}
\hfill
\begin{minipage}[t]{0.245\textwidth}
\centering
\begin{tabular}{lr}
\toprule
\rowcolor{instructcolor}\multicolumn{2}{l}{\textit{Science}} \\
Dataset & Size \\
\midrule
AI2D~\cite{kembhavi2016ai2d} & 16K \\
ImageCLEF~\cite{ionescu2024imageclef} & 80K \\
LLaVA-Med (FT)~\cite{li2023llavamed} & 51K \\
LLaVA-Med (PT)~\cite{li2023llavamed} & 467K \\
PathVQA~\cite{he2020pathvqa} & 20K \\
PMC-VQA~\cite{zhang2023pmcvqa} & 330K \\
ScienceQA~\cite{lu2022learn} & 6.3K \\
SLAKE~\cite{liu2021slake} & 9.5K \\
TQA~\cite{kembhavi2017tqa} & 25K \\
VisualWebInstruct~\cite{jia2025visualwebinstruct} & 1.1M \\
VQA-RAD~\cite{lau2018vqarad} & 1.8K \\
WebSight~\cite{laurencon2024unlocking} & 2.0M \\
\midrule
\textbf{Total} & \textbf{4.1M} \\
\midrule
\rowcolor{textcolor}\multicolumn{2}{l}{\textit{Text}} \\
Dataset & Size \\
\midrule
FLAN~\cite{wei2022finetuned} & 265M \\
FLAN-v2~\cite{longpre2023flan} & 457M \\
SlimOrca~\cite{lian2023slimorca} & 518K \\
UltraChat-200K~\cite{ding2023enhancing} & 463K \\
UltraFeedback~\cite{cui2024ultrafeedback} & 256K \\
WizardLM-Evol-70K~\cite{xu2024wizardlm} & 70K \\
LIMA~\cite{zhou2023lima} & 1.3K \\
No Robots~\cite{rajani2023no_robots} & 9.6K \\
Unnatural Instr.~\cite{honovich2023unnatural} & 69K \\
MOSS~\cite{sun2024moss} & 571K \\
Llama3-Magpie-Pro~\cite{xu2024magpie} & 1.0M \\
Magpie-Qwen2-Pro~\cite{xu2024magpie} & 1.0M \\
Firefly~\cite{yang2023firefly} & 1.6M \\
Dolly~\cite{conover2023dolly} & 15K \\
KOpen-Hermes-25~\cite{kopenhermes2024} & 60K \\
OpenAI-TLDR~\cite{stiennon2020learning} & 117K \\
Saraswati-CoT~\cite{saraswaticot2024} & 150K \\
CodeFeedback~\cite{zheng2024opencodeinterpreter} & 66K \\
Glaive-Code~\cite{glaive2023} & 136K \\
xCoder-80K~\cite{wang2024xcoder} & 80K \\
LeetCode~\cite{leetcode2023} & 2.4K \\
Evol-Code~\cite{luo2024wizardcoder} & 78K \\
LongCite-45K~\cite{zhang2024longcite} & 45K \\
LongInstruct-Para.~\cite{longinstruct2024} & 14K \\
Long-QLoRA~\cite{yang2023longqlora} & 37K \\
LongAlpaca~\cite{chen2024longlora} & 12K \\
GSM8K (Socratic)~\cite{cobbe2021gsm8k} & 7.5K \\
MetaMathQA~\cite{yu2024metamath} & 395K \\
MathQA~\cite{amini2019mathqa} & 30K \\
Numina-Math-1.5~\cite{numina_math_15} & 767K \\
Numina-Math-TIR~\cite{numina_math_tir} & 73K \\
Orca-Math~\cite{mitra2024orcamath} & 200K \\
InfinityMath~\cite{zhang2024infinitymath} & 101K \\
\midrule
\textbf{Total} & \textbf{730M} \\
\bottomrule
\end{tabular}
\end{minipage}
\end{table}

\begin{table}[!h]
\centering
\caption{\textbf{DCVLM pool per-dataset multimodal-token counts.} The mix combines \colorbox{capcolor}{captioning data}, \colorbox{mmdoccolor}{multimodal documents}, \colorbox{instructcolor}{visual instruction-tuning data} (organised by capability), and \colorbox{textcolor}{text-only data}. Token counts are measured using the InternVL-2.5 tokenizer~\citep{chen2024expanding}. Across all 160 datasets, the our pool contains \textbf{6.0T} multimodal tokens (${1536}$ tokens/sample on average).}
\label{tab:dcvlm-datamix-tokens}
\scriptsize
\setlength{\tabcolsep}{1.2pt}
\renewcommand{\arraystretch}{1.05}
\begin{minipage}[t]{0.245\textwidth}
\centering
\begin{tabular}{lr}
\toprule
\rowcolor{capcolor}\multicolumn{2}{l}{\textit{Captioning}} \\
Dataset & Size \\
\midrule
ReLAION-2B-en~\cite{schuhmann2022laion} & 2.6T \\
DataComp-1B~\cite{gadre2023datacomp} & 2.3T \\
AS-100M~\cite{wang2024allseeing} & 6.2B \\
GRIT (Cap.)~\cite{peng2023kosmos2} & 26.8B \\
InternVL-SA1B~\cite{chen2024internvl} & 26.5B \\
FaceCaption-15M~\cite{dai2024facecaption} & 23.2B \\
PixMo-Cap~\cite{deitke2025molmo} & 1.5B \\
ShareGPT-4o~\cite{chen2024sharegpt4v} & 160M \\
TextOCR-GPT4V~\cite{carter2024textocrgpt4v} & 66.4M \\
TextCaps~\cite{sidorov2020textcaps} & 284M \\
COCO (Cap.)~\cite{chen2015cococaptions} & 1.4B \\
OpenImages (Cap.)~\cite{kuznetsova2020openimages} & 1.3B \\
SEA-VL~\cite{cahyawijaya2025seavl} & 2.6B \\
\midrule
\textbf{Total} & \textbf{5.0T} \\
\midrule
\rowcolor{mmdoccolor}\multicolumn{2}{l}{\textit{Multimodal Docs}} \\
Dataset & Size \\
\midrule
MINT-HTML~\cite{awadalla2024mint} & 190B \\
MINT-PDF~\cite{awadalla2024mint} & 14.8B \\
OmniCC~\cite{li2024omnicorpus} & 228B \\
Multimodal Textbook~\cite{zhang2025textbook} & 2.6B \\
WanJuan~\cite{he2023wanjuan} & 4.3B \\
\midrule
\textbf{Total} & \textbf{440B} \\
\midrule
\rowcolor{instructcolor}\multicolumn{2}{l}{\textit{Cap. \& Know.}} \\
Dataset & Size \\
\midrule
Art500K~\cite{mao2017art500k} & 1.1B \\
LLaVA-595K~\cite{liu2024improved} & 195M \\
MMInstruct~\cite{liu2024mminstruct} & 920M \\
ShareGPT4V~\cite{chen2024sharegpt4v} & 3.0B \\
SVIT~\cite{zhao2023svit} & 9.5B \\
\midrule
\textbf{Total} & \textbf{14.7B} \\
\midrule
\rowcolor{instructcolor}\multicolumn{2}{l}{\textit{Chart \& Table}} \\
Dataset & Size \\
\midrule
BigDocsBench~\cite{rodriguez2025bigdocs} & 979M \\
Chart2Text~\cite{kantharaj2022chart2text} & 17.6M \\
ChartGemma~\cite{masry2024chartgemma} & 366M \\
ChartLlama~\cite{han2023chartllama} & 3.0M \\
ChartQA~\cite{masry2022chartqa} & 57.0M \\
ChartX~\cite{xia2024chartx} & 37.7M \\
CoSyn-400K~\cite{yang2025scaling} & 1.1B \\
DocStruct4M~\cite{hu2024docowl15} & 9.8B \\
DVQA~\cite{kafle2018dvqa} & 745M \\
FigureQA~\cite{kahou2017figureqa} & 2.4B \\
FinTabNet~\cite{zheng2021globaltable} & 15.8B \\
MMC-Instruct~\cite{liu2024mmc} & 861M \\
PixMo-Docs~\cite{deitke2025molmo} & 664M \\
PlotQA~\cite{methani2020plotqa} & 43.9B \\
PosterSum~\cite{saxena2025postersum} & 27.0M \\
SBSFigures~\cite{shinoda2024sbsfigures} & 11.4B \\
SciGraphQA~\cite{li2023scigraphqa} & 670M \\
SimChart9K~\cite{xia2023structchart} & 154M \\
SPIQA~\cite{pramanick2024spiqa} & 469M \\
TabMWP~\cite{lu2023tabmwp} & 42.0M \\
UniChart~\cite{masry2023unichart} & 13.1B \\
VisText~\cite{tang2023vistext} & 20.8M \\
\midrule
\textbf{Total} & \textbf{103B} \\
\bottomrule
\end{tabular}
\end{minipage}
\hfill
\begin{minipage}[t]{0.245\textwidth}
\centering
\begin{tabular}{lr}
\toprule
\rowcolor{instructcolor}\multicolumn{2}{l}{\textit{General QA}} \\
Dataset & Size \\
\midrule
AlgoPuzzleVQA~\cite{ghosal2024algopuzzlevqa} & 4.3M \\
ALLaVA~\cite{chen2024allava} & 3.3B \\
A-OKVQA~\cite{schwenk2022aokvqa} & 42.3M \\
Cambrian-GPT4o~\cite{tong2024cambrian} & 136M \\
EST-VQA~\cite{wang2020estvqa} & 44.6M \\
GQA~\cite{hudson2019gqa} & 2.3B \\
Hateful Memes~\cite{kiela2020hateful} & 17.9M \\
IconQA~\cite{lu2021iconqa} & 115M \\
iNaturalist-2018~\cite{vanhorn2018inaturalist} & 1.1B \\
LVIS-Instruct4V~\cite{wang2023lvisinstruct4v} & 608M \\
MMDU~\cite{liu2024mmdu} & 324M \\
OK-VQA~\cite{marino2019okvqa} & 21.5M \\
ProVision-10M~\cite{zhang2024provision} & 47.9B \\
SoM-LLaVA~\cite{yan2024listitems} & 1.7B \\
Spot-the-Diff~\cite{jhamtani2018spotthediff} & 5.7M \\
ViQuAE~\cite{lerner2022viquae} & 2.9M \\
VisDial~\cite{das2017visdial} & 322M \\
Visual7W~\cite{zhu2016visual7w} & 82.2M \\
VQAv2~\cite{goyal2017making} & 1.1B \\
VSR~\cite{liu2023vsr} & 19.3M \\
\midrule
\textbf{Total} & \textbf{59.2B} \\
\midrule
\rowcolor{instructcolor}\multicolumn{2}{l}{\textit{Grounding \& Counting}} \\
Dataset & Size \\
\midrule
All-Seeing-V2~\cite{wang2024allseeingv2} & 364M \\
LRV-Instruction~\cite{liu2024mitigating} & 820M \\
Objects365~\cite{shao2019objects365} & 4.9B \\
PixMo-Points~\cite{deitke2025molmo} & 582M \\
RefCOCO/+/g~\cite{kazemzadeh2014referitgame,yu2016modeling} & 151M \\
TallyQA~\cite{acharya2019tallyqa} & 611M \\
TolokaVQA~\cite{ustalov2023tolokavqa} & 98.3M \\
V3Det~\cite{wang2023v3det} & 469M \\
\midrule
\textbf{Total} & \textbf{8.0B} \\
\midrule
\rowcolor{instructcolor}\multicolumn{2}{l}{\textit{Math}} \\
Dataset & Size \\
\midrule
CLEVR-Math~\cite{lindstrom2022clevrmath} & 1.5B \\
Geometry3K~\cite{lu2021intergps} & 17.5M \\
GeomVerse~\cite{kazemi2023geomverse} & 27.0M \\
GeoQA+~\cite{cao2022geoqa} & 27.3M \\
MAVIS-Function~\cite{zhang2024mavis} & 728M \\
MAVIS-Geometry~\cite{zhang2024mavis} & 2.9B \\
UniGeo (Calc.)~\cite{chen2022unigeo} & 8.3M \\
UniGeo (Proof)~\cite{chen2022unigeo} & 17.6M \\
\midrule
\textbf{Total} & \textbf{5.2B} \\
\bottomrule
\end{tabular}
\end{minipage}
\hfill
\begin{minipage}[t]{0.245\textwidth}
\centering
\begin{tabular}{lr}
\toprule
\rowcolor{instructcolor}\multicolumn{2}{l}{\textit{Naive OCR}} \\
Dataset & Size \\
\midrule
AnyWord-3M~\cite{tuo2023anytext} & 1.2B \\
ArT~\cite{chng2019icdar} & 85.5M \\
CASIA~\cite{liu2011casia} & 2.0B \\
Chinese-OCR~\cite{xu2018chineseocr} & 15.8M \\
COCO-Text~\cite{veit2016cocotext} & 43.9M \\
CTW~\cite{yuan2019ctw} & 81.4M \\
EATEN~\cite{guo2019eaten} & 833M \\
HME-100K~\cite{yuan2022hme100k} & 132M \\
IAM~\cite{marti2002iam} & 11.5M \\
LSVT~\cite{sun2019icdar} & 687M \\
MTWI~\cite{he2018mtwi} & 18.5M \\
ParSynth-OCR-200K~\cite{singh2021parsynth} & 318M \\
POIE~\cite{kuang2023poie} & 4.9M \\
ReCTS~\cite{zhang2019icdar} & 39.0M \\
RenderedText~\cite{wendlerf2023renderedtext} & 31.8B \\
SROIE-2019~\cite{huang2019icdar} & 70.4M \\
SynthDoG~\cite{kim2022donut} & 5.1B \\
SynthText~\cite{gupta2016synthtext} & 1.8B \\
\midrule
\textbf{Total} & \textbf{44.2B} \\
\midrule
\rowcolor{instructcolor}\multicolumn{2}{l}{\textit{OCR QA}} \\
Dataset & Size \\
\midrule
ArXivQA~\cite{li2024multimodalarxiv} & 268M \\
Docmatix~\cite{laurencon2024building} & 4.4B \\
DocReason25K~\cite{hu2024docowl15} & 64.5M \\
DocVQA~\cite{mathew2021docvqa} & 132M \\
InfoVQA~\cite{mathew2022infographicvqa} & 2.5M \\
KVQA~\cite{shah2019kvqa} & 67.0M \\
LLaVAR~\cite{zhang2023llavar} & 696M \\
MapQA~\cite{chang2022mapqa} & 1.6B \\
MathWriting~\cite{gervais2024mathwriting} & 1.2B \\
MultiUI~\cite{liu2024harnessing} & 18.2B \\
OCR-VQA~\cite{mishra2019ocrvqa} & 1.8B \\
Screen2Words~\cite{wang2021screen2words} & 32.8M \\
ST-VQA~\cite{biten2019stvqa} & 64.5M \\
TextOCR~\cite{singh2021textocr} & 82.7M \\
TextVQA~\cite{singh2019towards} & 62.0M \\
VisualMRC~\cite{tanaka2021visualmrc} & 21.7M \\
\midrule
\textbf{Total} & \textbf{28.7B} \\
\bottomrule
\end{tabular}
\end{minipage}
\hfill
\begin{minipage}[t]{0.245\textwidth}
\centering
\begin{tabular}{lr}
\toprule
\rowcolor{instructcolor}\multicolumn{2}{l}{\textit{Science}} \\
Dataset & Size \\
\midrule
AI2D~\cite{kembhavi2016ai2d} & 34.6M \\
ImageCLEF~\cite{ionescu2024imageclef} & 171M \\
LLaVA-Med (FT)~\cite{li2023llavamed} & 116M \\
LLaVA-Med (PT)~\cite{li2023llavamed} & 1.0B \\
PathVQA~\cite{he2020pathvqa} & 38.5M \\
PMC-VQA~\cite{zhang2023pmcvqa} & 640M \\
ScienceQA~\cite{lu2022learn} & 12.9M \\
SLAKE~\cite{liu2021slake} & 9.6M \\
TQA~\cite{kembhavi2017tqa} & 18.4M \\
VisualWebInstruct~\cite{jia2025visualwebinstruct} & 1.5B \\
VQA-RAD~\cite{lau2018vqarad} & 4.0M \\
WebSight~\cite{laurencon2024unlocking} & 5.8B \\
\midrule
\textbf{Total} & \textbf{9.3B} \\
\midrule
\rowcolor{textcolor}\multicolumn{2}{l}{\textit{Text}} \\
Dataset & Size \\
\midrule
FLAN~\cite{wei2022finetuned} & 141B \\
FLAN-v2~\cite{longpre2023flan} & 177B \\
SlimOrca~\cite{lian2023slimorca} & 208M \\
UltraChat-200K~\cite{ding2023enhancing} & 490M \\
UltraFeedback~\cite{cui2024ultrafeedback} & 116M \\
WizardLM-Evol-70K~\cite{xu2024wizardlm} & 31.4M \\
LIMA~\cite{zhou2023lima} & 691K \\
No Robots~\cite{rajani2023no_robots} & 3.1M \\
Unnatural Instr.~\cite{honovich2023unnatural} & 9.9M \\
MOSS~\cite{sun2024moss} & 184M \\
Llama3-Magpie-Pro~\cite{xu2024magpie} & 530M \\
Magpie-Qwen2-Pro~\cite{xu2024magpie} & 418M \\
Firefly~\cite{yang2023firefly} & 338M \\
Dolly~\cite{conover2023dolly} & 2.2M \\
KOpen-Hermes-25~\cite{kopenhermes2024} & 30.0M \\
OpenAI-TLDR~\cite{stiennon2020learning} & 50.7M \\
Saraswati-CoT~\cite{saraswaticot2024} & 43.2M \\
CodeFeedback~\cite{zheng2024opencodeinterpreter} & 93.1M \\
Glaive-Code~\cite{glaive2023} & 55.3M \\
xCoder-80K~\cite{wang2024xcoder} & 75.4M \\
LeetCode~\cite{leetcode2023} & 915K \\
Evol-Code~\cite{luo2024wizardcoder} & 31.2M \\
LongCite-45K~\cite{zhang2024longcite} & 650M \\
LongInstruct-Para.~\cite{longinstruct2024} & 207M \\
Long-QLoRA~\cite{yang2023longqlora} & 120M \\
LongAlpaca~\cite{chen2024longlora} & 99.7M \\
GSM8K (Socratic)~\cite{cobbe2021gsm8k} & 2.0M \\
MetaMathQA~\cite{yu2024metamath} & 112M \\
MathQA~\cite{amini2019mathqa} & 6.8M \\
Numina-Math-1.5~\cite{numina_math_15} & 423M \\
Numina-Math-TIR~\cite{numina_math_tir} & 55.2M \\
Orca-Math~\cite{mitra2024orcamath} & 79.9M \\
InfinityMath~\cite{zhang2024infinitymath} & 45.3M \\
\midrule
\textbf{Total} & \textbf{323B} \\
\bottomrule
\end{tabular}
\end{minipage}
\end{table}

\cref{fig:pool-composition} summarises how our pool is split across the four data types, in terms of both samples and multimodal tokens. 
The detailed per-category sample and token totals are given in \cref{tab:dcvlm-datamix-samples,tab:dcvlm-datamix-tokens}.
The two views differ in informative ways:

\begin{itemize}[leftmargin=*,itemsep=1pt]
\item \textbf{Image-caption pairs} dominate both axes (74\% of samples, 83\% of tokens). 
The token share exceeds the sample share because every image-caption sample contributes $  $256 visual tokens \textit{per-tile} on top of a short caption, inflating the per-sample token count relative to the text-only data.
\item \textbf{Text-only} data shows the opposite asymmetry: 19\% of samples but only 5\% of tokens, because individual text samples are short (440 tok/sample on average) and contribute no visual tokens.
\item \textbf{Multimodal documents}, despite making up just 4\% of samples, contribute 7\% of tokens. 
They are the densest data type, having $\sim$3K tok/sample on average, since each sample typically interleaves several images with surrounding text.
\item \textbf{Instruction-tuning} data spans 8 capability categories and sits between image-caption pairs and multimodal documents in density ($\sim$2.2K tok/sample). 
Also in this case, the presence of multi-image samples contributes to a greater per-sample token average.
\end{itemize}

\subsection{Per-Dataset Variation}\label{app:pool-perdataset}

\figpoolscatter

The 160 datasets in the pool span six orders of magnitude in sample count---from $\sim$1K (e.g.\ ChartLlama, LIMA, ViQuAE) to $\sim$1.5B (ReLAION-2B-en, DataComp-1B)---and four orders of magnitude in tokens-per-sample (\cref{fig:pool-scatter}). We dig into the per-data-type statistics below:

\begin{itemize}[leftmargin=*,itemsep=1pt]
\item \textbf{Image-caption datasets} cluster tightly along the 1--2K tok/sample diagonal (\cref{fig:pool-scatter}). The narrow band reflects that image-caption tokens are dominated by the fixed-cost visual-token contribution ($\sim$256 tokens for a single $448{\times}448$ tile after pixel shuffling), with caption length contributing only secondary variation.
\item \textbf{Multimodal documents} sit a decade higher, in the 3--6K tok/sample regime, because each sample carries multiple images and longer interleaved text spans.
\item \textbf{Text-only datasets} occupy the widest tokens-per-sample band of any data type (100--15K), with two distinct clusters: short-form instruction data (Dolly, Unnatural-Instructions, MathQA, GSM8K-Socratic) at $\sim$200--400 tok/sample, and long-context corpora (LongCite-45K, LongInstruct-Para., LongAlpaca, Long-QLoRA) above 5K tok/sample.
\item \textbf{Instruction-tuning datasets} span the largest dynamic range in size but a relatively narrow tokens-per-sample band ($\sim$1--3K). The handful of high-density outliers (MMDU, MapQA, MathWriting) correspond to multi-turn or multi-image conversations.
\end{itemize}

\subsection{Data Sources and Licensing}\label{appsub:licensing}

We source our \dcvlm{} pool from four different data-types, each with a distinct sourcing strategy. Image-caption pairs are primarily sourced from web-crawled image-alt-text corpora (DataComp-1B~\citep{gadre2023datacomp}, ReLAION-2B~\citep{schuhmann2022laion}) and synthetic/human-curated caption datasets (PixMo-Cap~\citep{deitke2025molmo}, ShareGPT-4o~\citep{chen2024sharegpt4v}, GRIT~\citep{peng2023kosmos2}).
Multimodal documents come from web-crawled interleaved sources (MINT-1T-HTML~\citep{awadalla2024mint}, OmniCC~\citep{li2024omnicorpus}, WanJuan~\citep{he2023wanjuan}) and curated PDF corpora (Multimodal-Textbook~\citep{zhang2025textbook}, MINT-1T-PDF~\citep{awadalla2024mint}).
Instruction-tuning data is aggregated from academic benchmarks with train splits, synthetic generation pipelines, and existing curated datasets across 8 capability categories.
Text-only data combines general instruction sets (Dolly, FLAN/FLAN-v2, SlimOrca) with long-context (LongAlpaca, LongCite-45K) and code/math reasoning (Numina-Math, MetaMathQA, Glaive-Code) corpora. In~\cref{tab:dcvlm-datamix-licenses}, we provide the licensing information and original sources from which we collected each sub-dataset in our \dcvlm{} pool.

\begingroup
\scriptsize
\setlength{\tabcolsep}{3pt}
\renewcommand{\arraystretch}{1.05}
\begin{longtable}{@{}>{\raggedright\arraybackslash}p{0.19\textwidth}>{\raggedright\arraybackslash}p{0.59\textwidth}>{\raggedright\arraybackslash}p{0.17\textwidth}@{}}
\caption{\textbf{DCVLM pool per-dataset licenses.} Licenses are grouped according to the datamix categories: \colorbox{capcolor}{captioning data}, \colorbox{mmdoccolor}{multimodal documents}, \colorbox{instructcolor}{visual instruction-tuning data}, and \colorbox{textcolor}{text-only data}. ``Unknown'' indicates that a public license was not clearly specified in the listed source.}
\label{tab:dcvlm-datamix-licenses} \\
\toprule
Dataset & License & Source \\
\midrule
\endfirsthead
\caption[]{\textbf{DCVLM pool per-dataset licenses and sources} (continued).} \\
\toprule
Dataset & License & Source \\
\midrule
\endhead
\midrule
\multicolumn{3}{r}{\textit{Continued on next page}} \\
\endfoot
\bottomrule
\endlastfoot
\rowcolor{capcolor}\multicolumn{3}{l}{\textit{Captioning}} \\
ReLAION-2B-en & CC BY 4.0 (gated; access requires accepting HF terms/contact sharing) & \href{https://huggingface.co/datasets/laion/relaion2B-en-research}{source} \\
DataComp-1B & CC BY 4.0 & \href{https://huggingface.co/datasets/mlfoundations/datacomp_1b}{source} \\
AS-100M & Apache-2.0 & \href{https://huggingface.co/datasets/OpenGVLab/AS-100M}{source} \\
GRIT (Cap.) & MS-PL & \href{https://huggingface.co/datasets/zzliang/GRIT}{source} \\
InternVL-SA1B & MIT & \href{https://huggingface.co/datasets/OpenGVLab/InternVL-SA-1B-Caption}{source} \\
FaceCaption-15M & CC BY 4.0 + research/education notice & \href{https://huggingface.co/datasets/OpenFace-CQUPT/FaceCaption-15M}{source} \\
PixMo-Cap & ODC-BY-1.0 & \href{https://huggingface.co/datasets/allenai/pixmo-cap}{source} \\
ShareGPT-4o & MIT; HF-gated contact-sharing/access terms; source video copyrights/platform terms and academic-research notice apply & \href{https://huggingface.co/datasets/OpenGVLab/ShareGPT-4o}{source} \\
TextOCR-GPT4V & Apache-2.0 & \href{https://huggingface.co/datasets/lmms-lab/LLaVA-OneVision-Data/tree/main/textocr(gpt4v)}{source} \\
TextCaps & CC BY 4.0 & \href{https://dl.fbaipublicfiles.com/textvqa/data/textcaps/TextCaps_0.1_train.json}{source 1}; \href{https://dl.fbaipublicfiles.com/textvqa/images/train_val_images.zip}{source 2} \\
COCO (Cap.) & CC BY 4.0 for annotations; images retain original COCO/Flickr licenses & \href{https://storage.googleapis.com/sfr-vision-language-research/datasets/coco_karpathy_train.json}{source} \\
OpenImages (Cap.) & CC BY 4.0 for annotations; images retain original Open Images licenses & \href{https://storage.googleapis.com/localized-narratives/annotations/open_images_train_v6_localized_narratives-0000*-of-00010.jsonl}{source} \\
SEA-VL & CC BY-SA 4.0 & \href{https://huggingface.co/datasets/SEACrowd/sea-vl_crowdsourced}{source 1}; \href{https://huggingface.co/datasets/SEACrowd/sea-vl_crawling}{source 2} \\
\midrule
\rowcolor{mmdoccolor}\multicolumn{3}{l}{\textit{Multimodal Docs}} \\
MINT-HTML & CC BY 4.0 & \href{https://huggingface.co/datasets/mlfoundations/MINT-1T-HTML}{source} \\
MINT-PDF & CC BY 4.0 & \href{https://huggingface.co/collections/mlfoundations/mint-1t-6690216ca4d0df7e518dde1c}{source} \\
OmniCC & CC BY 4.0 & \href{https://huggingface.co/datasets/OpenGVLab/OmniCorpus-CC}{source} \\
Multimodal Textbook & Apache-2.0 & \href{https://huggingface.co/datasets/DAMO-NLP-SG/multimodal_textbook}{source} \\
WanJuan & CC BY 4.0 & \href{https://opendatalab.org.cn/WanJuan1.0}{source} \\
\midrule
\rowcolor{instructcolor}\multicolumn{3}{l}{\textit{Cap. \& Know.}} \\
Art500K & Custom non-commercial research-only terms; images retain third-party rights & \href{https://drive.google.com/file/d/1sWqPDINj7xSBaYmYxOsaYOLBwJZvg4BB/view}{source 1}; \href{https://drive.google.com/file/d/1HpEiuC7cQRBu0ykpbEH0iKb1Hh5rrvtE/view}{source 2} \\
LLaVA-595K & Other: must comply with CC-3M license and BLIP license (HF tag: other) & \href{https://huggingface.co/datasets/liuhaotian/LLaVA-CC3M-Pretrain-595K}{source} \\
MMInstruct & Apache-2.0 & \href{https://huggingface.co/datasets/yuecao0119/MMInstruct-GPT4V}{source} \\
ShareGPT4V & CC BY-NC 4.0 + OpenAI ToU & \href{https://huggingface.co/datasets/Lin-Chen/ShareGPT4V}{source} \\
SVIT & CC BY 4.0; OpenAI ToU and original Visual Genome/MS-COCO image/annotation licenses apply; HF-gated usage notice applies & \href{https://huggingface.co/datasets/BAAI/SVIT}{source} \\
\midrule
\rowcolor{instructcolor}\multicolumn{3}{l}{\textit{Chart \& Table}} \\
BigDocsBench & CC BY 4.0 for ServiceNow-generated parts; per-sample upstream terms and Llama 3.1 terms may apply & \href{https://huggingface.co/datasets/ServiceNow/BigDocs-Bench}{source} \\
Chart2Text & Unknown / not publicly specified & \href{https://github.com/JasonObeid/Chart2Text}{source 1}; \href{https://github.com/JasonObeid/Chart2TextImages}{source 2} \\
ChartGemma & Unknown / not publicly specified & \href{https://huggingface.co/datasets/ahmed-masry/ChartGemma}{source} \\
ChartLlama & Unknown / not publicly specified & \href{https://huggingface.co/datasets/listen2you002/ChartLlama-Dataset}{source} \\
ChartQA & Apache-2.0 for Cambrian-10M formatted version; original ChartQA terms may also apply & \href{https://huggingface.co/datasets/nyu-visionx/Cambrian-10M/resolve/main/chartqa.tar.gz}{source 1}; \href{https://huggingface.co/datasets/nyu-visionx/Cambrian-10M/resolve/main/jsons/Cambrian10M.jsonl}{source 2} \\
ChartX & Apache-2.0 & \href{https://huggingface.co/datasets/U4R/ChartX}{source} \\
CoSyn-400K & ODC-BY-1.0 (plus AI-generated-output/provider terms stated in card) & \href{https://huggingface.co/datasets/allenai/CoSyn-400K}{source} \\
DocStruct4M & Apache-2.0 & \href{https://huggingface.co/datasets/mPLUG/DocStruct4M}{source} \\
DVQA & Apache-2.0 for Cambrian-10M formatted version; original DVQA terms may also apply & \href{https://huggingface.co/datasets/nyu-visionx/Cambrian-10M/resolve/main/dvqa.tar.gz}{source 1}; \href{https://huggingface.co/datasets/nyu-visionx/Cambrian-10M/resolve/main/jsons/Cambrian10M.jsonl}{source 2} \\
FigureQA & Unknown / dataset files license not clearly stated; generation code MIT & \href{https://download.microsoft.com/download/c/3/1/c315c9d8-8239-487e-a895-2d3ff805b508/figureqa-train1-v1.tar.gz}{source} \\
FinTabNet & CDLA-Permissive-2.0 & \href{https://huggingface.co/datasets/bsmock/FinTabNet.c}{source} \\
MMC-Instruct & CC BY-SA 4.0 & \href{https://huggingface.co/datasets/xywang1/MMC}{source} \\
PixMo-Docs & ODC-BY-1.0 (plus AI-generated-output/provider terms stated in card) & \href{https://huggingface.co/datasets/allenai/pixmo-docs}{source} \\
PlotQA & CC BY 4.0 & \href{https://drive.google.com/file/d/1AYuaPX-Lx7T0GZvnsPgN11Twq2FZbWXL/view?usp=sharing}{source 1}; \href{https://drive.google.com/file/d/1bBSUutd-Die27ZH3QTMVhBjW9l5hAwrr/view?usp=sharing}{source 2}; \href{https://drive.google.com/file/d/1UNvkdq1YJD_ne6D3zbWtoQij37AtfpNp/view?usp=sharing}{source 3}; \href{https://drive.google.com/file/d/1VzWwxBVrlep17BGZU17GpLuGpwjyWbzq/view?usp=sharing}{source 4} \\
PosterSum & Unknown / not publicly specified & \href{https://huggingface.co/datasets/rohitsaxena/PosterSum}{source} \\
SBSFigures & Unknown / not publicly specified & \href{https://huggingface.co/datasets/omron-sinicx/sbsfigures}{source} \\
SciGraphQA & MIT & \href{https://huggingface.co/datasets/alexshengzhili/SciGraphQA-295K-train}{source} \\
SimChart9K & Unknown / not publicly specified & \href{https://drive.google.com/file/d/1M_NA3sIJNwCUfqB1HH0p4lsnEm3NbvsI/view?usp=sharing}{source} \\
SPIQA & CC BY 4.0 & \href{https://huggingface.co/datasets/google/spiqa}{source} \\
TabMWP & CC BY-NC-SA 4.0 for TabMWP dataset; MIT for repository code & \href{https://github.com/lupantech/PromptPG.git}{source} \\
UniChart & MIT for UniChart-pretrain-images; UniChart-pretrain-data license not publicly specified & \href{https://huggingface.co/datasets/ahmed-masry/UniChart-pretrain-images}{source 1}; \href{https://huggingface.co/datasets/ahmed-masry/unichart-pretrain-data}{source 2} \\
VisText & Unknown / not publicly specified on The Cauldron repo; original VisText terms may apply & \href{https://huggingface.co/datasets/HuggingFaceM4/the_cauldron/tree/main/vistext}{source} \\
\midrule
\rowcolor{instructcolor}\multicolumn{3}{l}{\textit{General QA}} \\
AlgoPuzzleVQA & Apache-2.0 & \href{https://huggingface.co/datasets/declare-lab/AlgoPuzzleVQA}{source} \\
ALLaVA & CC BY-NC 4.0 & \href{https://huggingface.co/datasets/FreedomIntelligence/ALLaVA-4V}{source} \\
A-OKVQA & Apache-2.0 for official repository; dataset archive has no separate license file; COCO image licenses apply & \href{https://prior-datasets.s3.us-east-2.amazonaws.com/aokvqa/aokvqa_v1p0.tar.gz}{source} \\
Cambrian-GPT4o & Apache-2.0 & \href{https://huggingface.co/datasets/nyu-visionx/Cambrian-10M}{source} \\
EST-VQA & Unknown / not publicly specified & \href{https://drive.google.com/file/d/15q8Hqx3uQm8mx4w3PZtactJVe1BRXug5/view}{source 1}; \href{https://drive.google.com/file/d/1M3fWhFWltjM_nFR0ALynUIwXatwKXb26/view}{source 2} \\
GQA & CC BY 4.0 & \href{https://downloads.cs.stanford.edu/nlp/data/gqa/questions1.2.zip}{source 1}; \href{https://downloads.cs.stanford.edu/nlp/data/gqa/images.zip}{source 2} \\
Hateful Memes & Custom non-commercial/research dataset terms & \href{https://www.kaggle.com/datasets/parthplc/facebook-hateful-meme-dataset}{source} \\
IconQA & CC BY-NC-SA 4.0 & \href{https://iconqa2021.s3.us-west-1.amazonaws.com/iconqa_data.zip}{source} \\
iNaturalist-2018 & Mixed image licenses; check per-image iNaturalist metadata & \href{https://github.com/visipedia/inat_comp/blob/master/2018/README.md}{source 1}; \href{https://ml-inat-competition-datasets.s3.amazonaws.com/2018/train_val2018.tar.gz}{source 2}; \href{https://ml-inat-competition-datasets.s3.amazonaws.com/2018/train2018.json.tar.gz}{source 3}; \href{https://ml-inat-competition-datasets.s3.amazonaws.com/2018/categories.json.tar.gz}{source 4} \\
LVIS-Instruct4V & Unknown / not publicly specified & \href{https://huggingface.co/datasets/X2FD/LVIS-Instruct4V}{source} \\
MMDU & CC BY-NC 4.0 + OpenAI ToU & \href{https://huggingface.co/datasets/laolao77/MMDU}{source} \\
OK-VQA & CC BY 4.0 for annotations; images retain original COCO/Flickr licenses & \href{https://okvqa.allenai.org/static/data/mscoco_train2014_annotations.json.zip}{source 1}; \href{https://okvqa.allenai.org/static/data/OpenEnded_mscoco_train2014_questions.json.zip}{source 2} \\
ProVision-10M & CC BY-NC 4.0 & \href{https://huggingface.co/datasets/Salesforce/ProVision-10M}{source} \\
SoM-LLaVA & Apache-2.0 & \href{https://huggingface.co/datasets/zzxslp/SoM-LLaVA}{source} \\
Spot-the-Diff & Unknown / not publicly specified on The Cauldron repo; original Spot-the-Diff terms may apply & \href{https://huggingface.co/datasets/HuggingFaceM4/the_cauldron/tree/main/spot_the_diff}{source} \\
ViQuAE & Unknown / not publicly specified & \href{https://huggingface.co/datasets/PaulLerner/viquae_dataset/resolve/main/train.jsonl}{source 1}; \href{https://huggingface.co/datasets/PaulLerner/viquae_images/resolve/main/images.tar.gz}{source 2} \\
VisDial & Unknown / not publicly specified & \href{https://huggingface.co/datasets/HuggingFaceM4/VisDial}{source} \\
Visual7W & MIT for Visual7W toolkit/repo; COCO image licenses apply; annotation license not separately specified & \href{http://vision.stanford.edu/yukezhu/visual7w_images.zip}{source 1}; \href{https://ai.stanford.edu/~yukez/papers/resources/dataset_v7w_telling.zip}{source 2}; \href{https://ai.stanford.edu/~yukez/papers/resources/dataset_v7w_pointing.zip}{source 3} \\
VQAv2 & CC BY 4.0 for annotations; images retain original COCO/Flickr licenses & \href{https://huggingface.co/datasets/HuggingFaceM4/the_cauldron/tree/main/vqav2}{source} \\
VSR & Apache-2.0 & \href{https://raw.githubusercontent.com/cambridgeltl/visual-spatial-reasoning/refs/heads/master/data/splits/random/train.jsonl}{source} \\
\midrule
\rowcolor{instructcolor}\multicolumn{3}{l}{\textit{Grounding \& Counting}} \\
All-Seeing-V2 & Apache-2.0 & \href{https://huggingface.co/datasets/OpenGVLab/AS-V2}{source} \\
LRV-Instruction & BSD-3-Clause for repository; source image/data terms may also apply & \href{https://github.com/FuxiaoLiu/LRV-Instruction?tab=readme-ov-file\#instruction-data}{source} \\
Objects365 & Academic-purpose only; annotations/website CC BY 4.0; images under Flickr terms and must not be redistributed; software MIT & \href{https://dorc.ks3-cn-beijing.ksyun.com/data-set/2020Objects365\%E6\%95\%B0\%E6\%8D\%AE\%E9\%9B\%86/train/zhiyuan_objv2_train.tar.gz}{source 1}; \href{https://dorc.ks3-cn-beijing.ksyun.com/data-set/2020Objects365\%E6\%95\%B0\%E6\%8D\%AE\%E9\%9B\%86/train/patch*.tar.gz}{source 2} \\
PixMo-Points & ODC-BY-1.0 & \href{https://huggingface.co/datasets/allenai/pixmo-points}{source} \\
RefCOCO/+/g & MS COCO image licenses; annotations license not clearly specified & \href{https://web.archive.org/web/20220413011718/https://bvisionweb1.cs.unc.edu/licheng/referit/data/refcoco.zip}{source 1}; \href{https://web.archive.org/web/20220413011656/https://bvisionweb1.cs.unc.edu/licheng/referit/data/refcocog.zip}{source 2}; \href{https://web.archive.org/web/20220413012904/https://bvisionweb1.cs.unc.edu/licheng/referit/data/refcoco+.zip}{source 3} \\
TallyQA & Apache-2.0 for TallyQA repo/annotations; referenced VQA 2.0/Visual Genome terms may apply & \href{https://github.com/manoja328/TallyQA_dataset/raw/master/tallyqa.zip}{source} \\
TolokaVQA & CC BY 4.0; images from CC BY-licensed MS COCO subset & \href{https://www.kaggle.com/datasets/dustalov/toloka-wsdm-cup-2023-vqa}{source} \\
V3Det & CC BY 4.0 for annotations/category tree/tools; Flickr/image terms for images & \href{https://huggingface.co/datasets/yhcao/V3Det_Backup}{source} \\
\midrule
\rowcolor{instructcolor}\multicolumn{3}{l}{\textit{Math}} \\
CLEVR-Math & CC BY 4.0 & \href{https://huggingface.co/datasets/dali-does/clevr-math}{source} \\
Geometry3K & Apache-2.0 & \href{https://huggingface.co/datasets/lmms-lab/LLaVA-OneVision-Data/tree/main/Geometry3K(MathV360K)}{source} \\
GeomVerse & Unknown / not publicly specified on The Cauldron repo; original GeomVerse terms may apply & \href{https://huggingface.co/datasets/HuggingFaceM4/the_cauldron/tree/main/geomverse}{source} \\
GeoQA+ & Apache-2.0 & \href{https://huggingface.co/datasets/lmms-lab/LLaVA-OneVision-Data/tree/main/GeoQA+(MathV360K)}{source} \\
MAVIS-Function & Unknown / not publicly specified & \href{https://huggingface.co/datasets/CaraJ/MAVIS-Function}{source} \\
MAVIS-Geometry & Unknown / not publicly specified & \href{https://huggingface.co/datasets/CaraJ/MAVIS-Geometry}{source} \\
UniGeo (Calc.) & Unknown / not publicly specified & \href{https://drive.google.com/drive/folders/1weZAzuxCyNsXK3NoUGo_7FWCAMnG4Mh4}{source} \\
UniGeo (Proof) & Unknown / not publicly specified & \href{https://drive.google.com/drive/folders/1weZAzuxCyNsXK3NoUGo_7FWCAMnG4Mh4}{source} \\
\midrule
\rowcolor{instructcolor}\multicolumn{3}{l}{\textit{Naive OCR}} \\
AnyWord-3M & Apache-2.0 & \href{https://huggingface.co/datasets/stzhao/AnyWord-3M}{source} \\
ArT & Unknown / not publicly specified & \href{https://drive.google.com/file/d/1BH6H75V-vW8G74ih9AXcsiucpqv4eshu/view}{source} \\
CASIA & Other / free for non-commercial use; Kaggle mirror license is other; original CASIA terms apply & \href{https://www.kaggle.com/datasets/pascalbliem/handwritten-chinese-character-hanzi-datasets}{source} \\
Chinese-OCR & Unknown / not publicly specified & \href{https://drive.google.com/file/d/1pM2F04X3itsGi3n-8d7Ppxg6MFqBLKbX/view?usp=sharing}{source} \\
COCO-Text & CC BY 4.0 & \href{https://github.com/bgshih/cocotext/releases/download/dl/cocotext.v2.zip}{source} \\
CTW & Unknown / not publicly specified & \href{https://cloud.tsinghua.edu.cn/d/6f1cfc516c324f41a07b/files/?p=\%2Fctw-public\%2Fctw-annotations.tar.gz&dl=1}{source 1}; \href{https://cloud.tsinghua.edu.cn/d/6f1cfc516c324f41a07b/files/?p=\%2Fctw-public\%2Fimages-trainval\%2Fctw-trainval-*-of-26.tar&dl=1}{source 2} \\
EATEN & Unknown / not publicly specified & \href{https://drive.google.com/file/d/10QPeNGbsCzUOdxESuiA6CROHVx2AkITb/view?usp=sharing}{source} \\
HME-100K & Apache-2.0 & \href{https://huggingface.co/datasets/lmms-lab/LLaVA-OneVision-Data/tree/main/hme100k}{source} \\
IAM & Custom non-commercial/research license & \href{https://huggingface.co/datasets/alpayariyak/IAM_Sentences}{source} \\
LSVT & Unknown / not publicly specified & \href{https://drive.google.com/file/d/18c4u7W_dDacJhe-mABYjleX3CekHb6lN/view}{source} \\
MTWI & Unknown / not publicly specified & \href{https://drive.google.com/file/d/1DDVa81xldAAy9iSAggvNvAIXhCVyBnQo/view?usp=drive_link}{source 1}; \href{https://drive.google.com/file/d/1MNZ3lVZzj6oXBkMKLRb_qdBenWF1s5ew/view?usp=drive_link}{source 2} \\
ParSynth-OCR-200K & Unknown / not publicly specified & \href{https://huggingface.co/datasets/hezarai/parsynth-ocr-200k}{source} \\
POIE & Unknown / not publicly specified & \href{https://drive.google.com/file/d/1eEMNiVeLlD-b08XW_GfAGfPmmII-GDYs/view}{source} \\
ReCTS & Unknown / not publicly specified & \href{https://drive.google.com/file/d/1orMtLhJt3rQl3pMoLm31eh-SmDG74W1K/view}{source} \\
RenderedText & Unknown / not publicly specified & \href{https://huggingface.co/datasets/wendlerc/RenderedText}{source} \\
SROIE-2019 & Unknown / not publicly specified & \href{https://huggingface.co/datasets/priyank-m/SROIE_2019_text_recognition}{source} \\
SynthDoG & MIT for SynthDoG code; generated dataset license not specified on listed HF repos & \href{https://huggingface.co/datasets/naver-clova-ix/synthdog-en}{source 1}; \href{https://huggingface.co/datasets/naver-clova-ix/synthdog-zh}{source 2}; \href{https://huggingface.co/datasets/naver-clova-ix/synthdog-ja}{source 3}; \href{https://huggingface.co/datasets/naver-clova-ix/synthdog-ko}{source 4} \\
SynthText & Custom research-only/non-commercial terms & \href{https://www.kaggle.com/datasets/wassefy/synthtext}{source} \\
\midrule
\rowcolor{instructcolor}\multicolumn{3}{l}{\textit{OCR QA}} \\
ArXivQA & CC BY-SA 4.0 & \href{https://huggingface.co/datasets/MMInstruction/ArxivQA}{source} \\
Docmatix & MIT & \href{https://huggingface.co/datasets/HuggingFaceM4/Docmatix}{source} \\
DocReason25K & Apache-2.0 & \href{https://huggingface.co/datasets/mPLUG/DocReason25K}{source} \\
DocVQA & Apache-2.0 for formatted HF version; original DocVQA terms may also apply & \href{https://huggingface.co/datasets/nyu-visionx/Cambrian-10M/resolve/main/docvqa.tar.gz}{source 1}; \href{https://huggingface.co/datasets/nyu-visionx/Cambrian-10M/resolve/main/jsons/Cambrian10M.jsonl}{source 2} \\
InfoVQA & Apache-2.0 & \href{https://huggingface.co/datasets/LIME-DATA/infovqa}{source} \\
KVQA & Unknown / not publicly specified & \href{https://huggingface.co/datasets/wusize/kvqa}{source} \\
LLaVAR & CC BY-NC 4.0; research-only/non-commercial; CLIP/LLaMA/Vicuna/GPT-4/LLaVA terms may apply & \href{https://drive.google.com/uc?id=1zWpqnAcaG_dUwkJJUvP9FH9zq__c-ODY}{source 1}; \href{https://drive.google.com/uc?id=1Ms7OCjcFQ18Whmujszpc9bTp0Jy0Dye4}{source 2}; \href{https://drive.google.com/uc?id=1_GCHFwrPGjp-9tZlDBwVkdz-L1ymchKY}{source 3}; \href{https://drive.google.com/uc?id=1NHO8lly6pUo-fdyOAyWeGiQJWRb9qggk}{source 4} \\
MapQA & Unknown / not publicly specified on The Cauldron repo; original MapQA terms may apply & \href{https://huggingface.co/datasets/HuggingFaceM4/the_cauldron/tree/main/mapqa}{source} \\
MathWriting & CC BY-NC-SA 4.0 & \href{https://storage.googleapis.com/mathwriting_data/mathwriting-2024.tgz}{source} \\
MultiUI & ODC-BY-1.0; HF-gated contact-sharing/access terms; public-web source content and LLM-provider terms may apply & \href{https://huggingface.co/datasets/neulab/MultiUI}{source} \\
OCR-VQA & Unknown / not publicly specified & \href{https://huggingface.co/datasets/howard-hou/OCR-VQA}{source} \\
Screen2Words & CC BY 4.0 & \href{https://huggingface.co/datasets/rootsautomation/RICO-Screen2Words}{source} \\
ST-VQA & Unknown / not publicly specified & \href{https://huggingface.co/datasets/vikhyatk/st-vqa}{source} \\
TextOCR & CC BY 4.0 & \href{https://dl.fbaipublicfiles.com/textvqa/images/train_val_images.zip}{source 1}; \href{https://dl.fbaipublicfiles.com/textvqa/data/textocr/TextOCR_0.1_train.json}{source 2} \\
TextVQA & Apache-2.0 for Cambrian/LLaVA formatted version; original TextVQA terms may also apply & \href{https://huggingface.co/datasets/nyu-visionx/Cambrian-10M/resolve/main/textvqa.tar.gz}{source 1}; \href{https://huggingface.co/datasets/nyu-visionx/Cambrian-10M/resolve/main/jsons/Cambrian10M.jsonl}{source 2} \\
VisualMRC & Unknown / not publicly specified & \href{https://drive.google.com/file/d/1kxCIG6Nns7L79i5s7xmvKS92foIDRm4T/view}{source} \\
\midrule
\rowcolor{instructcolor}\multicolumn{3}{l}{\textit{Science}} \\
AI2D & Apache-2.0 for Cambrian/LLaVA formatted version; original AI2D terms may also apply & \href{https://huggingface.co/datasets/nyu-visionx/Cambrian-10M/resolve/main/ai2d.tar.gz}{source 1}; \href{https://huggingface.co/datasets/nyu-visionx/Cambrian-10M/resolve/main/jsons/Cambrian10M.jsonl}{source 2} \\
ImageCLEF & Unknown / not publicly specified & \href{https://drive.google.com/file/d/1ZJcd15D699x0WO3ZPS_3CessN2ed17qR/view?usp=drive_link}{source} \\
LLaVA-Med (FT) & CC BY-NC 4.0; research/non-clinical-use restrictions; LLaMA/Vicuna/GPT-4 terms may apply & \href{https://huggingface.co/datasets/Shubhangi29/llava_med_instruct_60k_inline_mention_filtered}{source} \\
LLaVA-Med (PT) & CC BY-NC 4.0; research/non-clinical-use restrictions; LLaMA/Vicuna/GPT-4 terms may apply & \href{https://huggingface.co/datasets/Shubhangi29/llava_med_alignment_500k_chunk_1}{source 1}; \href{https://huggingface.co/datasets/Shubhangi29/llava_med_alignment_500k_chunk_2}{source 2}; \href{https://huggingface.co/datasets/Shubhangi29/llava_med_alignment_500k_chunk_3}{source 3}; \href{https://huggingface.co/datasets/Shubhangi29/llava_med_alignment_500k_chunk_4}{source 4}; \href{https://huggingface.co/datasets/Shubhangi29/llava_med_alignment_500k_chunk_5}{source 5} \\
PathVQA & MIT & \href{https://huggingface.co/datasets/flaviagiammarino/path-vqa}{source} \\
PMC-VQA & CC BY-SA (source PMC OA images/articles CC0 or CC BY) & \href{https://huggingface.co/datasets/xmcmic/PMC-VQA}{source} \\
ScienceQA & CC BY-SA 4.0 & \href{https://huggingface.co/datasets/derek-thomas/ScienceQA}{source} \\
SLAKE & CC BY 4.0 & \href{https://huggingface.co/datasets/BoKelvin/SLAKE}{source} \\
TQA & CC BY-NC 3.0 & \href{https://s3.amazonaws.com/ai2-vision-textbook-dataset/dataset_releases/tqa/tqa_train_val_test.zip}{source} \\
VisualWebInstruct & Apache-2.0 & \href{https://huggingface.co/datasets/TIGER-Lab/VisualWebInstruct}{source} \\
VQA-RAD & CC0-1.0 & \href{https://huggingface.co/datasets/flaviagiammarino/vqa-rad}{source} \\
WebSight & CC BY 4.0 + source-content licenses/disclosure condition & \href{https://huggingface.co/datasets/HuggingFaceM4/WebSight}{source} \\
\midrule
\rowcolor{textcolor}\multicolumn{3}{l}{\textit{Text}} \\
FLAN & CC BY 4.0 (Open-Orca/FLAN HF repo) & \href{https://huggingface.co/datasets/Open-Orca/FLAN}{source} \\
FLAN-v2 & Apache-2.0 & \href{https://huggingface.co/datasets/SirNeural/flan_v2}{source} \\
SlimOrca & MIT & \href{https://huggingface.co/datasets/Open-Orca/SlimOrca}{source} \\
UltraChat-200K & MIT & \href{https://huggingface.co/datasets/HuggingFaceH4/ultrachat_200k}{source} \\
UltraFeedback & MIT & \href{https://huggingface.co/datasets/openbmb/UltraFeedback}{source} \\
WizardLM-Evol-70K & MIT & \href{https://huggingface.co/datasets/WizardLMTeam/WizardLM_evol_instruct_70k}{source} \\
LIMA & Other; source-stricter license if applicable, otherwise CC BY-NC-SA & \href{https://huggingface.co/datasets/GAIR/lima}{source} \\
No Robots & CC BY-NC 4.0 & \href{https://huggingface.co/datasets/HuggingFaceH4/no_robots}{source} \\
Unnatural Instr. & MIT & \href{https://github.com/orhonovich/unnatural-instructions/raw/main/data/core_data.zip}{source} \\
MOSS & CC BY 4.0 & \href{https://huggingface.co/datasets/fnlp/moss-002-sft-data}{source} \\
Llama3-Magpie-Pro & Llama 3 license (HF license tag: llama3) & \href{https://huggingface.co/datasets/Magpie-Align/Llama-3-Magpie-Pro-1M-v0.1}{source} \\
Magpie-Qwen2-Pro & Unknown / not publicly specified on listed HF repo; generated with Qwen2, so Qwen terms may apply & \href{https://huggingface.co/datasets/Magpie-Align/Magpie-Qwen2-Pro-1M-v0.1}{source} \\
Firefly & Unknown / not publicly specified & \href{https://huggingface.co/datasets/YeungNLP/firefly-train-1.1M}{source} \\
Dolly & CC BY-SA 3.0 & \href{https://huggingface.co/datasets/databricks/databricks-dolly-15k}{source} \\
KOpen-Hermes-25 & MIT & \href{https://huggingface.co/datasets/MarkrAI/KOpen-HQ-Hermes-2.5-60K}{source} \\
OpenAI-TLDR & Unknown / source OpenAI TL;DR data terms not clearly specified & \href{https://huggingface.co/datasets/CarperAI/openai_summarize_tldr}{source} \\
Saraswati-CoT & OpenRAIL & \href{https://huggingface.co/datasets/knowrohit07/know-saraswati-cot}{source} \\
CodeFeedback & Apache-2.0 for source m-a-p/Code-Feedback; listed HF formatted repo has no license tag; OpenAI usage policy may apply & \href{https://huggingface.co/datasets/HuggingFaceH4/Code-Feedback}{source} \\
Glaive-Code & Apache-2.0 & \href{https://huggingface.co/datasets/glaiveai/glaive-code-assistant}{source} \\
xCoder-80K & Unknown / not publicly specified & \href{https://huggingface.co/datasets/banksy235/XCoder-80K}{source} \\
LeetCode & Llama 2 license (HF license tag: llama2) & \href{https://huggingface.co/datasets/RayBernard/leetcode}{source} \\
Evol-Code & CC BY-NC-SA 4.0 & \href{https://huggingface.co/datasets/nickrosh/Evol-Instruct-Code-80k-v1}{source} \\
LongCite-45K & Apache-2.0 & \href{https://huggingface.co/datasets/THUDM/LongCite-45k}{source} \\
LongInstruct-Para. & CC BY-SA 4.0 & \href{https://huggingface.co/datasets/yuyijiong/Long-Instruction-with-Paraphrasing}{source} \\
Long-QLoRA & Unknown / no license specified on listed HF repo; source dataset licenses may apply & \href{https://huggingface.co/datasets/YeungNLP/LongQLoRA-Dataset}{source} \\
LongAlpaca & CC BY-NC 4.0 for data/weights; research/non-commercial only & \href{https://huggingface.co/datasets/Yukang/LongAlpaca-12k}{source} \\
GSM8K (Socratic) & MIT & \href{https://huggingface.co/datasets/openai/gsm8k}{source} \\
MetaMathQA & MIT & \href{https://huggingface.co/datasets/meta-math/MetaMathQA}{source} \\
MathQA & Apache-2.0 & \href{https://math-qa.github.io/math-QA/data/MathQA.zip}{source} \\
Numina-Math-1.5 & Apache-2.0 & \href{https://huggingface.co/datasets/AI-MO/NuminaMath-1.5}{source} \\
Numina-Math-TIR & Apache-2.0 & \href{https://huggingface.co/datasets/AI-MO/NuminaMath-TIR}{source} \\
Orca-Math & MIT & \href{https://huggingface.co/datasets/microsoft/orca-math-word-problems-200k}{source} \\
InfinityMath & Apache-2.0 & \href{https://huggingface.co/datasets/BAAI/InfinityMATH}{source} \\
\end{longtable}
\endgroup

\subsection{Sample visualizations}
To enable quick and easy visualization of random samples from each of our 160 source datasets, we provide a visualization tool built with Streamlit here: \href{https://dcvlm-dataset-browser.streamlit.app/}{https://dcvlm-dataset-browser.streamlit.app/}.
The tool enables browsing samples from datasets, categorized by data type and afforded capability (for multimodal instruction tuning datasets).

\subsection{Multilingual Nature}\label{app:multilingual}

As can be seen from~\cref{tab:dcvlm-datamix-licenses}, the \dcvlm{} pool is sourced primarily from English-centric web corpora (DataComp~\citep{gadre2023datacomp}, ReLAION-2B-en~\citep{schuhmann2022laion}) and English instruction-tuning datasets, but it inherits a non-trivial multilingual tail from datasets that target other languages explicitly (e.g.\ Chinese instruction corpora, multilingual OCR generators) and from the natural multilingual contamination present in any large web crawl. To quantify this, we annotate every sample in the pool with two independent language detectors and take their top-1 prediction.

\paragraph{Annotation methodology.} For each sample we run two language detectors over the per-sample text (caption, instruction, or document content):
\begin{itemize}[leftmargin=15pt,topsep=2pt,itemsep=2pt]
  \item \textbf{Lingua}\footnote{\url{https://github.com/pemistahl/lingua-py}}: a lightweight character-$n$-gram detector covering 75 languages with good coverage of European tongues but limited support for languages with non-Latin scripts.
  \item \textbf{NLLB-218}~\citep{costa2022no}: the multilingual classifier shipped with the No-Language-Left-Behind translation model, covering 218 languages and substantially better than Lingua on Cyrillic, Devanagari, and Southeast-Asian scripts.
\end{itemize}
Each annotation stores the top-3 language hypotheses with calibrated scores. For the analysis below we report only the top-1 prediction from each detector. Both detectors operate on text with images stripped, so they describe the linguistic distribution of the \emph{textual} content of the pool, not of any visual content.

\paragraph{Pool-wide distribution.} Figure~\ref{fig:lang-distribution} shows the top-15 languages by sample count under each detector, over our small pool (each dataset is capped at 100{,}000 samples as a representative subset). We notice that the distribution is heavily tailed---English alone accounts for $91.1\%$ (Lingua) / $92.8\%$ (NLLB) of all samples. The single largest non-English language is Chinese at $3.9\%$ / $3.5\%$. Beyond Chinese, the remaining languages are spread thinly: French, German, Japanese, Korean, and Russian each contribute $0.2\%$--$0.8\%$, and the rest of the long tail (a further 70-110 languages depending on detector) sums to under $1\%$. Despite the long tail Lingua and NLLB largely agree---the two detectors return the same top-1 language on $96.6\%$ of samples---so our statistics are robust to the choice of detector.

\figlangdistribution

\paragraph{Where does the non-English content come from?} Roughly half of the pool's non-English samples trace to a small number of datasets that were either built for a specific non-English language or rely on synthetic multilingual generation:
\begin{itemize}[leftmargin=15pt,topsep=2pt,itemsep=2pt]
  \item \textbf{Chinese:} \texttt{firefly} (a general-purpose Chinese instruction corpus, $\sim$91\% non-English, contributing $\sim$33\% of all Chinese samples), \texttt{wanjuan} (curated Chinese multimodal interleaved documents, $\sim$84\% non-English), \texttt{internvl\_sa1b\_caption} (Chinese-translated SA-1B captions), and \texttt{magpie\_qwen2\_pro} (Chinese instruction synthesis).
  \item \textbf{French / German / Russian:} dominated by \texttt{flanv2}, whose translation tasks bundle large numbers of parallel corpora---alone responsible for $72.4\%$ of the pool's French, $78.3\%$ of its German, and $64.6\%$ of its Russian samples.
  \item \textbf{Japanese / Korean:} \texttt{synthdog}, a synthetic OCR-style image-text generator, supplies $76\%$ of Japanese and $88\%$ of Korean samples; the remaining mass comes from \texttt{eaten}, \texttt{mavis\_geometry}, and small instruction sets such as \texttt{kopen-hq-hermes-25-60K} ($100\%$ non-English).
  \item \textbf{Vietnamese / Arabic / Indonesian:} largely supplied by \texttt{sea-vl}, a Southeast-Asian vision-language dataset ($93\%$ Vietnamese, with smaller contributions to Arabic and Indonesian).
\end{itemize}
The remaining non-English content is the natural multilingual residue of the web crawls---although \texttt{relaion-2b-en} and \texttt{datacomp\_1b} are English-filtered upstream, $\sim$5\% of their captions are still flagged as non-English by NLLB, predominantly European languages (German, French, Italian, Dutch). This also corroborates the findings from~\citet{nguyen2024multilingual} and~\citet{pouget2024no}.

\paragraph{Implications.} The pool is overwhelmingly English by design: this matches the practice of most recent open VLM pretraining corpus we are aware of and reflects the asymmetry of available web data~\citep{udandarao2024no,parashar2024neglected}. The Chinese sub-pool is large enough to plausibly support Chinese-capable downstream models, as we show in our experiments in~\cref{sec:experiments}. Languages outside the top five are too sparse to sustain meaningful pretraining signal on their own---this is a limitation of the underlying source data rather than of \dcvlm{} per se, and addressing it would require deliberate ingestion of non-English vision-language datasets.

\paragraph{How much does English data drive downstream performance?} To test how dominant the English pretraining data is in our pool, we run a set of language ablation experiments at the \texttt{medium} scale (2B model, 25B tokens). Using the per-sample NLLB top-1 language annotations described above, we partition the pool into English and non-English samples and construct three training sets, holding the data mixture fixed: (\textit{i}) \textbf{All languages}, our default pool with no language-based selection; (\textit{ii}) \textbf{English-only}, which retains only samples whose detected language is English; and (\textit{iii}) \textbf{Non-English-only}, which retains only the complementary non-English tail. Since English accounts for ${\sim}92\%$ of the pool (\cref{fig:lang-distribution}), the English-only set is close in size to the full pool, whereas the non-English-only set is far smaller and is consequently seen with more repetitions to fill the same 25B-token budget. We report core-set results in \cref{tab:lang-ablation}.

\begin{table}[h]
    \centering
    \caption{\textbf{Language ablation experiments.} We partition the \dcvlm{} pool by language and train on (\textit{i}) all languages, (\textit{ii}) English-only samples, and (\textit{iii}) non-English-only samples. The English-only model nearly matches the full pool, while training only on the non-English tail is substantially worse. Multilingual benchmark performance is essentially unchanged across all three pools.}
    \label{tab:lang-ablation}
    \small
    \begin{tabular}{lccccccc}
        \toprule
        \textbf{Training pool} & \textbf{Gen} & \textbf{Know} & \textbf{OCR} & \textbf{Vision} & \textbf{MTL} & \textbf{Text} & \textbf{Core Avg} \\
        \midrule
        All languages (no filter) & 58.5 & 58.4 & 41.8 & 44.3 & 39.8 & 48.2 & \textbf{49.1} \\
        English-only               & 57.3 & 58.5 & 41.5 & 41.9 & 39.9 & 48.4 & 48.5 \\
        Non-English-only           & 53.2 & 56.6 & 38.5 & 42.0 & 39.1 & 45.4 & 46.1 \\
        \bottomrule
    \end{tabular}
\end{table}

Three observations stand out. First, \textbf{English data carries almost all of the signal}: the English-only model trails the full pool by only $0.6$pp, despite discarding ${\sim}8\%$ of the pool. Second, \textbf{the non-English tail alone is insufficient}: training exclusively on non-English samples drops performance by $3.0$pp relative to the full pool, with the largest regressions on General ($-5.3$pp), OCR ($-3.3$pp), and Text-Only ($-2.8$pp). Third, and most notably, \textbf{multilingual benchmark performance is essentially flat across all three pools} ($39.8$ / $39.9$ / $39.1$): neither removing non-English data nor training solely on it meaningfully moves the Multilingual score. 

We caution, however, that two confounders complicate the interpretation of this experiment: (\textit{i})
  since English makes up ${\sim}92\%$ of the pool, the non-English-only run draws from far fewer unique
  samples and incurs substantially more repetition than the English-only run to fill the same token budget, so
  its performance drop cannot be cleanly attributed to language as opposed to repetition; and (\textit{ii})
  our per-sample language annotations are noisy---especially for multilingual samples, where the detector is
  biased toward the head (high-resource) languages---making the English/non-English partition imperfect.

\clearpage

\section{Evaluation Suite Details}\label{app:eval_details}

In this section, we highlight our protocol to select and categorize benchmarks for evaluating models. 
We start with a candidate pool of 65 benchmarks, which we categorize according to the majority consensus of previous work: Cambrian-1 \cite{tong2024cambrian}, InternVL2.5 \& InternVL-3 \cite{chen2024expanding, zhu2025internvl3}, MM-1 \cite{mckinzie2024mm1}, Qwen2-VL \& Qwen2.5-VL \cite{wang2024qwen2, bai2025qwen25vltechnicalreport} into 9 distinct categories:
\begin{itemize}[leftmargin=*,itemsep=1pt]
    \item General Understanding: this category includes established multimodal benchmarks, as well as comprehensive, all-round benchmarks assessing several capabilities at once. 
    \item Knowledge-Centric benchmarks: these test factual, scientific and/or educational knowledge. 
    \item OCR \& Charts, testing the ability of models to read and understand visually rendered characters, diagrams, layout, and PDFs. 
    \item Vision-Centric benchmarks, requiring models to perceive spatially, perform fine-grained visual processing such as counting, and compare items across multiple images. 
    \item Multilingual benchmarks, which assess general knowledge in less common languages. 
    \item Text-Only benchmarks, testing whether the unimodal, language-only capabilities of the language model are preserved; 
    \item Safety benchmarks, which measure the ability of models to detect and, potentially, refuse to respond to, unsafe queries (e.g., violent, harmful and/or explicit content). 
    \item Hallucination benchmarks, stress-testing the capabilities of models to assess evident facts such as the existence of explicit objects in images; 
    \item Reasoning benchmarks, with a math- and logic-centric focus (including visual tasks such as rendered geometry problems and language-only tasks such as GSM-8K \cite{cobbe2021gsm8k}). 
\end{itemize}

Prior works are far from unanimous in how they organize these benchmarks: the same task is frequently placed in different categories by different works. For instance, AI2D is treated as a Knowledge benchmark by Cambrian-1 but as an OCR/document task by Qwen2-VL and InternVL, and MMMU is variously labelled as Knowledge, Mathematics, or Reasoning.
To obtain a single consistent taxonomy, we collect the category each prior work assigns to every benchmark and resolve disagreements by majority vote wherever a consensus exists, adjudicating the remaining ambiguous cases ourselves.
\Cref{tab:benchmark-categorization} reports, for each benchmark, the verbatim category given by every prior work alongside our final \dcvlm assignment. \Cref{tab:benchmarks} then lists the resulting mapping from each \dcvlm category to its benchmarks.
Whenever possible, we use benchmark implementations in VLMEvalKit \cite{duan2024vlmevalkit} and OpenCompass \cite{2023opencompass}, which we integrate into the toolkit we plan to release.

\begin{footnotesize}
\setlength{\tabcolsep}{3pt}
\renewcommand{\arraystretch}{1.15}
\begin{longtable}{@{}l *{5}{>{\raggedright\arraybackslash}p{1.95cm}}@{}}
\caption{\textbf{Benchmark categorization across prior works versus \dcvlm.} For each benchmark in our candidate pool we show the category assigned by Cambrian-1~\citep{tong2024cambrian}, MM-1~\citep{mckinzie2024mm1}, Qwen2-VL/Qwen2.5-VL~\citep{wang2024qwen2,bai2025qwen25vltechnicalreport}, and InternVL2.5/InternVL-3~\citep{chen2024expanding,zhu2025internvl3}, each \emph{verbatim} as reported by that work. Rows are grouped by the unified \dcvlm category. Prior works frequently disagree (e.g.\ AI2D is placed in Knowledge, OCR, or Diagram understanding by different works). We resolve each benchmark by majority consensus where one exists and adjudicate ambiguous cases ourselves. ``--'' marks a benchmark not categorized (or not used) by that work. 
} 
\\
\label{tab:benchmark-categorization}\\
\toprule
\textbf{Benchmark} & \textbf{Cambrian-1} & \textbf{MM-1} & \textbf{Qwen2/2.5-VL} & \textbf{InternVL2.5} & \textbf{InternVL-3} \\
\midrule
\endfirsthead
\toprule
\textbf{Benchmark} & \textbf{Cambrian-1} & \textbf{MM-1} & \textbf{Qwen2/2.5-VL} & \textbf{InternVL2.5} & \textbf{InternVL-3} \\
\midrule
\endhead
\bottomrule
\multicolumn{6}{r}{\footnotesize\itshape continued on next page}\\
\endfoot
\bottomrule
\endlastfoot
\addlinespace[2pt]
\multicolumn{6}{@{}l}{\cellcolor{gray!12}\textbf{General Understanding}}\\
\addlinespace[1pt]
MM-Bench & General & -- & General VQA & General & General \\
MM-Bench v1.1 & General & -- & General VQA & General & General \\
MME & General & -- & General VQA & General & General \\
GQA & General & -- & -- & -- & -- \\
SEEDBench-IMG & General & -- & -- & -- & -- \\
MMT-Bench & -- & -- & General VQA & -- & -- \\
MMStar & -- & -- & General VQA & General & General \\
COCO-Caption & -- & -- & -- & -- & -- \\
\addlinespace[2pt]
\multicolumn{6}{@{}l}{\cellcolor{gray!12}\textbf{Knowledge}}\\
\addlinespace[1pt]
MMMU & Knowledge & -- & Maths, Edu & Reasoning \& Maths & Reasoning \& Maths \\
ScienceQA & Knowledge & -- & -- & -- & -- \\
AI2D & Knowledge & -- & OCR, Docs, Diagram & OCR, Charts \& Docs & OCR, Charts \& Docs \\
\addlinespace[2pt]
\multicolumn{6}{@{}l}{\cellcolor{gray!12}\textbf{OCR \& Charts}}\\
\addlinespace[1pt]
CharXiv & -- & -- & -- & OCR, Charts \& Docs & OCR, Charts \& Docs \\
DocVQA & OCR \& Charts & -- & OCR, Docs, Diagram & OCR, Charts \& Docs & OCR, Charts \& Docs \\
ChartQA & OCR \& Charts & -- & OCR, Docs, Diagram & OCR, Charts \& Docs & OCR, Charts \& Docs \\
OCRBench & OCR \& Charts & -- & OCR, Docs, Diagram & OCR, Charts \& Docs & OCR, Charts \& Docs \\
TextVQA & OCR \& Charts & -- & OCR, Docs, Diagram & OCR, Charts \& Docs & OCR, Charts \& Docs \\
InfoVQA & -- & -- & OCR, Docs, Diagram & OCR, Charts \& Docs & OCR, Charts \& Docs \\
SEEDBench2+ & -- & -- & -- & OCR, Charts \& Docs & OCR, Charts \& Docs \\
OCRVQA & -- & -- & -- & -- & -- \\
\addlinespace[2pt]
\multicolumn{6}{@{}l}{\cellcolor{gray!12}\textbf{Vision-Centric}}\\
\addlinespace[1pt]
CV-Bench & Vision-Centric & -- & -- & -- & -- \\
Mantis-Eval & -- & -- & -- & Multi-Image Understanding & Multi-Image Understanding \\
3DSRBench & -- & -- & -- & -- & -- \\
MMT-Bench-MI & -- & -- & -- & Multi-Image Understanding & Multi-Image Understanding \\
BLINK & -- & -- & -- & Multi-Image Understanding & Multi-Image Understanding \\
MuirBench & -- & -- & -- & Multi-Image Understanding & Multi-Image Understanding \\
NaturalBench & -- & -- & -- & -- & -- \\
MMVP & Vision-Centric & -- & -- & -- & -- \\
RealWorldQA & Vision-Centric & -- & General VQA & Real-World Comprehension & Real-World Comprehension \\
MME-RealWorld & -- & -- & -- & Real-World Comprehension & Real-World Comprehension \\
VizWiz & -- & -- & -- & -- & -- \\
\addlinespace[2pt]
\multicolumn{6}{@{}l}{\cellcolor{gray!12}\textbf{Multilingual}}\\
\addlinespace[1pt]
MMMB & -- & -- & -- & Multilingual & Multilingual \\
MTL-MMBench & -- & -- & -- & Multilingual & Multilingual \\
MTVQA & -- & -- & OCR, Docs, Diagram & Multilingual & Multilingual \\
\addlinespace[2pt]
\multicolumn{6}{@{}l}{\cellcolor{gray!12}\textbf{Reasoning}}\\
\addlinespace[1pt]
MathVista & Knowledge & -- & Maths, Edu & Reasoning \& Maths & Reasoning \& Maths \\
MathVision & -- & -- & Maths, Edu & Reasoning \& Maths & Reasoning \& Maths \\
MathVerse & -- & -- & -- & Reasoning \& Maths & Reasoning \& Maths \\
WeMath & -- & -- & -- & -- & Reasoning \& Maths \\
DynaMath & -- & -- & -- & -- & Reasoning \& Maths \\
GSM8K & -- & -- & Maths & Maths & Maths \\
MATH & -- & -- & Maths & Maths & Maths \\
TheoremQA & -- & -- & -- & Maths & Maths \\
\addlinespace[2pt]
\multicolumn{6}{@{}l}{\cellcolor{gray!12}\textbf{Text-Only}}\\
\addlinespace[1pt]
GaoKao & -- & -- & -- & Comprehensive & Comprehensive \\
C-EVAL & -- & -- & -- & Comprehensive & Comprehensive \\
MMLU & -- & -- & -- & Comprehensive & Comprehensive \\
CMMLU & -- & -- & -- & Comprehensive & Comprehensive \\
TriviaQA & -- & General & -- & General & General \\
NaturalQ & -- & -- & -- & General & General \\
C3 & -- & -- & -- & General & General \\
RACE & -- & -- & -- & General & General \\
Winogrande & -- & General & -- & Reasoning & Reasoning \\
Hellaswag & -- & General & -- & Reasoning & Reasoning \\
HumanEval & -- & -- & Coding & Coding & Coding \\
\addlinespace[2pt]
\multicolumn{6}{@{}l}{\cellcolor{gray!12}\textbf{Hallucination}}\\
\addlinespace[1pt]
AMBER & -- & -- & -- & -- & -- \\
HallusionBench & -- & -- & General VQA & Hallucination & Hallucination \\
CRPE & -- & -- & -- & Hallucination & Hallucination \\
POPE & -- & -- & -- & Hallucination & Hallucination \\
\addlinespace[2pt]
\multicolumn{6}{@{}l}{\cellcolor{gray!12}\textbf{Safety}}\\
\addlinespace[1pt]
MMSafetyBench & -- & -- & -- & -- & -- \\
MLLMGuard & -- & -- & -- & -- & -- \\
VLGuard & -- & -- & -- & -- & -- \\
\end{longtable}
\end{footnotesize}

\begin{figure}[!h]
    \centering
    \includegraphics[width=0.8\linewidth]{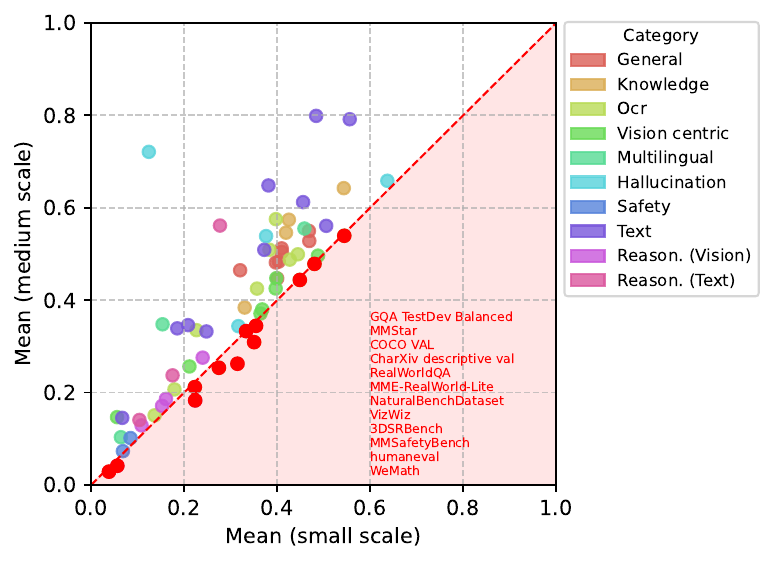}
    \caption{\textbf{Not all benchmarks monotonically increase with scale.} 
    We report the performance of 65 benchmarks (averaged over three runs at both the small and the medium scales) with a focus on their scaling behaviour. Average performance at the \texttt{small} scale is arranged on the x-axis; the y-axis displays average performance at the \texttt{medium} scale.
    Benchmarks falling below the identity line (highlighted in red in the figure) do \textit{not} improve from the \texttt{small} to the \texttt{medium} scale.}
    \label{fig:monotonic-evals}
\end{figure}

\begin{figure}
    \centering
    \includegraphics[width=\linewidth]{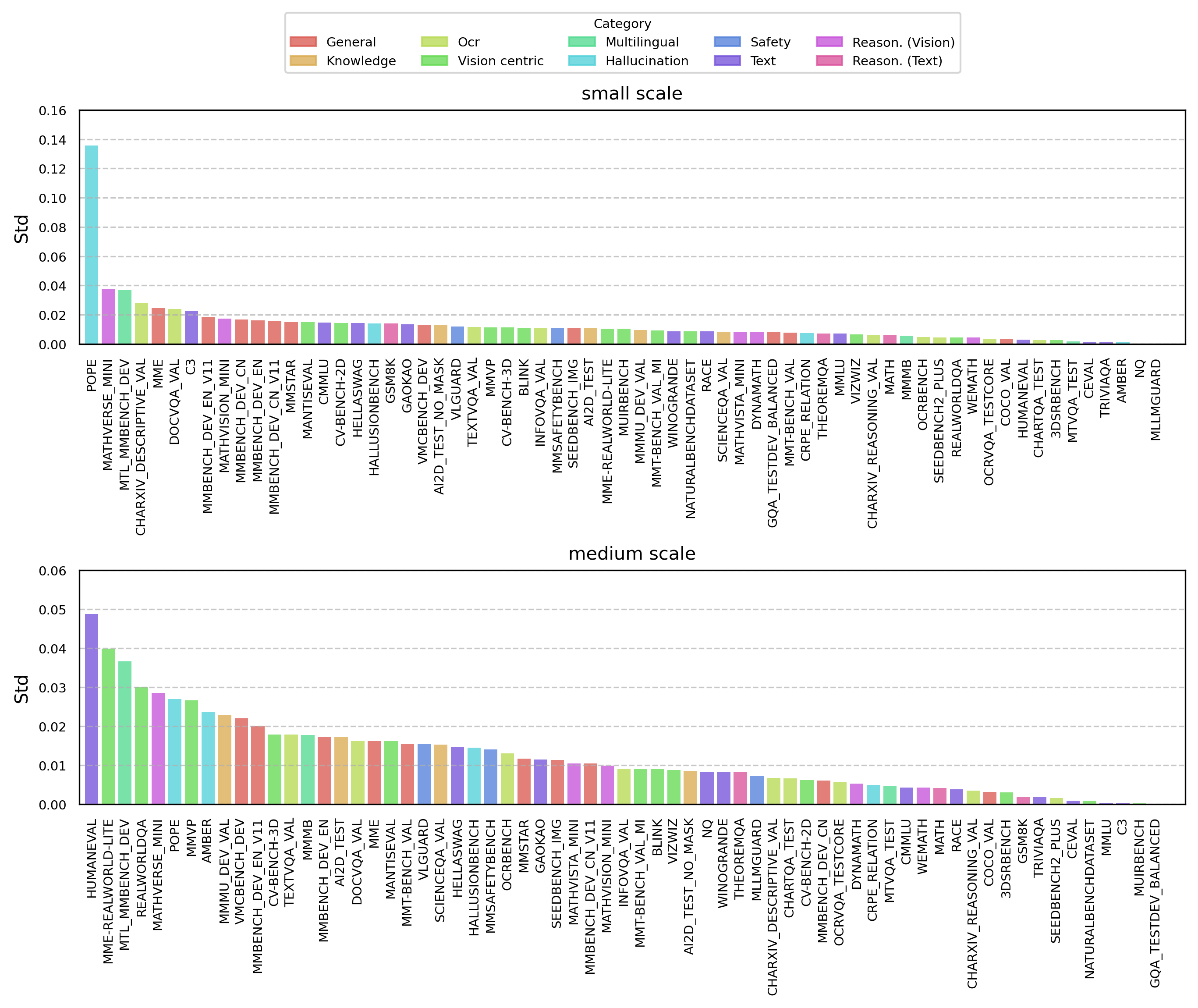}
    \caption{\textbf{Seed variance of the initial 65-benchmark pool.} We conduct runs at \texttt{small} and \texttt{medium} scales, to measure the seed variance of benchmarks in our initial 65 candidates. We remove POPE \cite{li2023evaluating}, as it leads to a standard deviation of up to 16\% over three runs at the small scale.}
    \label{fig:eval-stability}
\end{figure}

\textbf{Monotonicity selection}.
We filter 12 out of 65 benchmarks because they do not pass our first selection criterion: \textit{monotonicity}. 
To ensure reliable evaluation of data-centric decisions across scales, we filter our initial benchmark pool to retain only tasks that exhibit \textit{consistent improvement} with scale. 
Concretely, we train models at two scales (\texttt{small} and \texttt{medium}) for three different random seeds, for a total of six runs. We evaluate all resulting checkpoints and compute the per-scale average.
Since \dcvlm is a scale-centric benchmark, we rule out benchmarks for which performance does \textit{not} monotonically improve from small to medium scale, as such benchmarks would risk obscuring the signal attributable to data curation choices. 
This yields a curated subset of benchmarks (shown in \cref{fig:monotonic-evals} as points above the diagonal) that exhibit reliable scale-monotonic behavior. 

\textbf{Stability selection.} In parallel to assessing monotonicity, we also search for stable evaluations. 
For this reason, we further inspect the \textit{seed variance} of benchmarks using the same six runs. 
We perform stability filtering in two steps: globally and locally within each category. 

To provide an initial global picture, we display the standard deviation for all the initial 65 benchmarks at both scales in \cref{fig:eval-stability}.
We find striking instability with POPE \cite{li2023evaluating} compared to all other benchmarks ($16\%$ seed variance over three runs at the \texttt{small} scale), and we therefore discard it from our pool. 
This decreases the total from 53 to 52 remaining benchmarks, which we jointly call our \textbf{Extended} set. 

\begin{table}[h]
\centering
\caption{The initial unfiltered pool of 65 benchmarks we started from.}
\label{tab:benchmarks}
\resizebox{\textwidth}{!}{%
\begin{tabular}{ll}
\toprule
\textbf{Category} & \textbf{Benchmarks} \\
\midrule
General &
\begin{tabular}[t]{@{}l@{}}
MMBench-DEV-EN, MMBench-DEV-CN, MMBench-DEV-EN-V11, MMBench-DEV-CN-V11 \cite{liu2024mmbench}, MME \cite{fu2023mme},\\
GQA-TestDev-Balanced \cite{hudson2019gqa}, SEEDBench-IMG \cite{li2023seed}, MMT-Bench-VAL \cite{mmtbench}, MMStar \cite{chenwe}, COCO-VAL \cite{chen2015cococaptions}, VMCBench-DEV \cite{AutoConverter}
\end{tabular} \\
\midrule
Knowledge & MMMU-DEV-VAL \cite{yue2024mmmu}, ScienceQA-VAL \cite{lu2022learn}, AI2D-TEST, AI2D-TEST-NO-MASK \cite{kembhavi2016ai2d}\\
\midrule
OCR &
\begin{tabular}[t]{@{}l@{}}
DocVQA-VAL \cite{mathew2021docvqa}, ChartQA-TEST \cite{masry2022chartqa}, OCRBench \cite{liu2024ocrbench}, TextVQA-VAL \cite{singh2019towards}, InfoVQA-VAL \cite{mathew2022infographicvqa},\\
SEEDBench2-Plus \cite{li2024seed}, OCRVQA-TESTCORE \cite{mishra2019ocrvqa}, CharXiv-Descriptive-VAL, CharXiv-Reasoning-VAL \cite{wang2024charxiv}
\end{tabular} \\
\midrule
Vision-Centric &
\begin{tabular}[t]{@{}l@{}}
MMT-Bench-VAL-MI \cite{mmtbench}, BLINK \cite{fu2024blink}, MUIRBench \cite{wang2024muirbench}, RealWorldQA \cite{realworldqa2024}, MME-RealWorld-Lite \cite{zhang2024mme},\\
NaturalBenchDataset \cite{li2024naturalbench}, VizWiz \cite{gurari2018vizwiz}, 3DSRBench \cite{ma20253dsrbench}, MantisEval \cite{jiang2024mantis}, MMVP \cite{tong2024eyes}, CV-Bench-2D, CV-Bench-3D \cite{tong2024cambrian}
\end{tabular} \\
\midrule
Multilingual & MTVQA-TEST\cite{tang2025mtvqa}, MMMB\cite{sun2024parrot}, MTL-MMBench-DEV\cite{liu2024mmbench} \\
\midrule
Hallucination & HallusionBench\cite{guan2024hallusionbench}, CRPE-Relation\cite{wang2024allseeingv2}, POPE \cite{li2023evaluating}, AMBER\cite{wang2023amber} \\
\midrule
Safety & VLGuard\cite{zong2024safety}, MLLMGuard\cite{gu2024mllmguard}, MMSafetyBench\cite{liu2024mm} \\
\midrule
Text &
\begin{tabular}[t]{@{}l@{}}
C3\cite{xu2020clue}, C-Eval\cite{huang2023c}, CMMLU\cite{li2024cmmlu}, GaoKao\cite{zhang2023evaluating}, HellaSwag\cite{zellers2019hellaswag}, HumanEval\cite{chen2021evaluating},\\
MMLU\cite{hendrycks2020measuring}, NQ\cite{kwiatkowski2019natural}, RACE\cite{lai2017race}, TriviaQA\cite{joshi2017triviaqa}, WinoGrande\cite{sakaguchi2021winogrande}
\end{tabular} \\
\midrule
Reasoning & MathVista-MINI\cite{lu2023mathvista}, MathVision-MINI\cite{wang2024measuring}, MathVerse-MINI\cite{zhang2024mathverse}, WeMath\cite{qiao2025we}, DynaMath\cite{zou2024dynamath}, GSM8K\cite{cobbe2021training}, TheoremQA\cite{chen2023theoremqa}, MATH\cite{hendrycks2021measuring} \\
\bottomrule
\end{tabular}%
}
\end{table}

\begin{table}[h]
\centering
\caption{List of benchmarks in our 52-task \textbf{Extended} suite.}
\label{tab:extended-benchmarks}
\resizebox{\textwidth}{!}{%
\begin{tabular}{ll}
\toprule
\textbf{Category} & \textbf{Benchmarks} \\
\midrule
General &
\begin{tabular}[t]{@{}l@{}}
MMBench-DEV-EN, MMBench-DEV-CN, MMBench-DEV-EN-V11, MMBench-DEV-CN-V11,\\
MME, SEEDBench-IMG, MMT-Bench-VAL, VMCBench-DEV
\end{tabular} \\
\midrule
Knowledge & MMMU-DEV-VAL, ScienceQA-VAL, AI2D-TEST, AI2D-TEST-NO-MASK \\
\midrule
OCR &
\begin{tabular}[t]{@{}l@{}}
DocVQA-VAL, ChartQA-TEST, OCRBench, TextVQA-VAL, InfoVQA-VAL,\\
SEEDBench2-Plus, OCRVQA-TESTCORE, CharXiv-Reasoning-VAL
\end{tabular} \\
\midrule
Vision-Centric & MMT-Bench-VAL-MI, BLINK, MUIRBench, MantisEval, MMVP, CV-Bench-2D, CV-Bench-3D \\
\midrule
Multilingual & MTVQA-TEST, MMMB, MTL-MMBench-DEV \\
\midrule
Hallucination & HallusionBench, CRPE-Relation, AMBER \\
\midrule
Safety & VLGuard, MLLMGuard \\
\midrule
Text &
\begin{tabular}[t]{@{}l@{}}
C3, C-Eval, CMMLU, GaoKao, HellaSwag, MMLU, NQ, RACE, TriviaQA, WinoGrande
\end{tabular} \\
\midrule
Reasoning & 
\begin{tabular}[t]{@{}l@{}}
MathVista-MINI, MathVision-MINI, MathVerse-MINI, DynaMath, GSM8K, TheoremQA, MATH
\end{tabular} \\
\bottomrule
\end{tabular}%
}
\end{table}

\begin{table}[h]
\centering
\caption{List of benchmarks in our 33-task \textbf{Core} suite.}
\label{tab:core-benchmarks}
\resizebox{\textwidth}{!}{%
\begin{tabular}{ll}
\toprule
\textbf{Category} & \textbf{Benchmarks} \\
\midrule
General &
\begin{tabular}[t]{@{}l@{}}
MMBench-DEV-EN, MMBench-DEV-CN, MMBench-DEV-EN-V11, MMBench-DEV-CN-V11,\\
SEEDBench-IMG, MMT-Bench-VAL, VMCBench-DEV
\end{tabular} \\
\midrule
Knowledge & MMMU-DEV-VAL, ScienceQA-VAL, AI2D-TEST, AI2D-TEST-NO-MASK \\
\midrule
OCR & ChartQA-TEST, OCRBench, InfoVQA-VAL, SEEDBench2-Plus, OCRVQA-TESTCORE, CharXiv-Reasoning-VAL \\
\midrule
Vision-Centric & MMT-Bench-VAL-MI, BLINK, MUIRBench, CV-Bench-2D, CV-Bench-3D \\
\midrule
Multilingual & MTVQA-TEST, MMMB, MTL-MMBench-DEV \\
\midrule
Text & C-Eval, CMMLU, GaoKao, MMLU, NQ, RACE, TriviaQA, WinoGrande \\
\bottomrule
\end{tabular}%
}
\end{table}

\begin{table}[h]
\footnotesize
\centering
\caption{List of benchmarks in our 13-task \textbf{Validation} suite.}
\label{tab:validation-benchmarks}
\begin{tabular}{ll}
\toprule
\textbf{Category} & \textbf{Benchmarks} \\
\midrule
General     & MMBench-DEV-EN, MMBench-DEV-CN, VMCBench-DEV \\
Knowledge   & MMMU-DEV-VAL, AI2D-TEST \\
OCR         & OCRBench, SEEDBench2-Plus \\
Vision-Centric & BLINK, CV-Bench-2D, CV-Bench-3D \\
Multilingual & MMMB \\
Text        & C-Eval, MMLU \\
\bottomrule
\end{tabular}
\end{table}

\textbf{Extended, Core, and Validation suites.} Since we are primarily interested in evaluating pretrained models (according to our definition in \cref{sec:intro}), we create a smaller subset of tasks by excluding verticals that are mostly of interest for post-training strategies than they are for pretraining \cite{zong2024safety, guha2025openthoughts}: Safety, Hallucination, and Reasoning benchmarks. 
We then apply a second iteration of stability filtering that removes the most unreliable tasks per category according to the average of their standard deviation across small and medium scale runs. 
This corresponds to removing: (i) MME \cite{fu2023mme} from the General category, (ii) 
TextVQA and DocVQA from the OCR \& Charts category, (iii) C3 and HellaSwag from the Text-Only category, (iv) MMVP and MantisEval from the Vision-Centric category. 
We do not remove additional tasks from the Knowledge-Centric and the Multilingual categories, as they already contain a more modest number of tasks compared to other categories. 
This additional selection yields a 33-task suite, which we call our \textbf{Core} set.
As highlighted in the main body, we use this set to report all major results.
We additionally create a 13-task \textbf{Validation} set, which we use for faster experimentation, by picking the most established subset of benchmarks per category. 

To enumerate benchmarks and illustrate tier organization, we list Extended benchmarks in \cref{tab:extended-benchmarks}, Core in \cref{tab:core-benchmarks}, and Validation in \cref{tab:validation-benchmarks}.

\subsection{Issues with Grounding Benchmarks}
\label{app:grounding-detection-issues}

We initially considered including grounding and dense recognition benchmarks, such as
RefCOCO~\citep{kazemzadeh2014referitgame}, RefCOCO+~\citep{yu2016modeling},
RefCOCOg~\citep{nagaraja2016modeling}, and related COCO-derived tasks, in the
\dcvlm evaluation suite.
We ultimately excluded them, not because grounding is unimportant, but because the
standard grounding benchmarks introduce unusually severe contamination and split
ambiguities. These issues make it difficult to interpret them as reliable held-out
evaluations in a data-centric benchmark at the scale of \dcvlm.

The main issue is that many grounding, VQA, and captioning benchmarks are built on
overlapping subsets of MS COCO images~\citep{lin2014microsoft}. The RefCOCO family
uses COCO images, as do COCO Captions~\citep{chen2015cococaptions,karpathy2015deep},
OK-VQA~\citep{marino2019okvqa}, VSR~\citep{liu2023vsr}, and
TolokaVQA~\citep{ustalov2023tolokavqa}. As a result, the same underlying image can
appear as a training example for one task and as an evaluation example for another,
possibly with a different annotation format, e.g., a caption, a VQA pair, or a
referring expression. This makes image-level decontamination across the full training
pool difficult, especially when downstream datasets rename files, construct custom
splits, or mix COCO2014 and COCO2017 images in non-standard ways.

The RefCOCO variants further complicate this picture because they are not independent
held-out datasets. RefCOCO and RefCOCO+ were collected through ReferItGame and
commonly use the \texttt{unc} split, where train/test separation is performed at the
image level~\citep{kazemzadeh2014referitgame,yu2016modeling}. RefCOCOg was collected
through AMT and commonly uses the Google split, where the split is defined at the
object or annotation level~\citep{nagaraja2016modeling}. The variants also differ in
their linguistic supervision: RefCOCO contains relatively short expressions, RefCOCOg
contains longer and more detailed expressions, and RefCOCO+ restricts expressions to
appearance-based descriptions by excluding certain spatial phrases such as ``person on
the right''~\citep{yu2016modeling,nagaraja2016modeling}. Thus, a model trained on one
variant may have already seen the same image-object pair that appears in another
variant, only paired with a different referring expression.

This concern is amplified by common VLM pretraining practices. In several open training
recipes, grounding data is not treated as a single isolated benchmark split, but appears to group annotations from RefCOCO and RefCOCOg under the same
image~\footnote{\url{https://internvl.readthedocs.io/en/latest/get_started/chat_data_format.html\#grounding-detection-data}}. In such cases, evaluation on a nominally separate
RefCOCO-style split may partially measure familiarity with recurring COCO
image-object pairs rather than genuine grounding generalization.

For these reasons, we exclude grounding benchmarks from the main \dcvlm evaluation
suite. Including them would make it difficult to attribute performance differences
cleanly to data mixing, filtering and other curation choices, since gains could arise from uncontrolled overlap between COCO-derived training and evaluation data.
\clearpage

\section{Train-Test Decontamination}\label{app:decontamination}
To ensure that conclusions from our data-centric experiments are not confounded by train-test overlap, we perform strict two-way decontamination of all training sources against the \dcvlm-Extended evaluation suite.
We apply separate procedures for multimodal and text-only samples.

\subsection{Image-Based Decontamination}\label{appsub:image-decont}
For the multimodal portion of our pool, we decontaminate at the image level using the self-supervised copy-detection descriptor SSCD model~\citep{pizzi2022self} (ResNet-50 backbone). We embed every image in both the training pool and the \dcvlm-Extended evaluation suite under an \emph{identical} transform: we resize the shorter edge to $288$px while preserving aspect ratio, then apply ImageNet normalization. We deliberately favor fidelity over speed here~\citep{beyer2024vitspeed}---the $288$px input and the absence of any center crop avoid truncating charts, tables, and documents.

\smallsec{Matching procedure.}
We build a single exact (brute-force, flat $L_2$) FAISS index~\citep{douze2025faiss} over all evaluation-image descriptors. For each training image we retrieve its top-1 nearest evaluation image and convert the squared-$L_2$ distance $d$ to a cosine similarity $s = 1 - d/2$ (valid because SSCD descriptors are unit-normalized). 
A training sample is discarded if \emph{any} of its images attains $s \geq 0.75$ to \emph{any} evaluation image. For multi-image samples (e.g.\ multimodal documents) we thus aggregate by the maximum similarity over the sample's images. Because the evaluation index is small, exact search resolves millions of queries per second on a single GPU, so we use no approximate indexing. Our pipeline relies on a distributed implementation that builds on top of exact FAISS~\cite{douze2025faiss} search.

\begin{figure}[p]
    \centering
    \includegraphics[width=\linewidth]{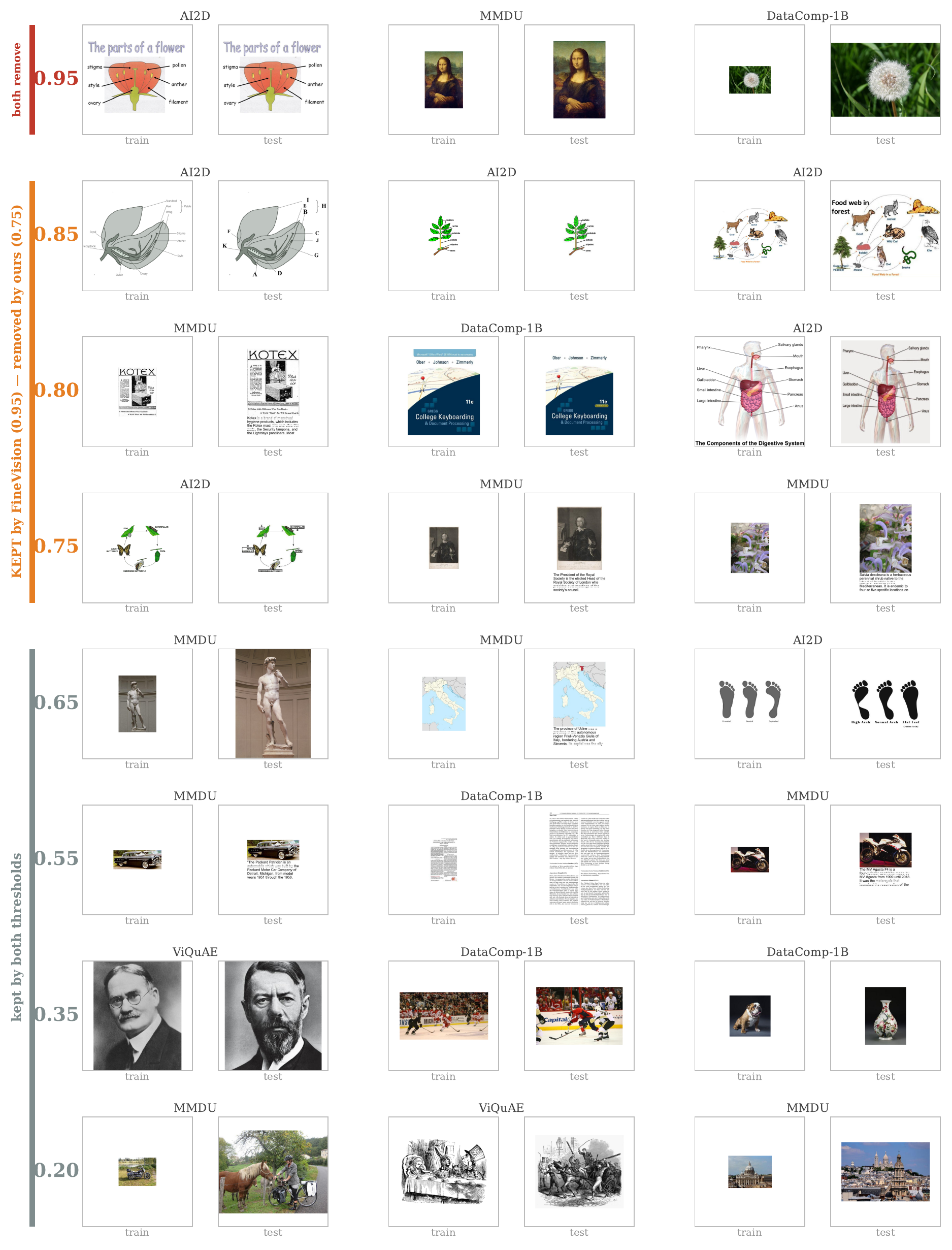}
    \caption{\textbf{Qualitative train/test image matches by SSCD similarity band.} Each pair shows a training-pool image (left) and its top-1 match in the \dcvlm-Extended evaluation suite (right). Rows are grouped by the maximum SSCD cosine similarity of the pair, and labels above each pair give the pool source. Only near-pixel-identical copies score $\geq 0.95$ (top row, removed by both protocols). The $0.75$--$0.95$ band (orange) consists almost entirely of genuine duplicates under benign transformations---label-stripped or re-typeset diagrams, re-crops with text overlays, re-scans---which FineVision's $0.95$ threshold \emph{keeps} but our $0.75$ threshold removes. Below $0.75$ (gray), matches are similar but genuinely distinct images (different photographs of the same statue, different maps of the same country, same-layout documents) and are retained by both protocols.}
    \label{fig:imagedecont-qualitative}
\end{figure}

\smallsec{Choosing the threshold.}
We use a similarity threshold of $0.75$, deliberately more conservative than the $0.95$ adopted by FineVision~\citep{wiedmann2025finevision}. \cref{fig:imagedecont-qualitative} shows representative train/test matches across the SSCD similarity scale, drawn from four pool sources that together span natural photographs (ViQuAE, DataComp-1B), diagrams (AI2D), and mixed document/chart-style imagery (MMDU). Only near-pixel-identical copies score $\geq 0.95$. The $0.75$--$0.95$ range is instead dominated by \emph{genuine} duplicates that have undergone benign transformations---diagrams with labels stripped or re-typeset, re-crops with added text overlays, and re-scans of the same source image. A $0.95$ threshold retains all of these, i.e.\ it admits many false negatives. We found this effect to be systematic for non-natural image domains---charts, tables, documents, and UI screenshots---where genuine train/test near-duplicates receive lower SSCD scores than natural photographs do. Our stricter $0.75$ cutoff removes this band at the cost of some false positives: we knowingly discard a modest amount of clean training data in exchange for substantially stronger decontamination guarantees. Below $0.75$, matches are visually or semantically similar but distinct images (different photographs of the same landmark, same-layout documents with different content) and are correctly retained. Note that even below the $0.75$ threshold, there are indeed some false negatives, however we found $0.75$ to be a stable threshold that balanced both precision and recall, enabling us to perform decontamination in a faithful manner while minimizing discard rates.

\begin{figure}[t]
    \centering
    \includegraphics[width=0.6\linewidth]{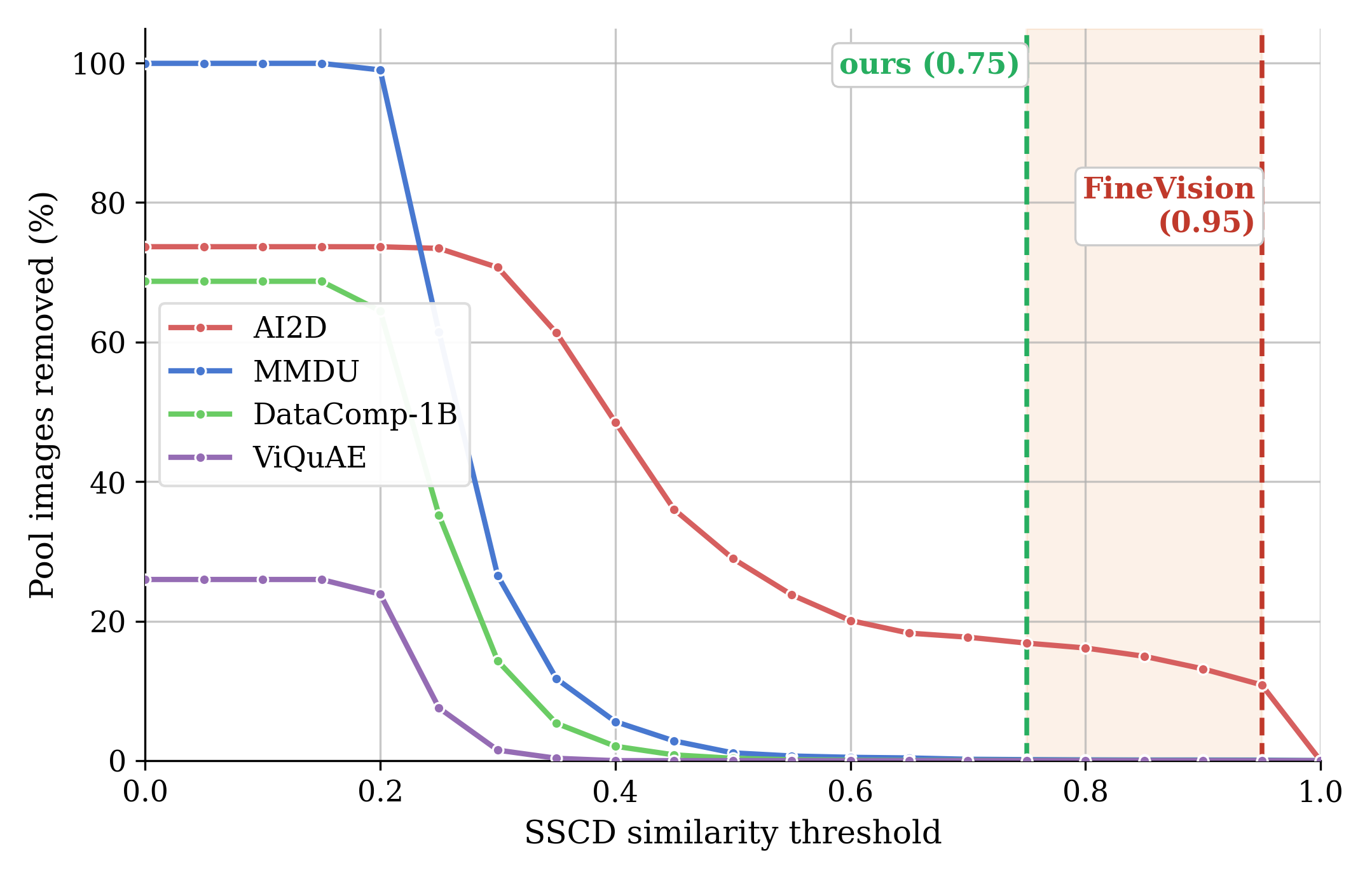}
    \caption{\textbf{Image-decontamination removal rate vs.\ SSCD similarity threshold.} Fraction of pool images removed as a function of the similarity threshold for four sources. The shaded band between our threshold ($0.75$, green) and FineVision's ($0.95$, red) is detected train/test overlap that FineVision's protocol leaves in the training pool.}
    \label{fig:imagedecont-sweep}
\end{figure}

\smallsec{Comparison to FineVision's threshold.}
\cref{fig:imagedecont-sweep} quantifies the gap between the two protocols: for each of the four sources of \cref{fig:imagedecont-qualitative} we plot the fraction of pool images removed as a function of the threshold. The region between $0.75$ and $0.95$ is overlap that FineVision's protocol leaves in the pool---for AI2D, the $0.75$ threshold removes $16.9\%$ of pool images versus only $10.9\%$ at $0.95$, i.e.\ over a third of the detected overlap survives the lenient threshold.

\smallsec{Removal rates on our DCVLM pool.}
The fraction of training images flagged varies sharply by source and is concentrated in datasets whose training and evaluation splits are drawn from the same underlying image distribution. At the $0.75$ threshold, notable per-source removal rates on our DCVLM-pool \texttt{large} scale include ScienceQA ($60.4\%$), ChartQA ($25.6\%$), AI2D ($25.0\%$), OCR-VQA ($2.8\%$), and InfoVQA ($100\%$, as its training and evaluation images entirely coincide). 

\subsection{Text-Based Decontamination}\label{appsub:text-decont}
\begin{figure}[t]
    \centering
    \begin{subfigure}[t]{0.32\linewidth}
        \includegraphics[width=\linewidth]{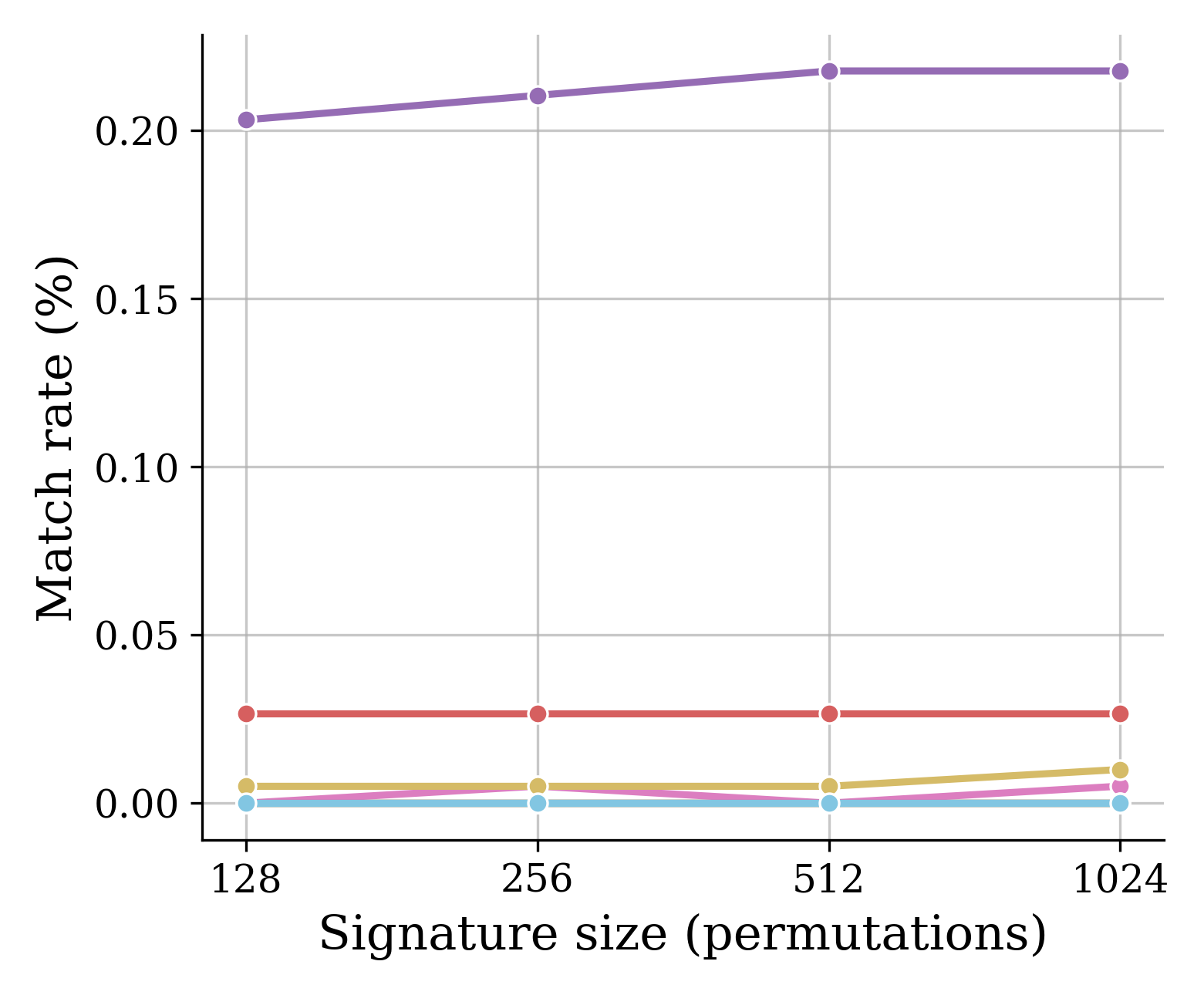}
        \caption{signature size has no effect.}
        \label{fig:textdecont-sigsize}
    \end{subfigure}
    \hfill
    \begin{subfigure}[t]{0.32\linewidth}
        \includegraphics[width=\linewidth]{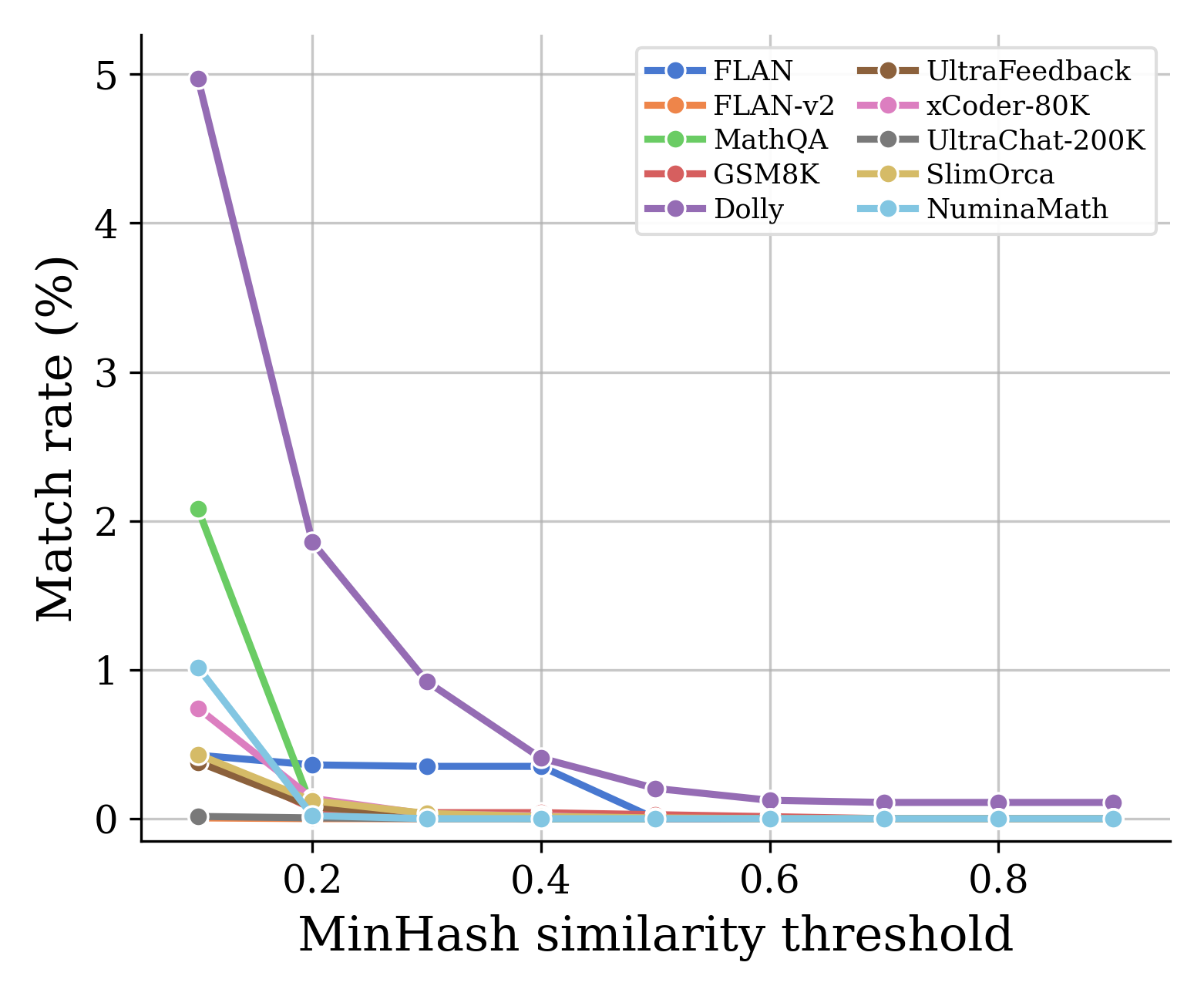}
        \caption{match rate vs.\ threshold.}
        \label{fig:textdecont-ngram5}
    \end{subfigure}
    \hfill
    \begin{subfigure}[t]{0.32\linewidth}
        \includegraphics[width=\linewidth]{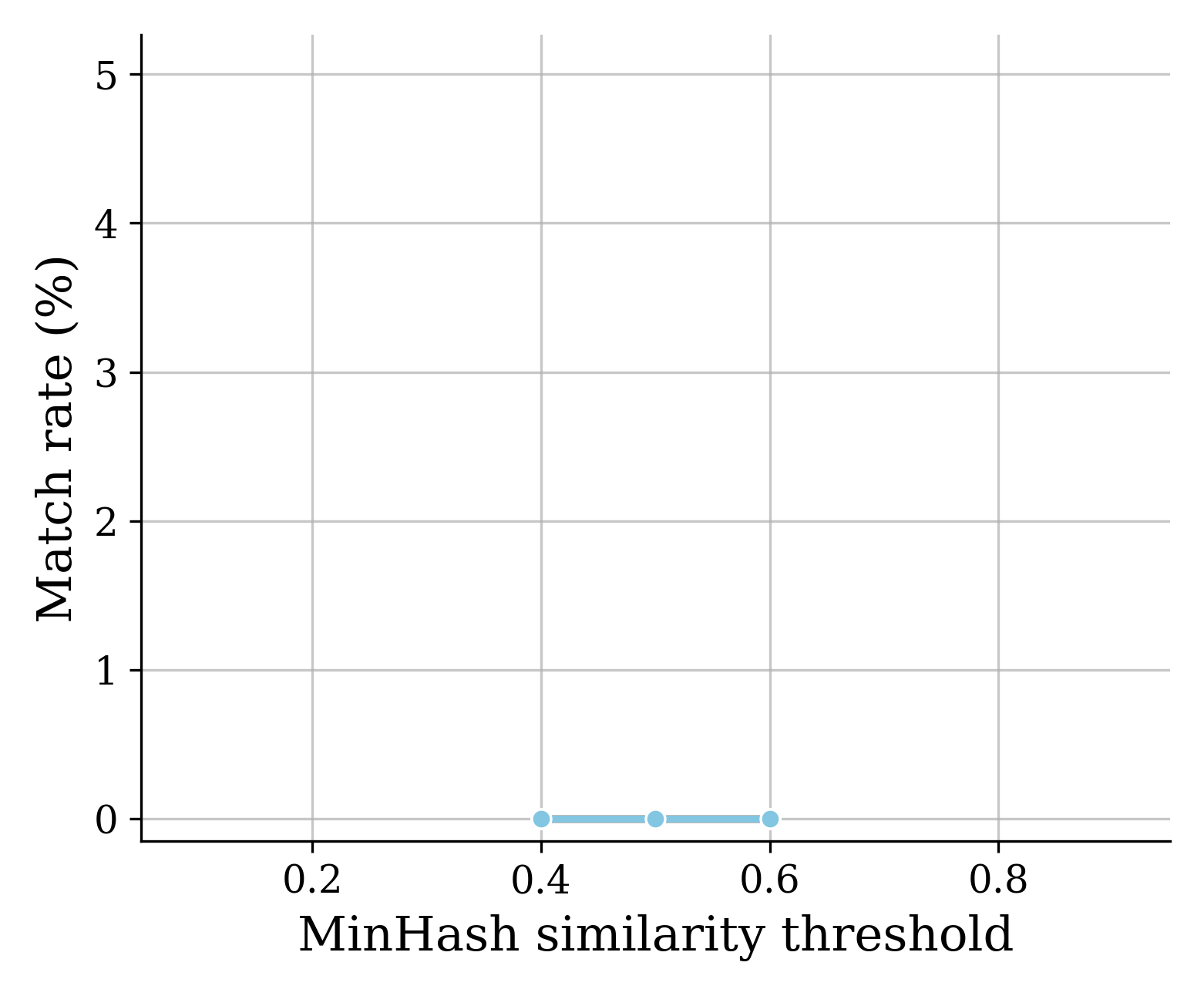}
        \caption{$13$-grams detect almost nothing.}
        \label{fig:textdecont-ngram13}
    \end{subfigure}
    \caption{\textbf{Text-decontamination hyperparameter sweep.} Per-dataset match rate (\% of training queries flagged) across the three MinHash design choices. Signature size is immaterial (a), so we use $128$ permutations. Longer $n$-grams miss short evaluation prompts and detect almost no overlap (c), so we use $5$-grams (b). Match rates are small in absolute terms and fall steeply with the threshold.}
    \label{fig:textdecont-sweep}
\end{figure}

\begin{figure}[t]
    \centering
    \includegraphics[width=0.5\linewidth]{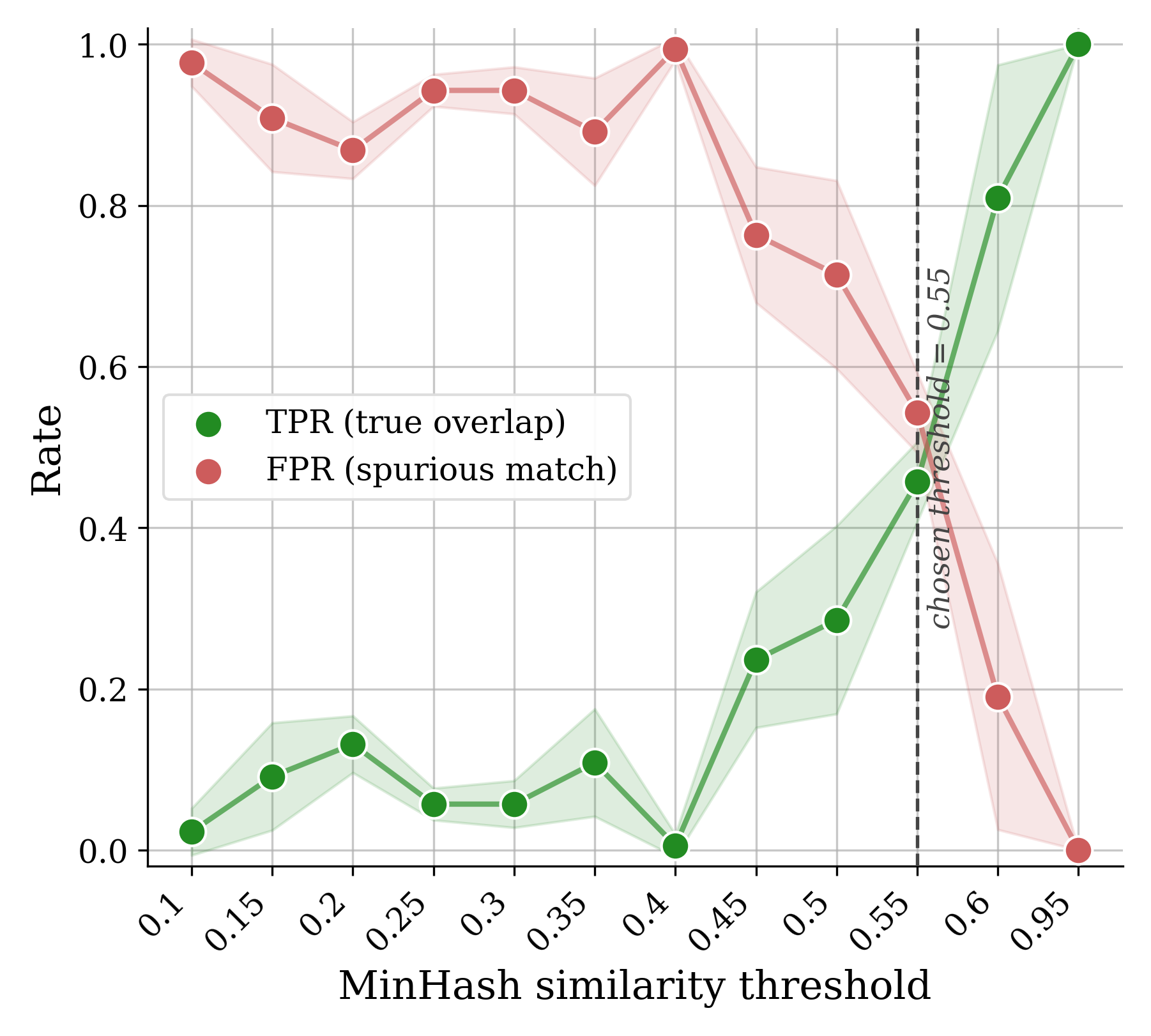}
    \caption{\textbf{Threshold selection via human annotation.} True-positive (TPR, green) and false-positive (FPR, red) rates of MinHash matches at each similarity bin, aggregated over seven annotators. Below ${\sim}0.55$ most matches are spurious (high FPR). The TPR climbs steeply just above it and the curves cross around $0.55$--$0.6$, motivating our first-stage threshold of $0.55$.}
    \label{fig:textdecont-threshold}
\end{figure}

For the text-only portion of our pool, we decontaminate at the \emph{prompt/question level}, following the protocol of Tulu-3~\citep{lambert2024t}: for each training sample we concatenate its human turns into a single query string and match it against the prompts of every text-only evaluation. We decontaminate against all text-only benchmarks in the InternVL-3 evaluation pool ($16$ benchmarks, namely MMLU, CMMLU, C-Eval, GaoKao, TriviaQA, NaturalQuestions, RACE, WinoGrande, HellaSwag, BBH, GSM8K, MATH, HumanEval, MBPP, MBPP-CN, TheoremQA). We use a two-stage MinHash + exact-substring pipeline, detailed below. \cref{fig:textdecont-sweep,fig:textdecont-threshold} justify our hyperparameter choices, and~\cref{tab:textdecont-examples} shows representative detected overlaps.

\smallsec{Matching procedure.}
We tokenize text with the Tulu-3 regular-expression tokenizer (lowercasing and splitting on whitespace and punctuation) and represent each sample by the set of its \emph{word-level} $5$-grams.\footnote{This tokenizer is tuned for whitespace-delimited languages and does not segment Chinese well; we treat the resulting under-matching of Chinese text as a known limitation.} We summarize each set with a $128$-permutation MinHash signature~\citep{broder1997resemblance} and estimate the Jaccard similarity between a training sample and its nearest evaluation sample by \emph{exact} (brute-force) signature comparison. Because the evaluation index is small (${\sim}86$k samples), exact top-1 search is fast enough that we do not require LSH-based approximate matching.

\smallsec{Two-stage filtering.}
In the first stage, we discard every training sample whose top-1 Jaccard similarity to any evaluation prompt exceeds $0.55$. In the second stage, training samples falling in the ambiguous band ($0.3 \leq \text{Jaccard} < 0.55$) are subjected to a \emph{bi-directional} exact-substring check: a sample is discarded if either the training query is contained verbatim in the matched evaluation prompt or vice versa. The second stage recovers true positives whose Jaccard score is diluted below $0.55$---for example, a short evaluation question embedded inside a longer training prompt. This combination balances recall (fuzzy near-duplicates caught by MinHash) with precision (the substring check filters out spurious matches in the intermediate band). Training samples too short to form a single $5$-gram are always retained.

\smallsec{Choosing the hyperparameters.}
We sweep the three free parameters of the MinHash stage and report per-dataset match rates over $10$ instruction datasets in~\cref{fig:textdecont-sweep}. (i) \emph{Signature size}: increasing the number of permutations from $128$ to $1024$ leaves match rates essentially unchanged (\cref{fig:textdecont-sigsize}), so we use $128$ for faster processing. (ii) \emph{$n$-gram length}: $5$-grams detect overlaps at meaningful rates, whereas $8$- and $13$-grams detect almost nothing (\cref{fig:textdecont-ngram13}), because many evaluation prompts are shorter than $8$ (let alone $13$) tokens and therefore never form a matching $n$-gram. We therefore use $5$-grams for higher recall. (iii) \emph{Threshold}: across all datasets the match rate falls steeply with the similarity threshold and is already below ${\sim}0.1\%$ by $0.5$ (\cref{fig:textdecont-ngram5}). At our operating point of $0.55$ fewer than $0.02\%$ of text samples are removed, so a conservative threshold costs almost no data.

\smallsec{Selecting the threshold via human annotation.}
The match-rate sweep does not by itself tell us \emph{which} threshold separates genuine contamination from coincidental overlap. To answer this quantitatively, we ran a small human-annotation study: we drew candidate matches spanning all similarity bins from the $10$ datasets above, and seven of the authors independently labeled each candidate as a true overlap (the two prompts ask the same question) or a false positive. Aggregating the true-positive (TPR) and false-positive (FPR) rates per similarity bin (\cref{fig:textdecont-threshold}), we find that below ${\sim}0.55$ the FPR is high (most matches are spurious, e.g.\ questions that share a template but have different answers), while the TPR rises steeply just above it and the two curves cross around $0.55$--$0.6$. We therefore adopt $0.55$ as the first-stage threshold, and rely on the second-stage substring check to recover the genuine matches that fall just below it.

\begin{table}[t]
    \centering
    \small
    \caption{\textbf{Representative text overlaps detected by our pipeline.} We present some training/evaluation prompt pairs with their estimated Jaccard similarity. High-similarity pairs are genuine near-duplicates and are removed (Stage 1). Pairs just below the $0.55$ threshold typically share a template but have different answers and are correctly retained.}
    \label{tab:textdecont-examples}
    \begin{tabularx}{\linewidth}{@{}c X X c@{}}
        \toprule
        \textbf{Jaccard} & \textbf{Training prompt} & \textbf{Evaluation prompt} & \textbf{Decision} \\
        \midrule
        $1.00$ & What is the loudest animal on Earth? & What is the loudest animal on earth? & \textcolor{lightred}{Removed} \\
        $1.00$ & Who was the first man to walk on the moon? & Who was the first man to walk on the Moon? & \textcolor{lightred}{Removed} \\
        $1.00$ & What is the capital city of Croatia? & What is the capital city of Croatia? & \textcolor{lightred}{Removed} \\
        $0.65$ & in which city was the first public opera house opened & Opened in 1637, in which city was the first public opera house? & \textcolor{lightred}{Removed} \\
        $0.62$ & According to Greek mythology, who was the first woman on earth? & In Greek mythology, who was the first woman on Earth? & \textcolor{lightred}{Removed} \\
        $0.59$ & What is the largest state in the US? & Which is the largest state in the US? & \textcolor{lightred}{Removed} \\
        \addlinespace
        $0.55$ & What is the tallest building in Seattle? & What is the tallest building in London? & \textcolor{nicergreen}{Kept} \\
        $0.51$ & who is considered the father of computers? & who is considered the father of modern cosmology & \textcolor{nicergreen}{Kept} \\
        $0.48$ & Write a Python function to find the maximum of three given numbers. & Write a Python function to find the maximum of two numbers. & \textcolor{nicergreen}{Kept} \\
        $0.48$ & What is the largest lake in Switzerland? & What is the largest lake in Central America? & \textcolor{nicergreen}{Kept} \\
        $0.45$ & What is the capital city of Malaysia? & What is the capital city of Colombia? & \textcolor{nicergreen}{Kept} \\
        $0.45$ & Who is the founder of SpaceX? & Who is the founder of Sikhism? & \textcolor{nicergreen}{Kept} \\
        $0.30$ & Caroline of Ansbach was the wife of which British monarch? & Mrs Maria Fitzherbert was the wife of which British monarch? & \textcolor{nicergreen}{Kept} \\
        $0.30$ & Nancy earns \$28 for working 4 hours. How many hours does she have to work to earn \$70? & Mai earns \$5.50 per hour at her after-school job. How many hours does she have to work to earn \$132? & \textcolor{nicergreen}{Kept} \\
        $0.28$ & What are the types of RVs? & What are the types of scanning? & \textcolor{nicergreen}{Kept} \\
        $0.28$ & What is the best way to make Indian ginger tea? & Which is the best way to make friends with an American? & \textcolor{nicergreen}{Kept} \\
        \bottomrule
    \end{tabularx}
\end{table}

\subsection{Overall removal rates}
\begin{figure}[t]
    \centering
    \includegraphics[width=0.62\linewidth]{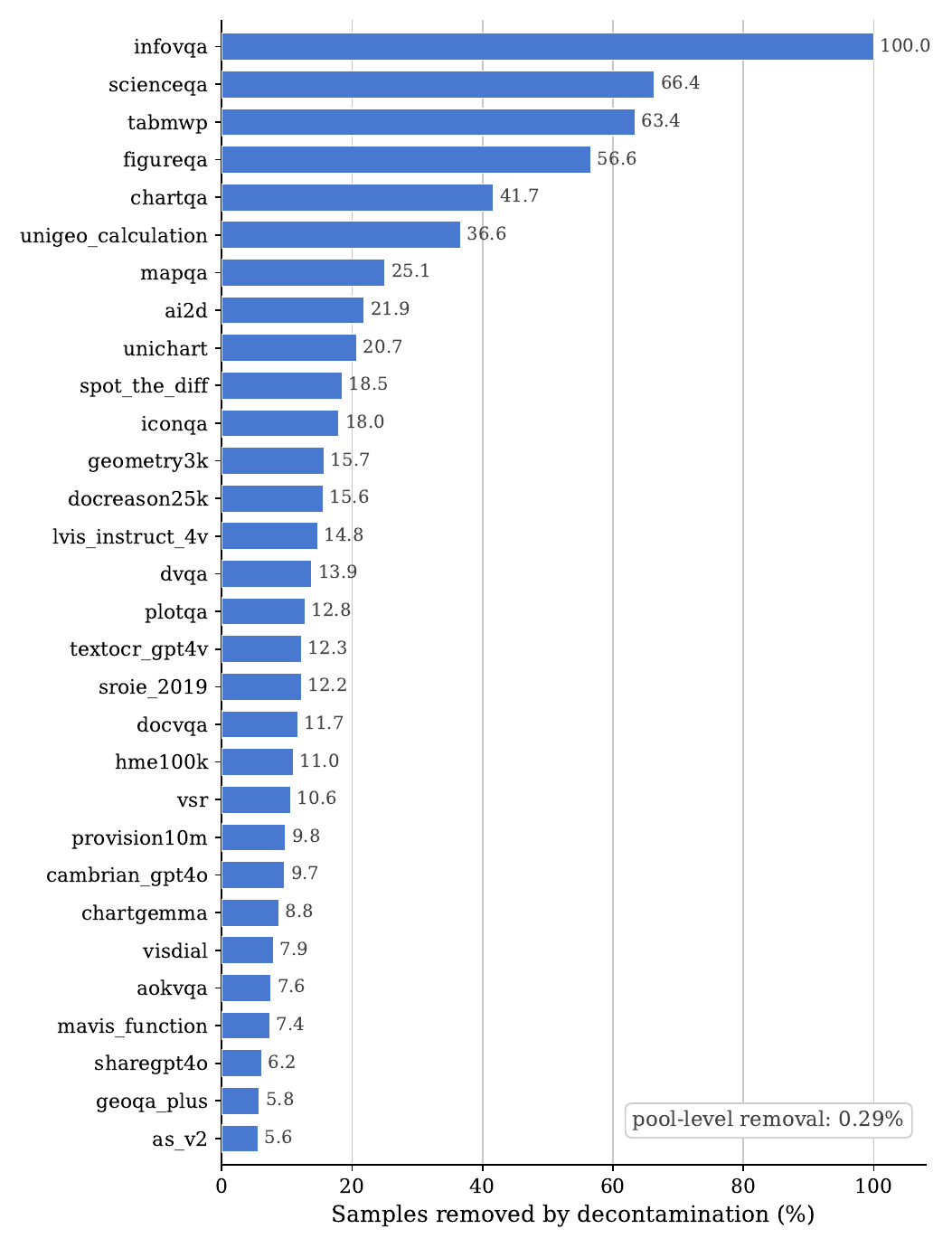}
    \caption{\textbf{Per-dataset removal rates of the full decontamination protocol.} The $30$ pool sources with the highest fraction of samples removed by the combined image-based (SSCD $s \geq 0.75$) and text-based (MinHash Jaccard $\geq 0.55$ + substring check) decontamination against the \dcvlm-Extended evaluation suite. Removal concentrates in sources whose training and evaluation splits share an underlying distribution (InfoVQA, ScienceQA, TabMWP). At the pool level only $0.29\%$ of samples are removed.}
    \label{fig:decont-removal-rates}
\end{figure}

\cref{fig:decont-removal-rates} reports per-dataset removal rates of the full decontamination protocol (image- and text-based combined) over our \texttt{medium} DCVLM-pool as a representative. Removal is concentrated in datasets whose training and evaluation splits are drawn from the same underlying distribution---InfoVQA loses $100\%$ of its samples (its training and evaluation images almost entirely coincide), ScienceQA $66.4\%$, TabMWP $63.4\%$, FigureQA $56.6\%$, ChartQA $41.7\%$, and AI2D $21.9\%$. At the pool level, however, decontamination is cheap---only $0.29\%$ of all training samples are removed. Strict decontamination therefore costs almost no data while removing precisely the samples most likely to inflate benchmark scores.

\clearpage

\section{Additional Experiments and Details}\label{app:filtering-details}
\begin{figure}
    \centering
    \includegraphics[width=\linewidth]{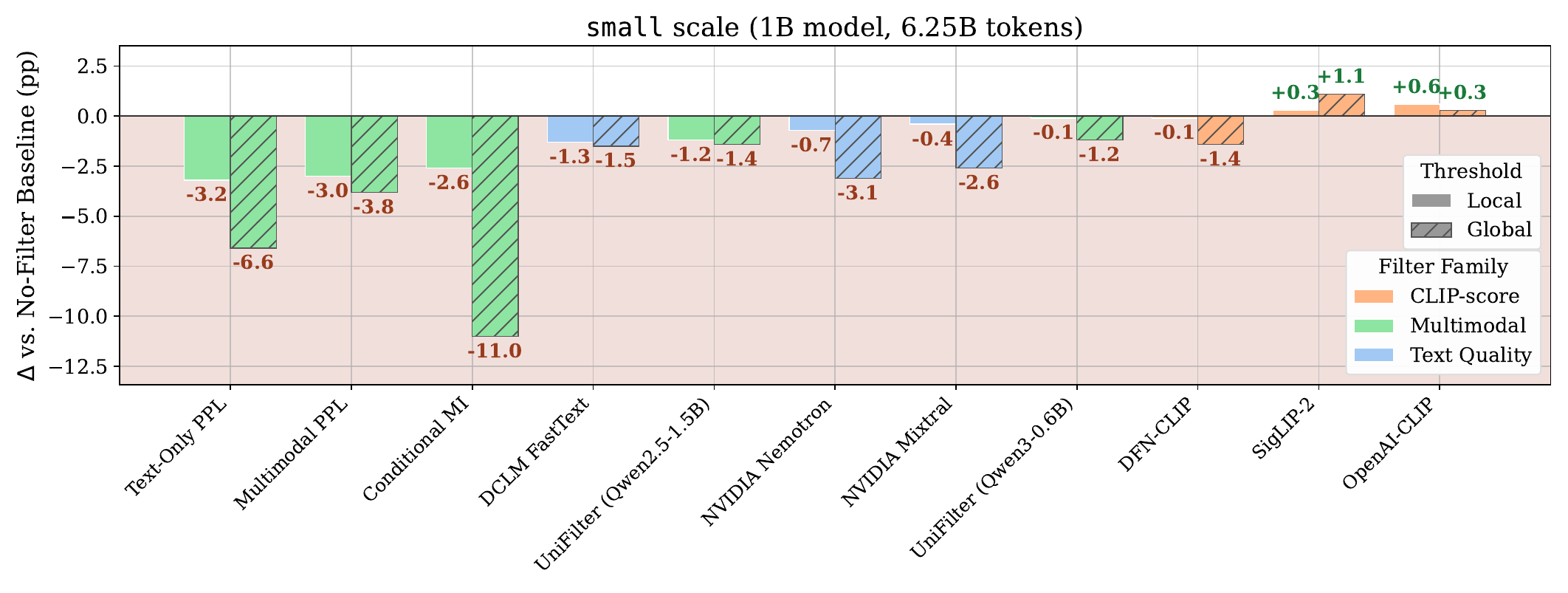}
    \caption{\textbf{Filtering rarely helps, but changing the data composition does move performance substantially (\textit{cont}).} Complementary experiments at the \texttt{small} scale of our benchmark confirm the outcome of \cref{sec:filtering}: established quality filtering rarely provides significant gains over a no-filter baseline, yet variations between global and local filtering remain frequent.}
    \label{fig:filtering-small}
\end{figure}

\subsection{Filtering rarely helps}\label{app:filtering-dont-help-sec}
\begin{table}[t]
\centering
\caption{\texttt{small} scale results for our local filtering experiments (for faster experimentation, we use our 13-task \textbf{Validation} set). We report within parentheses the data type to which the filter has been applied.}
\label{tab:local-filtering-val-small}
\small
\begin{tabular}{lrrrrrrr}
\toprule
\textbf{Filter} & \textbf{Gen} & \textbf{Know} & \textbf{OCR} & \textbf{Vision} & \textbf{MTL} & \textbf{Text} & \textbf{Val Avg} \\
\midrule
\rowcolor{gray!8} \multicolumn{8}{l}{\texttt{small} scale} \\
\midrule
\textit{No Filter} & 45.6 & 38.5 & 43.1 & 44.4 & 46.6 & 35.9 & 42.4 \\
Conditional MI (all) & 40.3 & 37.5 & 46.0 & 41.2 & 46.9 & 28.8 & 39.7 \\
Conditional MI (mmdoc) & 44.0 & 39.0 & 41.5 & 43.5 & 46.1 & 36.6 & 41.8 \\
Conditional MI (im-cap+mmdoc+inst) & 42.0 & 37.9 & 47.0 & 41.4 & 45.7 & 36.1 & 41.4 \\
DCLM FastText (all) & 43.9 & 36.7 & 42.8 & 41.4 & 45.7 & 36.0 & 41.0 \\
DCLM FastText (text-only) & 46.8 & 37.6 & 42.3 & 42.9 & 46.8 & 36.3 & 42.2 \\
DFN-CLIP (im-cap) & 46.8 & 39.0 & 41.2 & 42.7 & \textbf{48.5} & 37.1 & 42.4 \\
DFN-CLIP (im-cap+mmdoc) & 46.2 & 39.1 & 39.3 & 44.5 & 46.5 & 33.5 & 41.7 \\
DFN-CLIP (im-cap+mmdoc+inst) & 45.8 & 39.0 & 39.3 & 41.0 & 45.6 & 35.0 & 41.0 \\
IQA, Kadid-10k (im-cap+mmdoc+inst) & 44.4 & 37.4 & 42.1 & 41.8 & 46.9 & 36.4 & 41.3 \\
IQA, TID 2013 (im-cap+mmdoc+inst) & 44.9 & 37.6 & 40.4 & 42.8 & 47.0 & 36.6 & 41.5 \\
Length (all tokens) & 45.5 & 38.8 & 40.8 & 42.8 & 46.5 & 36.1 & 41.8 \\
Length (text-only tokens) & 41.8 & 38.2 & 41.9 & 41.1 & 45.7 & 34.2 & 40.2 \\
Length Trainable & 44.7 & 38.2 & 40.6 & 41.2 & 46.8 & 35.9 & 41.1 \\
Multimodal PPL (all) & 43.4 & 36.5 & 36.5 & 41.3 & 46.7 & 36.5 & 40.0 \\
Multimodal PPL (mmdoc) & 46.2 & 39.1 & 43.8 & 42.7 & 45.7 & 35.8 & 42.3 \\
Multimodal PPL (im-cap+mmdoc+inst) & 40.4 & 35.3 & 37.4 & 41.6 & 44.9 & 35.6 & 39.0 \\
NVIDIA Mixtral (all) & 46.4 & 38.7 & 40.6 & 44.0 & 47.6 & 37.6 & 42.5 \\
NVIDIA Mixtral (text-only) & \textbf{48.6} & 38.9 & 42.2 & 43.7 & 48.1 & 37.4 & \textbf{43.2} \\
NVIDIA Nemotron (all) & 46.1 & 38.2 & 41.8 & 42.7 & 47.8 & \textbf{37.7} & 42.3 \\
NVIDIA Nemotron (text-only) & 46.5 & 37.6 & 43.0 & 43.7 & 47.3 & 37.1 & 42.6 \\
OpenAI-CLIP (im-cap) & 46.7 & 38.9 & 45.1 & 44.4 & 46.8 & 36.7 & \textbf{43.2} \\
OpenAI-CLIP (im-cap+mmdoc) & 46.6 & 38.3 & 44.3 & \textbf{45.2} & 46.0 & 36.1 & 43.0 \\
OpenAI-CLIP (im-cap+mmdoc+inst) & 46.5 & 37.0 & 42.6 & 41.2 & 45.1 & 35.3 & 41.4 \\
SigLIP-2 (im-cap) & 45.9 & 38.1 & \textbf{47.1} & 42.9 & 46.4 & 35.7 & 42.7 \\
SigLIP-2 (im-cap+mmdoc) & 45.8 & 38.7 & 46.3 & 43.6 & 46.8 & 35.4 & 42.8 \\
SigLIP-2 (im-cap+mmdoc+inst) & 46.0 & 37.8 & 46.8 & 41.3 & 45.5 & 35.7 & 42.1 \\
Text-Only PPL (all) & 42.9 & 37.9 & 31.9 & 42.0 & 46.4 & 36.1 & 39.4 \\
Text-Only PPL (mmdoc) & 45.3 & 38.6 & 41.4 & 42.8 & 46.1 & 36.0 & 41.7 \\
Text-Only PPL (im-cap+mmdoc+inst) & 40.4 & 35.7 & 34.0 & 42.6 & 43.6 & 33.8 & 38.5 \\
UniFilter (Qwen2.5-1.5B) & 43.7 & 38.8 & 40.2 & 41.6 & 46.5 & 36.5 & 41.1 \\
UniFilter (Qwen3-0.6B) & 47.7 & \textbf{40.1} & 40.2 & 41.3 & 47.6 & 35.9 & 42.1 \\
\bottomrule
\end{tabular}
\end{table}

\begin{table}
\centering
\caption{\texttt{medium} scale results for our local filtering experiments (for faster experimentation, we use our 13-eval \textbf{Validation} set). We report in parentheses the data type to which the filter was applied.}
\label{tab:local-filtering-val-medium}
\small
\begin{tabular}{lrrrrrrr}
\toprule
\textbf{Filter} & \textbf{Gen} & \textbf{Know} & \textbf{OCR} & \textbf{Vision} & \textbf{MTL} & \textbf{Text} & \textbf{Val Avg} \\
\midrule
\rowcolor{gray!8} \multicolumn{8}{l}{\texttt{medium} scale} \\
\midrule
\textit{No Filter} & 60.5 & 50.8 & 56.9 & 47.4 & 60.4 & 47.3 & 53.4 \\
Conditional MI (all) & 55.7 & 48.4 & 57.9 & 44.7 & 59.1 & 39.4 & 50.1 \\
Conditional MI (mmdoc) & 60.1 & 50.8 & 57.4 & 48.9 & 61.0 & 47.7 & 53.8 \\
Conditional MI (im-cap+mmdoc+inst) & 59.1 & 50.0 & 56.9 & 45.0 & 60.4 & 47.0 & 52.3 \\
DCLM FastText (all) & 54.5 & 48.1 & 55.8 & 43.3 & 55.3 & 47.3 & 50.1 \\
DCLM FastText (text-only) & 59.9 & 50.7 & 57.3 & 48.9 & 61.6 & 47.4 & 53.8 \\
DFN-CLIP (im-cap) & 59.6 & 50.5 & 55.7 & \textbf{49.5} & 61.5 & 47.3 & 53.5 \\
DFN-CLIP (im-cap+mmdoc) & 60.0 & 49.5 & 56.1 & 48.4 & 60.7 & 47.2 & 53.2 \\
DFN-CLIP (im-cap+mmdoc+inst) & 59.2 & 50.6 & 55.2 & 45.6 & 60.3 & 47.7 & 52.4 \\
IQA, Kadid-10k (im-cap+mmdoc+inst) & 60.0 & 51.1 & 57.2 & 45.2 & 60.0 & 47.5 & 52.9 \\
IQA, TID 2013 (im-cap+mmdoc+inst) & 59.4 & 50.2 & 56.9 & 47.1 & 60.6 & 47.3 & 53.0 \\
Length, all tokens & 58.5 & 50.2 & 56.6 & 46.6 & 60.4 & 47.1 & 52.6 \\
Length, language tokens & 55.3 & 46.8 & 56.8 & 45.7 & 56.0 & 46.5 & 50.7 \\
Length, answer tokens & 60.1 & \textbf{51.3} & 57.1 & 44.5 & 60.4 & 47.4 & 52.8 \\
Multimodal PPL (all) & 59.9 & 50.3 & 58.8 & 45.3 & 61.0 & 46.9 & 53.0 \\
Multimodal PPL (mmdoc) & 59.6 & \textbf{51.3} & 56.8 & 47.6 & 61.7 & 47.7 & 53.5 \\
Multimodal PPL (im-cap+mmdoc+inst) & 59.2 & 50.0 & \textbf{59.9} & 46.8 & 58.3 & 47.0 & 53.1 \\
NVIDIA Mixtral (all) & 60.2 & 50.1 & 57.0 & 47.5 & 60.1 & 47.5 & 53.3 \\
NVIDIA Mixtral (text-only) & 61.0 & 51.0 & 57.0 & 48.6 & \textbf{62.6} & \textbf{47.9} & \textbf{54.1} \\
NVIDIA Nemotron (text-only) & \textbf{61.8} & 49.9 & 57.2 & 46.8 & 62.0 & 47.5 & 53.6 \\
OpenAI-CLIP (im-cap) & 59.7 & 50.4 & 57.8 & 47.2 & 59.9 & 47.5 & 53.2 \\
OpenAI-CLIP (im-cap+mmdoc) & 60.4 & 50.1 & 58.0 & 47.4 & 61.6 & 47.3 & 53.5 \\
OpenAI-CLIP (im-cap+mmdoc+inst) & 55.9 & 48.5 & 57.4 & 44.8 & 59.1 & 47.2 & 51.3 \\
SigLIP-2 (im-cap) & 60.2 & 50.3 & 57.7 & 48.7 & 60.8 & 47.4 & 53.7 \\
SigLIP-2 (im-cap+mmdoc) & 60.1 & 50.8 & 59.2 & 47.8 & 61.6 & 47.5 & 53.9 \\
SigLIP-2 (im-cap+mmdoc+inst) & 58.0 & 48.9 & 57.8 & 43.7 & 59.6 & 47.3 & 51.7 \\
Text-Only PPL (all) & 59.5 & 50.1 & 57.7 & 46.0 & 60.1 & 47.2 & 52.8 \\
Text-Only PPL (mmdoc) & 60.2 & 50.3 & 57.7 & 48.9 & 60.6 & 47.4 & 53.8 \\
Text-Only PPL (im-cap+mmdoc+inst) & 59.3 & 51.1 & 57.5 & 47.1 & 59.0 & 47.2 & 53.1 \\
UniFilter (Qwen2.5-1.5B) & 58.2 & 49.6 & 55.8 & 43.8 & 59.9 & 47.2 & 51.6 \\
UniFilter (Qwen3-0.6B) & 59.5 & 50.1 & 57.1 & 46.9 & 61.1 & 46.8 & 52.9 \\
\bottomrule
\end{tabular}
\end{table}

\Cref{fig:filtering-small} reports the filtering results at the \texttt{small} scale (1B model, 6.25B tokens), complementing the \texttt{medium}-scale results in \Cref{sec:filtering}. 
The conclusions are fully consistent: local filtering uniformly fails to improve over the no-filter baseline. 
The only positive outcomes still arise from global CLIP-score filtering (SigLIP-2 $+1.1$pp and OpenAI-CLIP $+0.6$pp), in line with \cref{fig:filtering}, yet, once again, these gains are lower than what one would expect from prior work.

Consistently, global filtering and local filtering produce different results also at this scale. 
Overall, these results reinforce the central finding of \cref{sec:filtering} that filtering is not a reliable curation strategy, and lead to hypotheses that observed performance variations might be primarily driven by the underlying mixture changes.

\begin{table}[t]
\centering
\small
\caption{Global filtering results (validation set).}
\label{tab:global-filtering-val}
\begin{tabular}{lrrrrrrr}
\toprule
\textbf{Filter} & \textbf{Gen} & \textbf{Know} & \textbf{OCR} & \textbf{Vision} & \textbf{MTL} & \textbf{Text} & \textbf{Val Avg} \\
\midrule
\rowcolor{gray!8} \multicolumn{8}{l}{\texttt{small} scale} \\
\midrule
\textit{No Filter} & 45.6 & 38.5 & 43.1 & \textbf{44.4} & 46.6 & 35.9 & 42.4 \\
Conditional MI & 25.2 & 30.2 & 39.1 & 41.2 & 41.2 & 22.1 & 32.5 \\
DCLM FastText & 40.6 & 36.9 & 41.8 & 41.6 & 45.9 & 36.8 & 40.3 \\
DFN-CLIP & 43.1 & 40.5 & 34.4 & 42.3 & 48.0 & 36.7 & 40.6 \\
Multimodal PPL & 38.2 & 38.7 & 40.9 & 40.9 & 46.1 & 36.1 & 39.6 \\
NVIDIA Mixtral & 38.0 & 37.9 & 36.7 & 41.9 & 47.4 & 37.2 & 39.3 \\
NVIDIA Nemotron & 35.8 & 36.8 & 34.5 & 42.8 & 46.3 & 37.4 & 38.4 \\
OpenAI-CLIP & 45.3 & 37.6 & \textbf{44.4} & 43.9 & 48.3 & 36.4 & 42.5 \\
SigLIP-2 & \textbf{46.0} & \textbf{41.3} & 44.3 & 44.2 & \textbf{48.5} & 37.2 & \textbf{43.4} \\
Text-Only PPL & 33.3 & 37.8 & 22.3 & 43.6 & 45.8 & 37.1 & 36.2 \\
UniFilter (Qwen2.5-1.5B) & 43.9 & 36.5 & 37.0 & 44.2 & 46.2 & 37.5 & 41.0 \\
UniFilter (Qwen3-0.6B) & 45.4 & 38.5 & 35.7 & 42.1 & 46.8 & \textbf{37.8} & 41.0 \\
\midrule
\rowcolor{gray!8} \multicolumn{8}{l}{\texttt{medium} scale} \\
\midrule
\textit{No Filter} & 60.5 & 50.8 & 56.9 & 47.4 & 60.4 & 47.3 & 53.4 \\
Conditional MI & 55.9 & 46.9 & 56.0 & 44.4 & 54.9 & 35.3 & 48.6 \\
DCLM FastText & 59.5 & 51.1 & 58.7 & 48.0 & 62.2 & \textbf{47.5} & 53.8 \\
DFN-CLIP & 59.9 & 49.9 & 56.3 & \textbf{49.7} & \textbf{63.2} & 46.9 & 53.7 \\
Multimodal PPL & 60.4 & 50.6 & 59.6 & 45.2 & 60.3 & 47.4 & 53.2 \\
NVIDIA Mixtral & \textbf{61.1} & 49.4 & 59.1 & 48.3 & 61.6 & 47.4 & 54.0 \\
NVIDIA Nemotron & 60.4 & 50.4 & 57.0 & 45.8 & 61.8 & \textbf{47.5} & 53.1 \\
OpenAI-CLIP & 59.7 & \textbf{52.6} & 58.5 & 47.4 & \textbf{63.2} & \textbf{47.5} & 54.0 \\
SigLIP-2 & 60.6 & 52.2 & \textbf{60.1} & 48.2 & 62.9 & 47.0 & \textbf{54.5} \\
Text-Only PPL & 58.2 & 51.4 & 57.8 & 45.3 & 61.1 & 47.2 & 52.7 \\
UniFilter (Qwen2.5-1.5B) & 58.8 & 51.1 & 58.1 & 44.2 & 60.3 & 47.4 & 52.5 \\
UniFilter (Qwen3-0.6B) & 56.5 & 48.4 & 54.3 & 43.8 & 56.9 & 47.0 & 50.6 \\
\bottomrule
\end{tabular}
\end{table}

\textbf{Additional filtering experiments.} For completeness, we report the full benchmarking we conducted with a variety of additional configurations for the filters in \cref{sec:filtering}. 
Specifically, we tested several local filtering variants of the common filters presented in \cref{sec:filtering}:
\begin{itemize}[leftmargin=*, itemsep=2pt, topsep=0pt, parsep=0pt, partopsep=0pt]
    \item for CLIP-score, we experiment with filtering not only image-caption pairs alone, but also multimodal documents and instruction tuning data. For the former, we score a document according to the average CLIP-score of all its adjacent image-text snippets. For the latter, we employ image-to-question similarity. In case of multi-image examples, we use an identical strategy to multimodal documents and score the closest image to the question in the conversation history. For multi-turn samples, we average across all adjacent pairs as we do for multimodal documents. We use the same three models described in \cref{sec:filtering}
    \item for text quality classifiers, we additionally experiment with filtering only the language portion of our pool. This experiment is motivated by the hypothesis that text-only sources might be more in-distribution with respect to the classifiers themselves. Identically to CLIP-score, we report all three classifiers of the main body.
    \item For Text-Only Perplexity, Multimodal Perplexity, and Conditional Mutual Information filters we experiment with applying them to all multimodal data types jointly (image-caption pairs, instruction-tuning data, multimodal documents), as well as subset combinations. 
\end{itemize}

We additionally experimented with basic filtering strategies:
\begin{itemize}[leftmargin=*, itemsep=2pt, topsep=0pt, parsep=0pt, partopsep=0pt]
    \item Length heuristics, that remove the shortest and the longest samples (removes bottom-10\% shortest and top-90\% longest at both scales). We experiment with several interpretations of ``length'': (i) the count of all tokens, (ii) the count of answer tokens, on top of which the loss is ultimately computed during training; and (iii) the count of language tokens (i.e., question + answer tokens); 
    \item Image Quality Assessment (IQA) models, to remove low-quality images. We test with the ARNIQA~\citep{agnolucci2024arniqa} suite of image quality models. Specifically, we use the regressors trained on the Kadid-10k~\citep{lin2019kadid} and Tid-2013~\citep{ponomarenko2013color} datasets, since they are one of the highest performing image quality assessment classifiers on standard IQA benchmarks.  
\end{itemize}

Except for length heuristics, which use the same (bottom-10\%, top-90\%) percentiles on both scales, all experiments are conducted with an aggressive top-10\% filtering threshold on the small scale and a more lenient top-40\% threshold on the medium scale, following \cref{sec:filtering}. 

Given the large number of experiments, we opted for faster iterations on our 13-task Validation suite. 
As anticipated in the main body, no configuration provides significant gains. 

For completeness, we report Validation results for global filtering in \cref{tab:global-filtering-val} as well (we did not test all variants here, but rather only those presented in \cref{fig:filtering}).

\subsection{Data formatting}\label{appsec:formatting}

We use the Chat Markup Language (ChatML) format~\citep{openai2022chatml} for training, which assigns a role (``system'', ``user'', or ``assistant'') to all messages in a conversation. 
ChatML is a structured format explicitly separating conversation roles (e.g., system, user, assistant) and demarcating message boundaries using specific tokens (e.g., \verb+<|im_start|>+ and \verb+<|im_end|>+).
Text-only and instruction-tuning data typically come in this format already, while image-caption pairs can be easily formatted using templates such as ``\textit{Describe the image}'' to serve as user messages. 
However, 
there is no consensus on how to format multimodal documents,
with prior works exploring contrasting methods~\citep{awadalla2024mint,li2024omnicorpus,zhu2023multimodal,laurenccon2023obelics}. 

\begin{wraptable}{r}{0.27\textwidth}
    \vspace{-12pt}
    \centering
    \caption{Multi-turn is better than in-context formatting.}
    \label{tab:mmdoc-formatting}
    \small
    \begin{tabular}{lc}
        \toprule
        \textbf{Format} & \textbf{Val Avg} \\
        \midrule
        In-context  & 35.5 \\
        Multi-turn  & 37.9 \\
        \bottomrule
    \end{tabular}
    \vspace{-8pt}
\end{wraptable}

We thus conduct an experiment at the \texttt{small} scale to compare two strategies:
\textbf{in-context}, which uses the last text chunk as assistant response, and concatenates everything %
except the last text chunk into the user message
vs \textbf{multi-turn}, which interleaves images and text chunks into multiple turns, so that images are user messages and all text chunks are assistant responses
To ensure reliable signal, we use a uniform mixture across data types ($25$\% for all data types). 
From~\cref{tab:mmdoc-formatting}, we find multi-turn formatting superior to in-context formatting.
We thus adopt multi-turn formatting for all our experiments. %

\subsection{Diminishing returns of downstream filtering}
\label{app:filtering-exp}
This section provides further details on the upstream vs.\ downstream filtering experiment in~\cref{sec:filtering}.

We start from the base \dcvlm{} small-pool of 160 source datasets. By virtue of our collection procedure (\cref{sec:pool}), this pool has already been ``upstream-filtered'' by the source dataset providers, so we treat it as the ``Upstream-filtered=100\%'' setting. It follows our pool's base mixture ratio: $75$\% image-caption pairs, $18$\% text-only, $4$\% multimodal documents, and $3$\% multimodal instruction-tuning data. The \texttt{datacomp\_1b} subset alone accounts for roughly $47$\% of the image-caption portion.

To construct the other two settings, we progressively replace pre-filtered image-caption data with raw, unfiltered pairs crawled directly from the released \href{https://huggingface.co/datasets/mlfoundations/datacomp_pools}{CommonPool} URLs:
\begin{itemize}
    \item \textbf{Upstream-filtered=65\%:} replace pre-filtered \texttt{datacomp\_1b} with its unfiltered counterpart, yielding $0.47 \times 75 \approx 35$\% unfiltered data.
    \item \textbf{Upstream-filtered=25\%:} replace the \emph{entire} image-caption portion with raw unfiltered \texttt{datacomp\_1b}, yielding $75$\% unfiltered data.
\end{itemize}

For each setting, we apply CLIP-score downstream filtering on image-caption pairs using OpenAI-CLIP-ViT-L/14~\citep{radford2021learning}, and train \texttt{small}-scale models on the resulting datasets.

The gain from downstream filtering shrinks monotonically with the level of upstream curation: $+2.4$pp at Upstream-filtered=25\%, $+1.3$pp at $65$\%, and only $+0.6$pp at $100$\%. The first result confirms that filtering \emph{does} help on genuinely noisy data, but the trend establishes that \textit{additional filtering on top of already-curated data operates in a regime of diminishing returns}.

\subsection{Temperature-scaled sampling}\label{appsub:temperature}

Several of our most task-relevant datasets are already present in the pretraining pool: the training splits of ScienceQA, AI2D, ChartQA, InfoVQA and OCR-VQA all live in our instruction-tuning sources.
It is therefore natural to ask why the corresponding benchmarks do not benefit more, and a simple hypothesis presents itself---these datasets may be \emph{severely undersampled} during training.
By default we sample sources \emph{within} each data type in proportion to their size, $p(d) \propto \mathrm{len}(d)$ (where $\mathrm{len}(d)$ is the number of samples $d$ contributes to the pool), so a handful of large web-scale sources can dominate the mixture and drown out small but high-quality or diverse datasets.

Prior work tackles this imbalance in two ways: by hand-setting mixture ratios based on intuition (e.g.\ the Nemotron~\citep{diao2025nemotron} and OLMo-2~\citep{olmo20242} series, and FineWeb~\citep{penedo2024fineweb}), or by principled downweighting of large sources---most notably the square-root sampling used in the Molmo line of work~\citep{deitke2025molmo,clark2026molmo2}.
We study this finer, within-type sampling in a principled manner via temperature-scaled sampling,
\begin{equation}
    p(d) \;\propto\; \mathrm{len}(d)^{1/T}\,,
\end{equation}
where $T=1$ recovers length-proportional sampling, $T=2$ corresponds exactly to the square-root sampling of Molmo, and $T\rightarrow\infty$ approaches uniform sampling over datasets (\emph{flattening}, which upweights small sources). We additionally explore $T<1$, which \emph{sharpens} the distribution toward the largest sources.
We apply the sweep $T \in \{0.5, 0.8, 1.0, 2.0, 4.0\}$ to our balanced mixture ($40$\% image-caption, $5$\% multimodal documents, $15$\% text, $40$\% instruction-tuning, see \cref{sec:mixing}), holding all other settings fixed, and report results on our 13-task Validation suite. We run these experiments at the \texttt{small} scale of our benchmark.

\begin{figure}[t]
    \centering
    \includegraphics[width=0.72\linewidth]{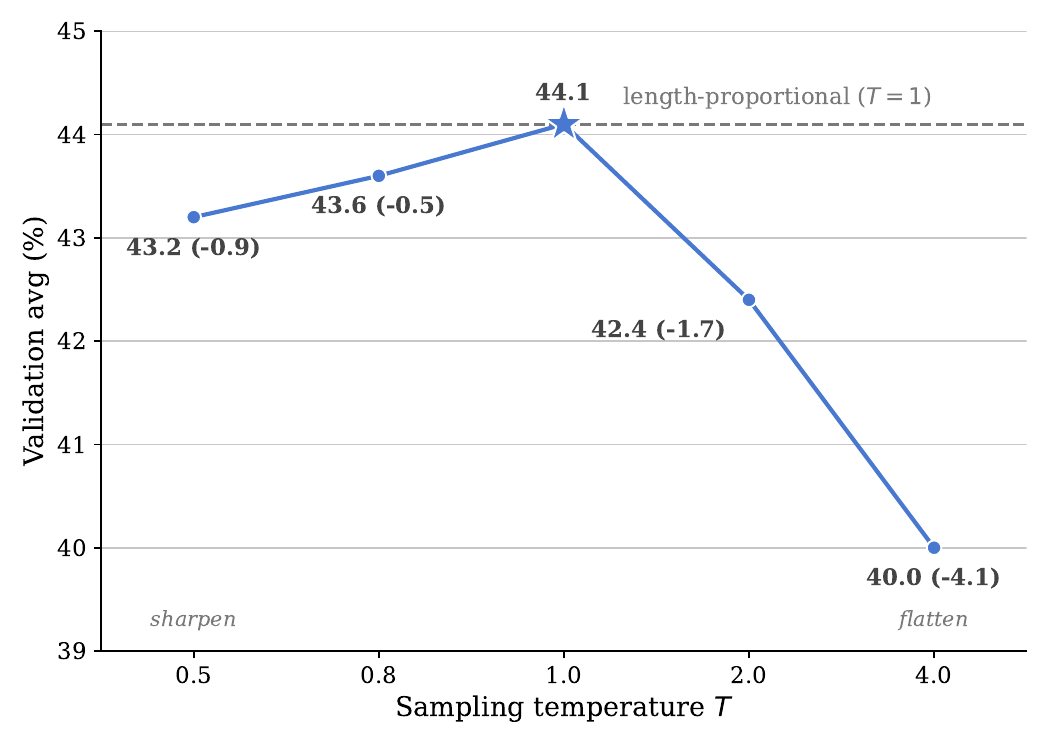}
    \caption{\textbf{Length-proportional sampling ($T=1$) is near-optimal.} Validation average as a function of sampling temperature $T$ in $p(d)\propto\mathrm{len}(d)^{1/T}$. Both sharpening ($T<1$) and flattening ($T>1$) degrade performance relative, with the near-uniform $T=4$ setting losing $4.1$pp.}
    \label{fig:temperature}
\end{figure}

As shown in \cref{fig:temperature}, the undersampling hypothesis does not hold.
The default $T=1$ is already near-optimal, and flattening the distribution---which upweights exactly the small, task-relevant datasets we suspected of being undersampled---consistently \emph{hurts}: square-root sampling ($T=2$) drops to $42.4$\% ($-1.7$pp) and the near-uniform $T=4$ falls to $40.0$\% ($-4.1$pp), below even the \fv{} baseline.
Sharpening offers no benefit either ($T=0.8$: $43.6$\%, $-0.5$pp; $T=0.5$: $43.2$\%, $-0.9$pp).
We hypothesize the contradicting results between ours and the Molmo series are due to the differing mixture ratios and pool composition---we believe there might be some non-trivial effects between the broader data-type mixing ratios and the within-data-type source sampling, which would be an interesting exploration for future work.
Given our results, we hence retain length-proportional ($T=1$) sampling throughout---noting that the cross-type mixing ratios of \cref{sec:mixing} are a far more impactful lever than the within-type temperature.

\subsection{Synthetic recaptioning}\label{appsub:recaptioning}

Replacing noisy web-crawled alt-text with synthetic captions is a well-established lever for improving contrastive (CLIP-style) pretraining~\citep{li2024if,yu2024capsfusion,ghosh2025concept}.
We test whether the same holds for generative VLM pretraining.
We hold the images and their training order fixed and vary \emph{only} the caption text on the DataComp~\citep{gadre2023datacomp} subset of our image-caption pool, comparing three variants:
\begin{itemize}[leftmargin=*, itemsep=2pt, topsep=0pt, parsep=0pt, partopsep=0pt]
    \item \textbf{Alt-text (original):} the raw web alt-text from the original DataComp-1B image-caption pairs.
    \item \textbf{Synthetic short:} concise captions generated by Qwen2-VL-7B~\citep{wang2024qwen2}, conditioned on the image, its alt-text, and a set of detected concepts~\citep{ghosh2025concept}.
    \item \textbf{Synthetic spatial:} the short captions further augmented with explicit 2D/3D spatial-relationship descriptions, to probe whether richer, layout-aware captions help the more vision-centric benchmarks.
\end{itemize}
We source all caption variants using the released datasets of~\citet{ghosh2025concept}. We run these experiments on a mixture of $65$\% image-caption, $5$\% multimodal documents, $15$\% text, and $15$\% instruction-tuning data, at the \texttt{small} scale of our benchmark.
Crucially, the caption intervention is applied \emph{only} to DataComp-1B, not to the entire image-caption portion: since DataComp-1B accounts for roughly $47$\% of the image-caption data, the captions are rewritten for only $\approx\!47\%\times65\% \approx 30\%$ of the \emph{overall} training mixture, while every other source (including image-caption datasets that already come with synthetic captions) retains its original text.
The images, mixture ratios, and data order are identical across the three runs, so any difference is attributable solely to DataComp-1B captions. As with other experiments, we report on the 13-task Validation suite.

\begin{table}[t]
    \centering
    \caption{\textbf{Recaptioning image-caption pairs does not improve VLM pretraining.} Validation scores (\%) per category for the three caption variants. Synthetic captions are generated by Qwen2-VL-7B~\citep{wang2024qwen2}. $\Delta$ is the change in average relative to the original alt-text baseline.}
    \label{tab:recaptioning-val}
    \small
    \begin{tabular}{lccccccc}
        \toprule
        \textbf{Captions} & \textbf{General} & \textbf{Knowledge} & \textbf{OCR} & \textbf{Vision} & \textbf{MTL} & \textbf{Text} & \textbf{Val Avg} \\
        \midrule
        Alt-text (original) & 48.0 & 40.0 & 46.8 & 43.4 & 49.5 & 36.6 & \textbf{44.1} \\
        Synthetic short     & 46.6 & 39.2 & 44.9 & 42.5 & 49.8 & 36.5 & 43.2 \\
        Synthetic spatial   & 48.7 & 39.7 & 45.9 & 42.5 & 50.7 & 36.7 & 44.0 \\
        \bottomrule
    \end{tabular}
\end{table}

\Cref{tab:recaptioning-val} shows that \textbf{neither synthetic captioning scheme provides a meaningful improvement}: original alt-text reaches $44.1$\% average, synthetic-short $43.2$\% ($-0.9$pp), and synthetic-spatial $44.0$\% ($-0.1$pp).
Per-category, the changes are small and inconsistent in sign---spatial captions help General ($+0.7$) and MTL ($+1.2$) but hurt OCR ($-0.9$) and Vision ($-0.9$), netting essentially zero overall.
This contrasts with the consistent gains recaptioning brings to CLIP training, and we hypothesize that the difference is in our pretraining setup: the instruction-tuning portion of the mixture already supplies abundant high-quality, densely-described image-text pairs, so perhaps the marginal value of rewriting the image-caption captions is low.
In other words, at the VLM pretraining stage the \emph{text content} of image-caption pairs matters far less than their \emph{proportion} in the overall mixture (\cref{sec:mixing})---given the substantial cost of running a 7B captioner over the full pool, we use the original alt-text throughout.

We caveat that this null result is specific to our setup, and a different configuration could plausibly surface gains.
Two factors likely dampen the effect here.
First, the intervention touches only $\approx\!30\%$ of the training mixture (the DataComp subset)---recaptioning a larger fraction of the pool, or starting from a pool that is entirely alt-text, may tell a different story.
Second, and more importantly, our image-caption pool \emph{already} contains several sources beyond DataComp-1B and ReLAION that come with high-quality synthetic or re-written captions.
In their presence, rewriting the captions of one additional subset has little marginal effect---much as additional filtering yields diminishing returns once a pool is already well-curated (\cref{app:filtering-exp}).
The benefit of recaptioning may thus be largely realized already, and washed out, by the synthetic captions present elsewhere in the pool.

\subsection{Online Filtering}\label{appsub:online_filtering}

Prior data curation benchmarks such as DataComp~\citep{gadre2023datacomp} and DCLM~\citep{li2024datacomp} employ \emph{offline resharding}: given a raw pool and a set of metadata annotations, the entire dataset is filtered according to a predicate and written to disk as a new, smaller dataset, which is then used for training.
While this makes training efficient---the dataloader never encounters rejected samples---it incurs substantial storage costs, particularly when sweeping over filtering configurations.
For example, determining the optimal CLIP-score threshold requires training runs at several percentiles, each of which produces a separate copy of the filtered pool on disk.
With dozens of filters and multiple threshold values, storage requirements can grow by an order of magnitude.

Both DataComp and DCLM cite the computational cost of \emph{online filtering}---applying predicate conditions during training---as the reason for preferring offline resharding.
We revisit this assumption and find that, in our setting, online filtering matches the throughput of unfiltered training, effectively eliminating the storage bottleneck.
We describe our implementation in detail below.

\subsubsection{Metadata Annotation}

Each sample in our pool is annotated with scores from every filter we consider.
Concretely, for a given filter $f$ (e.g., \texttt{oai-clip-vit-b-32}), we run a batch inference job over the entire pool that computes a scalar quality score $s_f(x) \in \mathbb{R}$ for each sample $x$.
These scores are stored as sidecar metadata files alongside the original WebDataset shards: for each shard \texttt{xxxxx.tar}, we produce a corresponding \texttt{xxxxx.json} file containing a dictionary mapping sample keys to their scores across all filters.
A single sample's metadata entry looks like:
\begin{verbatim}
  "sample_00042_017": {
      ...
      "clip_score_clip_vitl14_224_standard": 31.984,
      "clip_score_siglip2_b16_224_standard": 5.421,
      "text_quality_dclmbaseline_fasttext_score": 0.091,
      "text_quality_nvidia_edu_mixtral_score": 0.122,
      "iqa_kadid10k": 0.574
      ...
  }
\end{verbatim}
This annotation step is performed once per pool and is fully parallelizable across shards.
Adding a new filter (e.g., a new CLIP model variant) requires scoring the entire pool only with the new filter; existing annotations are preserved and do not need to be recomputed.
In total, our small pool ships with over 100 precomputed metadata annotations per sample.

\subsubsection{Threshold Pre-Computation}

Given a filter $f$ and a desired retention rate $p$ (e.g., ``keep the top 60\%''), we need a threshold $\tau_{f,d}$ for each sub-dataset $d$ (in the local filtering setting) or a single global threshold $\tau_f$ (in the global filtering setting).
We pre-compute these thresholds for all filters at percentile increments of 5 (i.e., $p \in \{0.1, 0.15, 0.2, 0.25, \ldots, 0.9, 0.95\}$) using \texttt{torch.quantile()} with linear interpolation over the full score distribution of each sub-dataset (for local filtering) or the full merged score distribution across all sub-datasets (for global filtering).
The output is a lightweight JSON lookup table that maps each filter to the score value corresponding to each percentile, which is then used as the filtering threshold:
\begin{verbatim}
    filter_name: {
      "0.10": 0.052,   # 10th percentile
      "0.15": 0.095,   # 15th percentile
      "0.20": 0.118,   # 20th percentile
      ...
      "0.90": 0.891    # 90th percentile
      "0.95": 0.945    # 95th percentile
    }
\end{verbatim}

\subsubsection{Runtime Integration}

During training, the dataloader reads samples from WebDataset shards as usual, but before a sample is yielded to the training loop, it passes through a lightweight predicate function registered via \texttt{wds.select()}.
The predicate performs the following steps for each sample:
\begin{enumerate}[leftmargin=*, itemsep=2pt]
    \item Look up the sample's pre-computed filter scores from the sidecar metadata (loaded into a hash map at initialization).
    \item For each active filter, compare the sample's score against the corresponding threshold for that particular sub-dataset.
    \item If the score falls below the threshold (i.e., the sample is in the rejected percentile), return \texttt{False}; the sample is silently skipped and the dataloader advances to the next sample in the shard.
    \item If all active filters pass, return \texttt{True}; the sample proceeds to tokenization, packing, and batching as normal.
\end{enumerate}
For multi-filter configurations (e.g., applying both a CLIP-score filter and a text-quality filter simultaneously), the predicates are composed with logical AND: a sample must pass \emph{all} active filters to be retained.
Since the predicate evaluation involves only a dictionary lookup and a scalar comparison per filter, the per-sample cost is on the order of microseconds---negligible compared to the millisecond-scale cost of image decoding, tokenization, and GPU transfer.

Switching between filtering configurations requires changing only the filter names and percentile values in a configuration file; no data needs to be resharded or copied.
This enables rapid iteration: practitioners can launch dozens of filtering experiments against the same on-disk pool, varying filters, thresholds, and combinations, without any storage overhead beyond the original pool and its sidecar metadata.

\subsubsection{Overhead Measurement}\label{sec:exp_online_filtering}

A natural concern is that high rejection rates will cause the dataloader to ``spin''---reading and discarding many samples before finding one that passes the filter, thereby starving the GPUs.
We benchmark this directly on 2 nodes (8 A100 GPUs), with per-device batch size 4 and effective sequence length 8192, across two representative datasets with different characteristics: ShareGPT4V (high-quality synthetic captions, relatively uniform scores) and MINT-HTML (web-crawled multimodal documents, highly variable scores).

\begin{figure}[t]
    \centering
    \includegraphics[width=\linewidth]{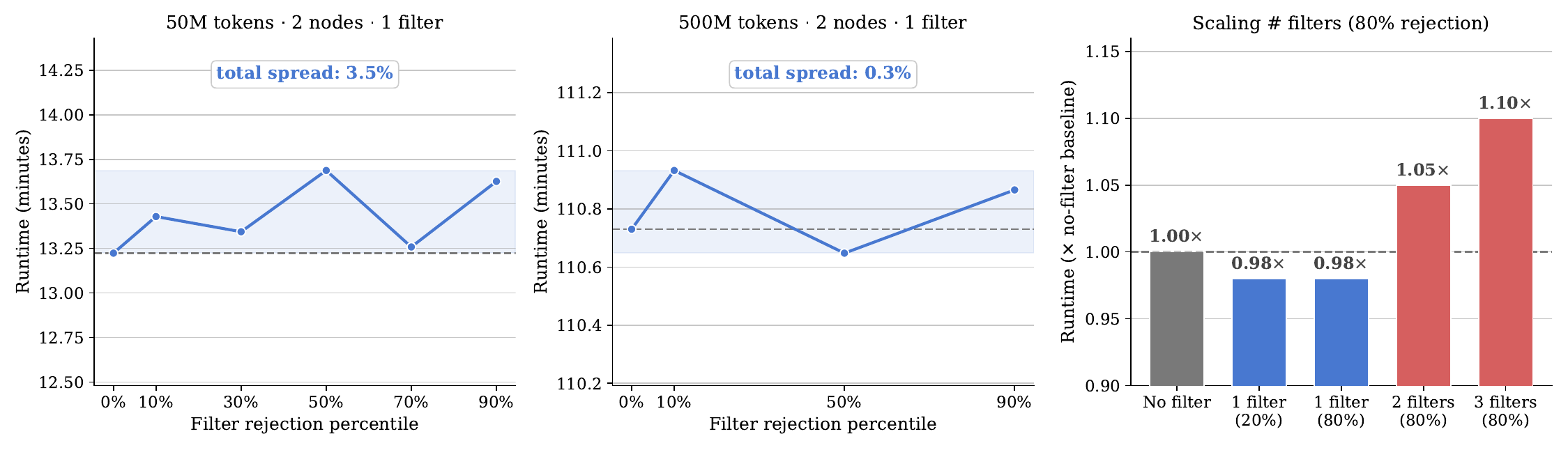}
    \caption{\textbf{Online filtering adds negligible overhead to training.}
    \textbf{Left:} Runtime for 50M tokens (2 nodes, 8 GPUs, 1 filter) across rejection percentiles---total spread is 3.5\%, within normal run-to-run variance.
    \textbf{Center:} Same measurement at 500M tokens, where the spread shrinks to 0.3\%, confirming that any per-sample overhead is amortized over longer runs.
    \textbf{Right:} Scaling the number of simultaneous filters at 80\% rejection each.
    Even with 3 filters applied concurrently, the overhead is only $1.10\times$ baseline, confirming that online filtering is a scalable approach.}
    \label{fig:online-filtering-overhead}
\end{figure}

\cref{fig:online-filtering-overhead} summarizes the results across three experimental axes.

First, when varying the rejection percentile of a single filter from 0\% to 90\% (\cref{fig:online-filtering-overhead}, left), runtime at 50M tokens remains within 3.5\% of the unfiltered baseline, and this spread shrinks to just 0.3\% at 500M tokens (\cref{fig:online-filtering-overhead}, center)---well within normal run-to-run variance.
The lack of degradation even at 90\% rejection (where 9 out of 10 samples are discarded) can be attributed to two factors: (i)~the predicate check is cheap relative to downstream processing, allowing the dataloader to skip rejected samples faster than the GPU can consume accepted ones; and (ii)~our models are more compute-bound than the CLIP models trained in DataComp, since every sample passes through both a vision encoder and a full language model---this leaves more headroom in the data-loading pipeline, making it naturally tolerant of the additional filtering overhead.

Second, when scaling the number of simultaneous filters at a fixed 80\% rejection rate per filter (\cref{fig:online-filtering-overhead}, right), applying 1--2 filters actually runs \emph{faster} than the unfiltered baseline ($0.98\times$), likely because rejected samples bypass the more expensive downstream stages (image decoding, pixel-shuffle, tokenization, packing).
Even with 3 concurrent filters---the most aggressive configuration we tested, where the compound rejection rate exceeds 99\%---the overhead is only $1.10\times$ the unfiltered baseline.

These results demonstrate that our online filtering is a practical and storage-efficient alternative to offline resharding for VLM pretraining at our scale, enabling rapid iteration over filtering configurations without materializing each filtered subset on disk.
\clearpage

\section{Fine-grained sweep of mixture optimization experiments}\label{app:data-mixing-fine-sweep}

In~\cref{sec:mixing} of the main paper, we conducted a full scaling grid across $3$ model sizes (1B, 2B, 4B), $3$ token budgets (6.25B, 12.5B, 25B) and $3$ mixing ratios (Instruction-heavy, Balanced, Caption-heavy). Here, we aim to do a finer-grained sweep over $9$ total mixtures. 
To that end, we conduct a dense line search between the orpotion of image-caption data and instruction-tuning data, which means that we gradually decrease the proportion of crawled Image-Text data in favour of Instruction Tuning data. We keep the multimodal docs and text proportion fixed at $5$\% and $15$\% respectively, and only vary the image-caption to instruction-tuning data proportion from $65$\% captioning-$15$\% instruction to $10$\% captioning-$70$\% instruction.

\begin{figure}[!h]
    \centering
    \includegraphics[width=0.9\linewidth]{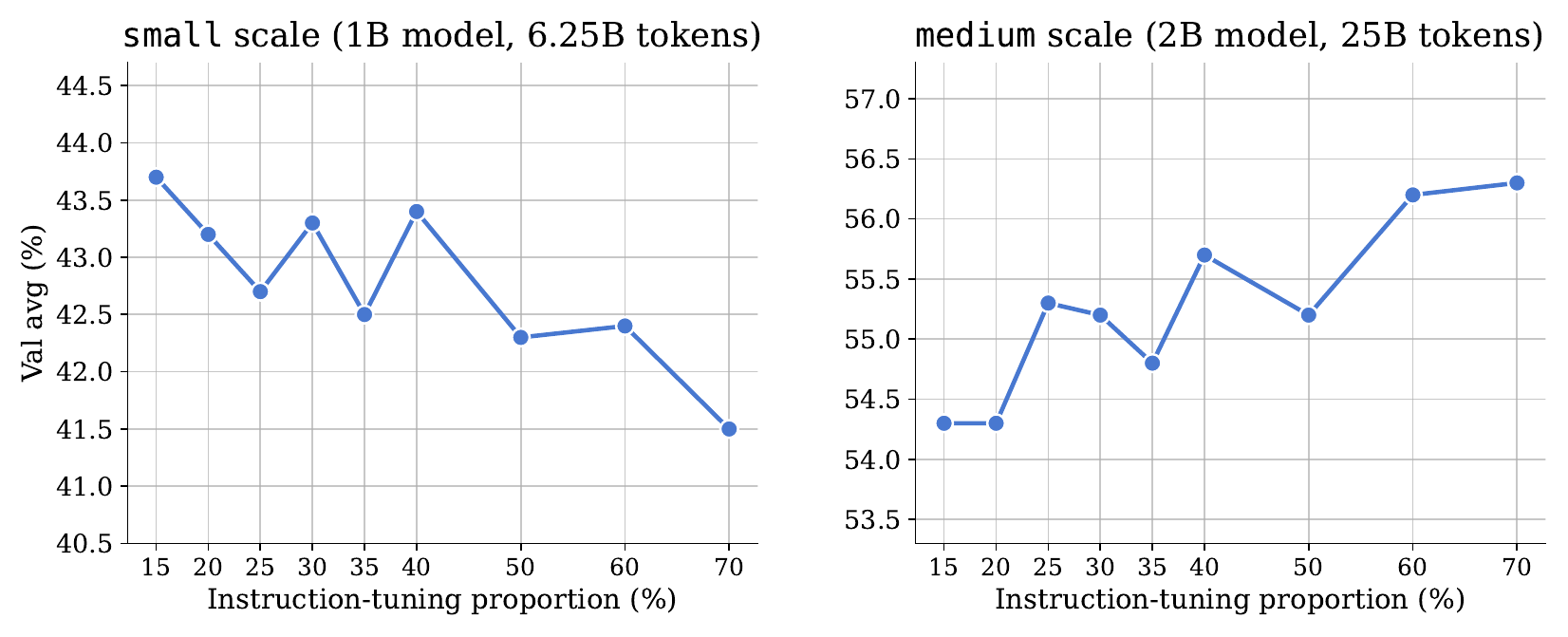}
    \caption{\textbf{The optimal pretraining data mixture is scale-dependent.} At the \texttt{small} scale, performance is flat or slightly degrades as the instruction-tuning share grows; at the \texttt{medium} scale, the trend reverses, with performance improving monotonically and peaking at $65$\% instruction-tuning data. This highlights the need for scale-aware data curation.}
    \label{fig:fine-mixing}
\end{figure}

In~\cref{sec:mixing} of the main paper, we conducted a full scaling grid across $3$ model sizes (1B, 2B, 4B), $3$ token budgets (6.25B, 12.5B, 25B), and $3$ mixing ratios (\textit{Instruction-heavy}, \textit{Balanced}, \textit{Caption-heavy}). Here we complement that grid with a finer-grained sweep over $9$ mixtures at two scales: \texttt{small} (1B model, 6.25B tokens) and \texttt{medium} (2B model, 25B tokens). Concretely, we conduct a dense line search between image-caption and instruction-tuning data gradually trading off image-text data for instruction-tuning data, while keeping the multimodal-document and text-only shares fixed at $5\%$ and $15\%$ respectively. The image-caption-to-instruction-tuning split is varied from $65\%$ caption / $15\%$ instruction (Caption-heavy mix in~\cref{sec:mixing}) down to $10\%$ caption / $70\%$ instruction (Instruction-heavy mix in~\cref{sec:mixing}) in 8 increments, yielding the $9$ operating points reported in~\cref{fig:fine-mixing}.

We find that the optimal data mixture at the \texttt{small} scale is the $65$\% caption-$15$\% instruction mix, and performance gracefully degrades as we increase the proportion of instruction-tuning data in the mixture. Conversely, the trend reverses for the \texttt{medium} experiment, wherein the best mixture ratio is exactly at the opposite operating point of $10$\% caption-$70$\% instruction. This highlights the importance of \textit{scale-aware data-curation}---strategies that are optimal at one scale need not transfer to larger ones, and hence one must be prudent while making such transfer decisions, which has also been discussed in prior works~\citep{goyal2024scaling,mizrahi2025language,maithinking1}.

\clearpage

\section{Packing Implementation Details}\label{app:packing}

Our pretraining corpus mixes data types whose sequence lengths span several orders of magnitude---from single image-caption pairs of a few hundred tokens to multimodal documents and multi-turn instruction conversations that approach the full context window.
Training on such data without packing forces every example to be padded to a common length, wasting a large fraction of every batch on padding tokens that carry no gradient.
The standard remedy is \emph{sequence packing}: concatenating several short samples into one fixed-length training example so that almost every token is a real token.
For vision-language models, however, packing must respect a constraint that text-only packing does not: each image is split by dynamic tiling into a variable number of tiles, and every tile expands into a fixed block of visual tokens that the vision encoder must process.
A pack is therefore bounded by \emph{two} budgets at once, and we pack against both.

\subsection{Dual constraints}

Each emitted training example must simultaneously satisfy:
\begin{itemize}[leftmargin=*, itemsep=2pt]
    \item \textbf{Token budget} $L = 8192$: the total number of (text $+$ visual) tokens may not exceed the LLM context length we train at.
    \item \textbf{Image-tile budget} $M = 24$: the total number of image tiles---summed over all images of all samples in the pack---may not exceed $M$. This caps the vision-encoder footprint of a single forward pass and is the constraint unique to the multimodal setting.
\end{itemize}
A pack is considered \emph{complete} as soon as it reaches \emph{either} budget exactly, since a single text-heavy multimodal document can saturate $L$ with few images, while a gallery of single-tile thumbnails can saturate $M$ while leaving $L$ far from full.

\begin{figure}[!h]
    \centering
    \includegraphics[width=\linewidth]{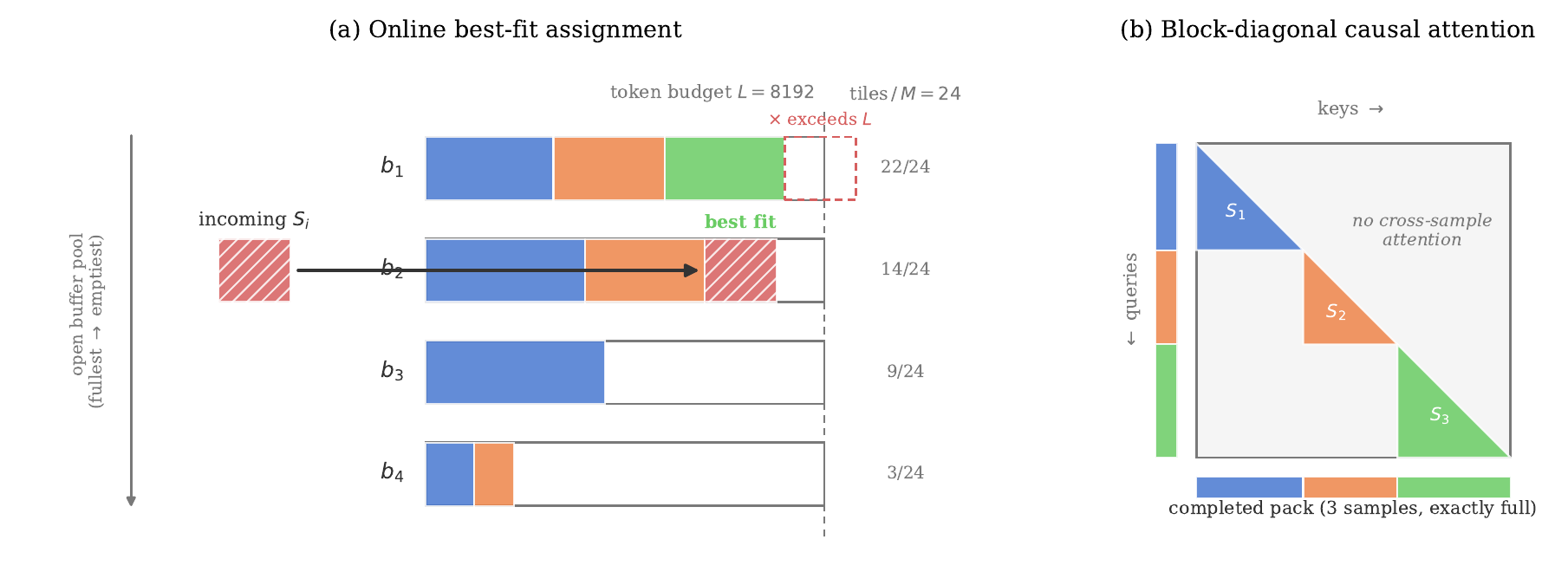}
    \caption{\textbf{Streaming best-fit sequence packing.}
    \textbf{(a)}~Incoming samples are assigned online to a pool of open buffers kept sorted from fullest to emptiest. A sample $S_i$ is placed in the \emph{first} (hence fullest) buffer whose token budget $L$ \emph{and} image-tile budget $M$ both still accommodate it. Here $b_1$ would overflow $L$, so $S_i$ lands in $b_2$. A buffer is emitted once it hits either budget exactly (or when the pool overflows its size cap).
    \textbf{(b)}~Within a completed pack, a variable-length attention mask enforces block-diagonal causal attention, so each packed sample attends only within its own segment and position ids restart per segment---numerically identical to unpacked training.}
    \label{fig:packing-schematic}
\end{figure}

\subsection{Streaming best-fit bin-packing}

Samples are produced one at a time by the WebDataset mixing stream described in \cref{appsub:online_filtering}---already filtered, templated, and tokenized---so we never see the full dataset and cannot solve the offline (optimal) bin-packing problem.
We instead maintain a pool of \emph{open buffers}, each a partially-filled pack, and assign samples greedily as they arrive (\cref{fig:packing-schematic}a).
The pool is kept sorted in order of \emph{decreasing} image-tile occupancy (fullest first), an invariant we exploit to make assignment a single linear scan.
For each incoming sample $S_i$:
\begin{enumerate}[leftmargin=*, itemsep=2pt]
    \item \textbf{Best-fit search.} Scan the pool from fullest to emptiest and select the \emph{first} buffer $b$ for which adding $S_i$ would violate neither budget, i.e.\ $\mathrm{tiles}(b) + \mathrm{tiles}(S_i) \le M$ \emph{and} $\mathrm{tokens}(b) + \mathrm{tokens}(S_i) \le L$. Because the pool is fullest-first, this returns the most-occupied pack that can still host $S_i$---a best-fit choice that drives buffers toward completion rather than leaving many half-full.
    \item \textbf{Assignment.} If such a buffer is found, $S_i$ is concatenated into it. Otherwise $S_i$ seeds a fresh singleton buffer. Each token of $S_i$ is tagged with a within-pack sample index (\texttt{data\_index}), recording which original sample it belongs to (used below to prevent cross-sample attention).
    \item \textbf{Re-insertion.} The (possibly grown) buffer is re-inserted at the position that restores the fullest-first ordering.
\end{enumerate}

A buffer is emitted as a finished training example under one of two conditions:
\begin{itemize}[leftmargin=*, itemsep=2pt]
    \item \textbf{Exactly full:} it reaches either budget exactly ($\mathrm{tiles}=M$ or $\mathrm{tokens}=L$). If this happens, these are the fully well-packed examples and are emitted immediately, fullest first.
    \item \textbf{Pool overflow:} if the number of open buffers exceeds a cap $B = 20$, the fullest buffer is force-emitted even if not exactly full. This bounds the memory and latency cost of holding many partially-filled packs and is the only source of imperfectly-filled examples in practice.
\end{itemize}
No explicit padding is introduced during packing itself. The only padding is whatever the collator adds to align tensors within a micro-batch, and is compensated for by per-sample loss weighting (see details below).

\subsection{Preventing cross-sample contamination}

A single pack contains several independent samples that must not attend to one another, or the model would condition predictions on unrelated context.
We enforce this at the attention level rather than by re-padding.
The per-token \texttt{data\_index} tags are converted at collation time into cumulative segment boundaries (\texttt{cu\_seqlens}) and passed---in place of a dense attention mask---into a patched attention kernel (\texttt{flash\_attn\_varlen\_func}) for the LM backbone.
This yields exact \textbf{block-diagonal attention} (\cref{fig:packing-schematic}b): every sample attends only within its own segment, and position ids restart at zero at each segment boundary, so a packed sample is numerically identical to the same sample trained in isolation.
Finally, because samples within a pack have very different lengths, we re-weight the per-token loss by a function of each segment's effective (loss-bearing) token count, so that a single long sample does not dominate the pack's gradient relative to the other short samples in a pack.

\subsection{Stateful and resumable packing}

Packing is both streaming and stateful: at any instant a worker holds a pool of partially-filled buffers that are not derivable from the dataloader position alone.
To make training resumable without re-seeing or dropping data at a checkpoint boundary, each \texttt{(rank, worker)} pair persists its open buffer pool to disk (\texttt{rank<r>\_worker<w>\_buffers.pt}) alongside the WebDataset shard/sample/epoch cursors that record exactly how far each source has been consumed.
On resume, the pool and cursors are restored and packing continues from precisely the state it was checkpointed in.
This buffer state is written as part of the same \texttt{dataset\_state} mechanism used for dataloader resumption.

In all our runs we pack to $L = 8192$ tokens and $M = 24$ image tiles, with a pool cap of $B = 20$ open buffers.

\clearpage

\section{Annotation Infrastructure}\label{app:annotation}

To run our filtering experiments, we need to \emph{annotate} all samples in our data pool with the respective filtering metric---for every sample, and for every filter, we compute a scalar score for each filter (e.g., compute the CLIP score or the decontamination score according to a set of reference images).
At our scale, this amounts to scoring trillions of tokens under many filters whose resource profiles range from cheap CPU-only text-based classifiers (e.g., length) to GPU-bound encoders (e.g., CLIP score).
We build a Ray-based\footnote{\href{https://github.com/ray-project/ray}{https://github.com/ray-project/ray}} annotation framework that runs inside a Slurm allocation, parallelizes across nodes, and tolerates common failures (decode errors, OOMs, transient file-system issues, Slurm preemption) that any multi-day multi-node annotation job encounters.
We plan to release the framework as part of the DCVLM benchmark.

\subsection{Pipeline and Actor Model}\label{app:annotation-pipeline}

A run is parameterised by an input directory of WebDataset~\citep{webdataset} shards and a list of filter names.
Each filter declares its per-actor CPU, GPU, and memory requirements via a \texttt{BaseFilter} interface.
Because a worker runs the selected filters sequentially on a shared decoded batch, the framework sizes the actor pool from the maximum declared requirement across filters on each axis (CPU, GPU, DRAM).
A run proceeds in three overlapping phases:

\begin{enumerate}[leftmargin=*,itemsep=1pt]
\item \textbf{Extraction.} One Ray task per tar reads the raw bytes (without decoding any sample components like images, json files etc.), groups samples into batches of $B$ samples, and places each batch in the Ray object store.
Decoding is deferred so that PIL images never round-trip through the object store, which keeps inter-node traffic to compressed bytes only.
\item \textbf{Processing.} A fixed pool of \texttt{FilterExecutionWorker} actors is created once at the start of the run. 
For each batch, the actor decodes once and runs every filter on the decoded samples in sequence, returning per-sample results and per-filter statistics.
The actor model where each actor has some form of state is important.
Loading the scoring model dominates per-batch cost for most filters, so we want to pay for it once per actor rather than once per batch, and decoding once per batch lets multi-filter runs share work.
\item \textbf{Streaming writes.} As soon as all batches for a tar resolve, we schedule a write task.
This decouples write IO from active extraction and inference, and enables continuation after a Slurm interruption.
\end{enumerate}

We support two output modes.
In \texttt{--only-write-metadata} mode, each input tar \texttt{xxxxx.tar} creates a sibling \texttt{xxxxx.json} mapping every sample key to a flat dict of filter scores.
The original shards are left untouched.
In the alternative full-rewrite mode, the same scores are embedded as a \texttt{filters.json} field inside a freshly written tar.
This is convenient when a downstream consumer wants a self-contained WebDataset.

\subsection{Backpressure and Fault Tolerance}\label{app:annotation-fault-tolerance}

A \texttt{max\_batches\_in\_flight} cap bounds the number of outstanding processing tasks, which in turn bounds the working set of the Ray object store.
Without it, extraction outruns the GPU pool by orders of magnitude and PIL-decoded images blow up RAM (and trigger costly object-store spilling).

We tolerate failures at four levels with intentionally minimal recovery logic.
(i) Sample decode errors are retried $N$ times then skipped or raised per a configurable handler.
(ii) A batch that raises during filtering is resubmitted up to \texttt{max\_resubmits} times (default $5$); the actor is health-checked via a \texttt{ping} before being returned to the idle pool.
(iii) If an actor dies (\texttt{RayActorError}, e.g.\ kernel OOM-kill or a CUDA fault), the in-flight task is resubmitted and a best-effort replacement actor is spawned, so the pool does not shrink monotonically over the run.
(iv) At the job level, atomic .tmp+rename writes are combined with a startup pass that skips tars whose final output already exists, so an interrupted job can be resumed by simply re-running it.
The core assumption we rely on is that filters are deterministic, which makes duplicate computation under retry safe.
\clearpage

\section{Additional SFT Results}\label{app:sft}

The SFT control experiments in \cref{sec:controls} rely on two design choices that we fix once and reuse across all 54 runs: the SFT token budget and the peak learning rate. We first describe how we set both (\cref{app:sft-hparams}), and then verify that our conclusions are robust to the choice of SFT dataset (\cref{app:sft-dataset}).

\subsection{SFT Compute Budget and Learning-Rate Selection}\label{app:sft-hparams}

\smallsec{SFT token budget.} Rather than fine-tuning each checkpoint with an arbitrary fixed schedule, we scale the SFT budget proportionally to each checkpoint's pretraining budget. We estimate the SFT-to-pretraining token ratio of InternVL3~\citep{zhu2025internvl3} to be $\approx\!0.29$ and apply this multiplier to each of our three pretraining budgets, yielding SFT budgets of $1.8$B, $3.6$B, and $7.2$B tokens for checkpoints pretrained on $6.25$B, $12.5$B, and $25$B tokens, respectively. Holding this ratio fixed ensures that larger-compute checkpoints receive proportionally more fine-tuning, rather than coupling our conclusions to a single absolute SFT schedule.

\smallsec{Hyperparameter considerations.} For pretraining we use a fixed peak LR of $2\times10^{-5}$ across all scales, chosen via a sweep spanning three orders of magnitude. For SFT, common practice across recent VLMs is to keep the LR within the same order of magnitude as pretraining: InternVL-2.5~\citep{chen2024expanding} and Nemotron Nano-V2-VL~\citep{deshmukh2025nvidia} reuse the pretraining LR unchanged, MM1~\citep{mckinzie2024mm1} and MiMo-VL~\citep{coreteam2025mimovltechnicalreport} stay within the same order of magnitude (the latter increasing it slightly), and PerceptionLM~\citep{cho2025perceptionlm} reuses the LR but halves the global batch size, effectively doubling it under the linear-scaling rule. Only Molmo~\citep{deitke2025molmo} drops by a full order of magnitude (from $2\times10^{-4}$ to $1\times10^{-5}$), and additionally uses separate LRs for the language model, projector, and vision encoder.

\smallsec{Learning-rate sweep.} Guided by this, we keep the global batch size fixed at $1024$ (matching pretraining) and sweep the SFT peak LR over $\{4\times10^{-6},\, 2\times10^{-5},\, 4\times10^{-5}\}$, i.e.\ one order of magnitude around the pretraining value. All runs use a cosine-decay schedule with $3\%$ warmup and weight decay $0.01$. We sweep on two representative $2$B checkpoints---both pretrained on the Instruction-heavy mix that is optimal at this scale---one pretrained for $6.25$B tokens (fine-tuned for $1.8$B SFT tokens) and one for $12.5$B tokens (fine-tuned for $3.6$B), using LLaVA-665K~\citep{liu2023visual} as the SFT dataset.

\cref{tab:sft-lr-sweep} reports the results. The lowest LR, $4\times10^{-6}$, is best on Core Avg at both checkpoints, while the highest LR, $4\times10^{-5}$, is consistently worst---most visibly on Text ($-4.4$pp and $-4.5$pp relative to $4\times10^{-6}$ at the $6.25$B and $12.5$B checkpoints), indicating that an overly aggressive SFT LR erodes the language capabilities established during pretraining. The pretraining LR ($2\times10^{-5}$) is a close runner-up. We therefore adopt a peak LR of $4\times10^{-6}$ for all SFT runs in this work. Reassuringly, SFT lifts every checkpoint above its pretrained starting point at the chosen LR ($+2.4$ and $+2.8$), without altering the underlying pretraining ranking.

\begin{table}[!h]
    \centering
    \caption{\textbf{SFT learning-rate sweep.} We fine-tune two $2$B checkpoints (pretrained on the Instruction-heavy mix for $6.25$B and $12.5$B tokens) with LLaVA-665K~\citep{liu2023visual} at three peak LRs, keeping the global batch size fixed at $1024$. The lowest LR ($4\times10^{-6}$) is best at both checkpoints, while the highest LR degrades Text performance. Best Core Avg per block in \textbf{bold}. Categories follow \cref{tab:main-results}.}
    \label{tab:sft-lr-sweep}
    \small
    \begin{tabular}{lccccccc}
        \toprule
        \textbf{Setting} & \textbf{Gen} & \textbf{Know} & \textbf{OCR} & \textbf{Vision} & \textbf{MTL} & \textbf{Text} & \textbf{Core Avg} \\
        \midrule
        \rowcolor{gray!8} \multicolumn{8}{l}{\texttt{2B} model, $6.25$B pretraining tokens $\rightarrow$ $1.8$B SFT tokens} \\
        Pretrained (no SFT)             & 55.1 & 56.9 & 41.0 & 45.0 & 39.6 & 47.2 &  47.9 \\
        \quad + SFT, LR $4\times10^{-6}$ & 59.9 & 60.5 & 43.1 & 45.9 & 43.8 & 47.2 & \textbf{50.3} \\
        \quad + SFT, LR $2\times10^{-5}$ & 61.3 & 60.5 & 42.4 & 45.7 & 43.6 & 45.0  & 49.9 \\
        \quad + SFT, LR $4\times10^{-5}$ & 59.5 & 58.3 & 40.3 & 44.1 & 41.6 & 42.8 & 47.8 \\
        \midrule
        \rowcolor{gray!8} \multicolumn{8}{l}{\texttt{2B} model, $12.5$B pretraining tokens $\rightarrow$ $3.6$B SFT tokens} \\
        Pretrained (no SFT)             & 59.0 & 59.0 & 43.9 & 45.1 & 42.4 & 48.1  & 50.0 \\
        \quad + SFT, LR $4\times10^{-6}$ & 63.8 & 62.1 & 48.6 & 47.6 & 44.4 & 48.0 & \textbf{52.8} \\
        \quad + SFT, LR $2\times10^{-5}$ & 64.5 & 61.1 & 46.4 & 47.8 & 45.9 & 45.6 & 52.0 \\
        \quad + SFT, LR $4\times10^{-5}$ & 63.8 & 59.9 & 44.8 & 47.5 & 44.6 & 43.5  & 50.8 \\
        \bottomrule
    \end{tabular}
\end{table}

\subsection{Robustness to the SFT Dataset}\label{app:sft-dataset}

\begin{figure}[!h]
    \centering
    \includegraphics[width=0.8\linewidth]{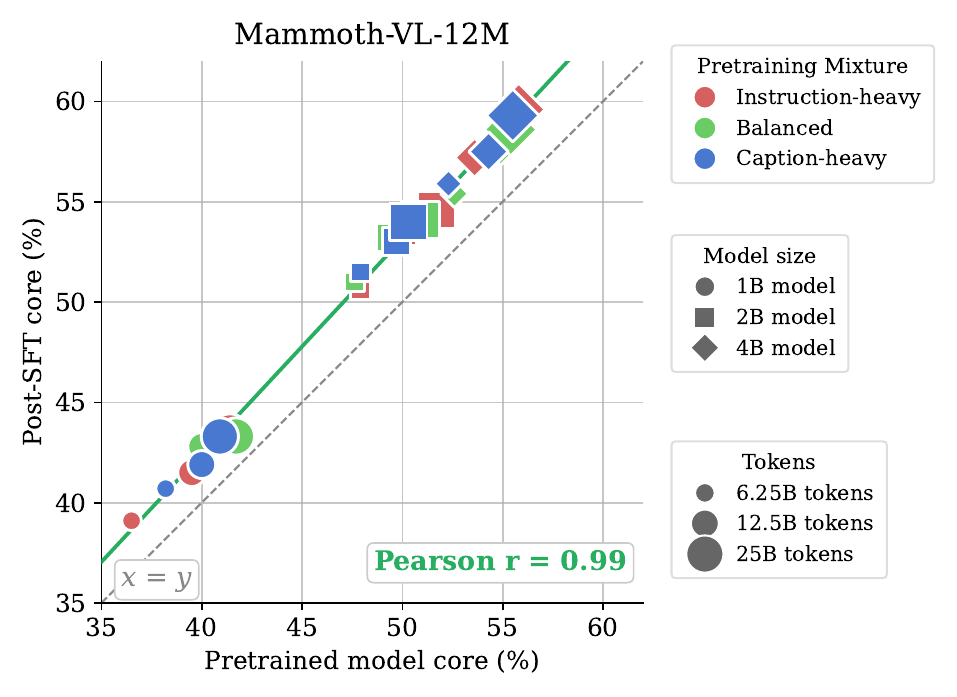}
    \caption{\textbf{Pretraining performance predicts post-SFT performance (\textit{cont.}).} We extend the analysis of \cref{sec:controls} (\cref{fig:sft-correlation}) to another SFT dataset, i.e., Mammoth-VL-12M \cite{guo2025mammoth}. Remarkably, this leaves our main observation that \textit{pretraining predicts post-SFT performance} unchanged.}
    \label{fig:sft-correlation-mammoth}
\end{figure}

\Cref{fig:sft-correlation-mammoth} replicates the analysis of \Cref{sec:controls} using Mammoth-VL-12M~\citep{guo2025mammoth} as the SFT dataset in place of LLaVA-665K \cite{li2024llava}.
Results are fully consistent: pretraining and post-SFT scores remain near-perfectly correlated (Pearson $r{=}0.99$; Spearman $\rho{=}0.99$), and the pretraining ranking is preserved across all 27 checkpoints.
These results confirm that the conclusion is robust to the choice of SFT dataset: a stronger pretraining checkpoint consistently yields a stronger fine-tuned model.
In particular, the advantage of Instruction-heavy pretraining mixtures is \textit{not} diminished by the additional instruction-tuning signal introduced during SFT.

\clearpage

\clearpage

\section{Extended Suite Evaluations}\label{app:extended-evals}

\Cref{tab:extended-results} reports results on our 52-task Extended evaluation suite, which augments the Core suite with Hallucination (\textbf{Hall}), Safety (\textbf{Safety}), and Reasoning (\textbf{Reason}) categories.

\textsc{\dcvlm-Baseline} maintains its overall advantage over \textsc{FineVision} at both scales, achieving $+3.0$pp and $+3.9$pp gains at \texttt{large} and \texttt{x-large}, respectively. Gains are most pronounced in Vision-Centric ($+23.1$pp / $+21.3$pp) and General Understanding ($+8.4$pp / $+8.5$pp) tasks, consistent with the Core suite trends. 

Regarding newly added categories, Hallucination scores also favor \textsc{\dcvlm-Baseline} at both scales ($+3.4$pp / $+3.0$pp), suggesting our data mixture yields better-calibrated visual grounding.
However, two notable weaknesses emerge. 
First, \textsc{\dcvlm-Baseline} substantially underperforms on Safety ($-17.7$pp / $-19.2$pp across both scales). Second, \textsc{FineVision} retains an advantage on OCR ($+6.5$pp / $+5.2$pp) and Reasoning ($+4.3$pp / $+4.9$pp), suggesting its document-rich sources benefit reading and understanding details, text, figures, and equations in images. 
We leave the exploration of how to address these gaps, particularly regarding safety alignment, for future work.

\begin{table}[h!]
\caption{\textbf{Extended} benchmark results comparing \textsc{\dcvlm-Baseline} to \textsc{FineVision} at \texttt{large} (top) and \texttt{x-large} (bottom) scales. Category acronyms: \textbf{Gen} = General -- \textbf{Know} = Knowledge-Centric -- \textbf{OCR} = OCR \& Charts -- \textbf{Vision} = Vision-Centric -- \textbf{MTL} = Multilingual -- \textbf{Hall} = Hallucination -- \textbf{Text} = Text-Only -- \textbf{Reason} = Reasoning.}\label{tab:extended-results}
\centering
\resizebox{\linewidth}{!}{
\begin{tabular}{lcccccccccc}
\toprule
\textbf{Dataset}  & \textbf{Gen} & \textbf{Know} & \textbf{OCR} & \textbf{Vision} & \textbf{MTL} & \textbf{Hall} & \textbf{Safety} & \textbf{Text} & \textbf{Reason} & \textbf{Ext Avg} \\
\midrule
\rowcolor{gray!8} \multicolumn{11}{l}{\texttt{large} scale (4B model, 100B tokens)} \\
\textsc{FineVision} & 60.7 & 70.7 & \textbf{65.2} & 30.9 & 45.1 & 65.6 & 39.8 & 56.3 & 38.9 & 53.0 \\
\textsc{\dcvlm-Baseline} (ours) & 69.1 & 67.6 & 58.7 & 54.0 & 50.9 & \textbf{69.0} & 22.1 & 59.7 & 34.6 & 56.0 \\
\midrule
\rowcolor{gray!8} \multicolumn{11}{l}{\texttt{x-large} scale (8B model, 200B tokens)} \\
\textsc{FineVision} & 65.3 & 72.8 & 64.1 & 39.2 & 48.4 & 65.9 & \textbf{43.5} & 60.4 & \textbf{44.4} & 56.6 \\
\textsc{\dcvlm-Baseline} (ours) & \textbf{73.8} & \textbf{73.0} & 58.9 & \textbf{60.5} & \textbf{56.1} & 68.9 & 24.3 & \textbf{67.1} & 39.5 & \textbf{60.5} \\
\bottomrule
\end{tabular}
}
\end{table}

\clearpage

\section{Safety Preprocessing: Harmful, Toxic and Biased Content}\label{app:safety}
As with all large uncurated data-pools (largely sourced from the web), especially for ones that are fully open-sourced, there can be concerns regarding releasing harmful and unsafe content openly. We invested some efforts into ensuring that our raw data-pool has as little unsafe content as possible, via two main analysis methods---(1) Text-based removal and (2) Image-level Safety Analysis.
\subsection{Unsafe, Biased and Toxic Text Removal}
To ensure data quality and safety, we filter unsafe, biased, and toxic content in DCVLM pool using Detoxify \cite{Detoxify}, a transformer-based multilingual toxicity classifier. The classifier produces scores across multiple harm categories which include toxicity, severe toxicity, obscenity, threat, insult and identity attack. We retain a sample only if every category score falls at or below a threshold of $0.1$, following the filtering criteria used in DataComp-CLIP \cite{gadre2023datacomp}. 

\subsection{Image-level Safety Analysis}
We analyze a subset of $20,000$ image–text pairs to assess image safety. For this purpose, we employ ShieldGemma2 \cite{zeng2025shieldgemma2}, a 4B-parameter image content moderation model built on Gemma 3, which provides robust safety risk predictions across multiple categories, including sexually explicit content, violence and gore, and dangerous content. Within the sampled dataset, the model classifies $19$ images as dangerous, $117$ as sexually explicit, and $4$ as containing violence and gore, each with a confidence score of at least $90\%$. This suggests that potentially unsafe content represents only a small fraction of the dataset, indicating a relatively low overall presence of harmful imagery and further supports using the analyzed subset as a representative estimate of the safety characteristics of the full dataset.
\clearpage

\section{Per-Dataset Core Results}\label{app:per-dataset-evals}

For completeness and ease of inspection, we report the per-dataset scores underlying the per-category Core averages of \cref{tab:main-results}. Each table below corresponds to one of our six Core categories and lists every constituent benchmark for the same set of models and scales as \cref{tab:main-results}. Averaging a row's datasets within a table recovers that row's per-category number in \cref{tab:main-results}, and averaging across all 33 datasets (six tables) recovers its Core Avg. 

\begin{table}[ht]
\centering
\caption{\textbf{Per-dataset Core results: General Understanding (Gen).} Per-dataset breakdown of the \textbf{Gen} column of \cref{tab:main-results}. The final \textbf{Gen} column is the row mean over these 7 datasets and reproduces \cref{tab:main-results}. \textbf{Bold} = best per scale.}
\label{tab:per-dataset-general}
\resizebox{\linewidth}{!}{
\begin{tabular}{lcccccccc}
\toprule
\textbf{Method} & \textbf{MMB-EN} & \textbf{MMB-CN} & \textbf{MMB-EN$_{\text{v1.1}}$} & \textbf{MMB-CN$_{\text{v1.1}}$} & \textbf{SEED-IMG} & \textbf{MMT} & \textbf{VMC} & \textbf{Gen} \\
\midrule
\rowcolor{gray!8} \multicolumn{9}{l}{\texttt{small} scale (1B model, 6.25B tokens)} \\
\textsc{LLaVA-OneVision-1.5} & 12.0 & 22.3 & 11.5 & 21.7 & 35.1 & 22.5 & 31.9 & 22.4 \\
\textsc{Nemotron-VL-2} & 16.4 & 5.4 & 14.0 & 4.6 & 38.6 & 29.2 & 31.7 & 20.0 \\
\textsc{FineVision} & \textbf{39.2} & 34.3 & 36.7 & 33.2 & \textbf{50.7} & \textbf{39.5} & 47.4 & 40.1 \\
\textsc{\dcvlm-Baseline} (ours) & 38.7 & \textbf{38.3} & \textbf{36.9} & \textbf{37.2} & 46.6 & 38.3 & \textbf{47.5} & \textbf{40.5} \\
\midrule
\rowcolor{gray!8} \multicolumn{9}{l}{\texttt{medium} scale (2B model, 25B tokens)} \\
\textsc{LLaVA-OneVision-1.5} & 29.9 & 31.1 & 29.1 & 30.4 & 44.2 & 29.9 & 38.2 & 33.3 \\
\textsc{Nemotron-VL-2} & 49.2 & 48.5 & 46.4 & 47.0 & 62.1 & 41.1 & 46.2 & 48.6 \\
\textsc{FineVision} & 59.3 & 49.8 & 57.4 & 47.3 & \textbf{64.8} & 47.4 & 61.2 & 55.3 \\
\textsc{\dcvlm-Baseline} (ours) & \textbf{66.2} & \textbf{62.6} & \textbf{64.2} & \textbf{60.9} & 64.5 & \textbf{51.6} & \textbf{66.0} & \textbf{62.3} \\
\midrule
\rowcolor{gray!8} \multicolumn{9}{l}{\texttt{large} scale (4B model, 100B tokens)} \\
\textsc{Nemotron-VL-2} & 27.1 & 26.3 & 21.8 & 23.1 & 42.5 & 35.2 & 44.6 & 31.5 \\
\textsc{FineVision} & 67.1 & 53.4 & 64.0 & 51.5 & 66.5 & 42.7 & 67.5 & 59.0 \\
\textsc{\dcvlm-Baseline} (ours) & \textbf{72.2} & \textbf{71.2} & \textbf{69.3} & \textbf{68.0} & \textbf{71.5} & \textbf{57.8} & \textbf{69.1} & \textbf{68.4} \\
\midrule
\rowcolor{gray!8} \multicolumn{9}{l}{\texttt{x-large} scale (8B model, 200B tokens)} \\
\textsc{FineVision} & 67.3 & 69.2 & 63.9 & 66.1 & 66.0 & 46.2 & 65.6 & 63.5 \\
\textsc{\dcvlm-Baseline} (ours) & \textbf{79.0} & \textbf{75.2} & \textbf{76.7} & \textbf{73.0} & \textbf{74.1} & \textbf{59.7} & \textbf{73.4} & \textbf{73.0} \\
\bottomrule
\end{tabular}
}
\end{table}

\begin{table}[ht]
\centering
\caption{\textbf{Per-dataset Core results: Knowledge-Centric (Know).} Per-dataset breakdown of the \textbf{Know} column of \cref{tab:main-results}. The final \textbf{Know} column is the row mean over these 4 datasets and reproduces \cref{tab:main-results}. \textbf{Bold} = best per scale.}
\label{tab:per-dataset-knowledge}
\resizebox{0.8\linewidth}{!}{
\begin{tabular}{lccccc}
\toprule
\textbf{Method} & \textbf{MMMU} & \textbf{SciQA} & \textbf{AI2D} & \textbf{AI2D$_{\text{nm}}$} & \textbf{Know} \\
\midrule
\rowcolor{gray!8} \multicolumn{6}{l}{\texttt{small} scale (1B model, 6.25B tokens)} \\
\textsc{LLaVA-OneVision-1.5} & 21.4 & 40.7 & 38.2 & 39.0 & 34.8 \\
\textsc{Nemotron-VL-2} & 24.7 & 48.9 & 42.1 & 43.0 & 39.7 \\
\textsc{FineVision} & 35.7 & \textbf{58.1} & \textbf{43.3} & \textbf{45.5} & \textbf{45.7} \\
\textsc{\dcvlm-Baseline} (ours) & \textbf{36.8} & 52.4 & 42.0 & 43.4 & 43.6 \\
\midrule
\rowcolor{gray!8} \multicolumn{6}{l}{\texttt{medium} scale (2B model, 25B tokens)} \\
\textsc{LLaVA-OneVision-1.5} & 25.6 & 48.4 & 47.2 & 51.0 & 43.0 \\
\textsc{Nemotron-VL-2} & 28.2 & 65.1 & 59.3 & 65.6 & 54.5 \\
\textsc{FineVision} & 42.7 & \textbf{72.2} & \textbf{63.0} & \textbf{72.7} & \textbf{62.7} \\
\textsc{\dcvlm-Baseline} (ours) & \textbf{44.4} & 70.6 & 58.5 & 68.5 & 60.5 \\
\midrule
\rowcolor{gray!8} \multicolumn{6}{l}{\texttt{large} scale (4B model, 100B tokens)} \\
\textsc{Nemotron-VL-2} & 32.7 & 65.2 & 54.6 & 62.7 & 53.8 \\
\textsc{FineVision} & 40.9 & \textbf{89.5} & \textbf{69.7} & \textbf{82.6} & \textbf{70.7} \\
\textsc{\dcvlm-Baseline} (ours) & \textbf{47.1} & 75.8 & 68.2 & 79.5 & 67.7 \\
\midrule
\rowcolor{gray!8} \multicolumn{6}{l}{\texttt{x-large} scale (8B model, 200B tokens)} \\
\textsc{FineVision} & 46.1 & \textbf{87.4} & 73.3 & 84.4 & 72.8 \\
\textsc{\dcvlm-Baseline} (ours) & \textbf{51.1} & 79.5 & \textbf{75.1} & \textbf{86.3} & \textbf{73.0} \\
\bottomrule
\end{tabular}
}
\end{table}

\begin{table}[ht]
\centering
\caption{\textbf{Per-dataset Core results: OCR \& Charts (OCR).} Per-dataset breakdown of the \textbf{OCR} column of \cref{tab:main-results}. The final \textbf{OCR} column is the row mean over these 6 datasets and reproduces \cref{tab:main-results}. \textbf{Bold} = best per scale.}
\label{tab:per-dataset-ocr}
\resizebox{\linewidth}{!}{
\begin{tabular}{lccccccc}
\toprule
\textbf{Method} & \textbf{ChartQA} & \textbf{OCRBench} & \textbf{InfoVQA} & \textbf{SEED2+} & \textbf{OCRVQA} & \textbf{CharXiv} & \textbf{OCR} \\
\midrule
\rowcolor{gray!8} \multicolumn{8}{l}{\texttt{small} scale (1B model, 6.25B tokens)} \\
\textsc{LLaVA-OneVision-1.5} & 0.0 & 17.4 & 0.0 & 27.1 & 0.0 & 4.7 & 8.2 \\
\textsc{Nemotron-VL-2} & 0.2 & 2.6 & 0.2 & 30.6 & 8.5 & 5.2 & 7.9 \\
\textsc{FineVision} & \textbf{35.3} & \textbf{53.2} & \textbf{31.2} & \textbf{45.7} & 35.6 & 9.1 & \textbf{35.0} \\
\textsc{\dcvlm-Baseline} (ours) & 22.9 & 43.6 & 28.8 & 42.7 & \textbf{44.2} & \textbf{15.6} & 33.0 \\
\midrule
\rowcolor{gray!8} \multicolumn{8}{l}{\texttt{medium} scale (2B model, 25B tokens)} \\
\textsc{LLaVA-OneVision-1.5} & 0.0 & 59.1 & 0.0 & 36.1 & 0.0 & \textbf{30.8} & 21.0 \\
\textsc{Nemotron-VL-2} & 0.0 & 45.3 & 0.1 & 48.2 & 7.6 & 18.3 & 19.9 \\
\textsc{FineVision} & \textbf{57.4} & \textbf{75.6} & \textbf{57.3} & 57.7 & \textbf{48.9} & 14.1 & \textbf{51.8} \\
\textsc{\dcvlm-Baseline} (ours) & 44.2 & 63.3 & 41.8 & \textbf{57.8} & 46.5 & 21.1 & 45.8 \\
\midrule
\rowcolor{gray!8} \multicolumn{8}{l}{\texttt{large} scale (4B model, 100B tokens)} \\
\textsc{Nemotron-VL-2} & 1.3 & 45.2 & 3.3 & 44.7 & 26.9 & 20.4 & 23.6 \\
\textsc{FineVision} & \textbf{67.1} & \textbf{82.4} & \textbf{69.9} & \textbf{63.6} & 42.6 & \textbf{27.8} & \textbf{58.9} \\
\textsc{\dcvlm-Baseline} (ours) & 49.0 & 75.6 & 47.1 & 63.0 & \textbf{65.6} & 24.0 & 54.0 \\
\midrule
\rowcolor{gray!8} \multicolumn{8}{l}{\texttt{x-large} scale (8B model, 200B tokens)} \\
\textsc{FineVision} & \textbf{76.4} & \textbf{83.8} & \textbf{72.7} & 65.0 & 17.9 & 29.3 & \textbf{57.5} \\
\textsc{\dcvlm-Baseline} (ours) & 34.8 & 66.9 & 53.3 & \textbf{66.0} & \textbf{69.9} & \textbf{29.9} & 53.5 \\
\bottomrule
\end{tabular}
}
\end{table}

\begin{table}[ht]
\centering
\caption{\textbf{Per-dataset Core results: Vision-Centric (Vision).} Per-dataset breakdown of the \textbf{Vision} column of \cref{tab:main-results}. The final \textbf{Vision} column is the row mean over these 5 datasets and reproduces \cref{tab:main-results}. \textbf{Bold} = best per scale.}
\label{tab:per-dataset-vision-centric}
\resizebox{\linewidth}{!}{
\begin{tabular}{lcccccc}
\toprule
\textbf{Method} & \textbf{MMT-MI} & \textbf{BLINK} & \textbf{MUIR} & \textbf{CVB-2D} & \textbf{CVB-3D} & \textbf{Vision} \\
\midrule
\rowcolor{gray!8} \multicolumn{7}{l}{\texttt{small} scale (1B model, 6.25B tokens)} \\
\textsc{LLaVA-OneVision-1.5} & 22.8 & 16.1 & 23.1 & 28.7 & 48.2 & 27.8 \\
\textsc{Nemotron-VL-2} & 29.7 & 28.2 & \textbf{26.4} & 46.0 & 37.3 & 33.5 \\
\textsc{FineVision} & \textbf{39.6} & \textbf{39.2} & 23.0 & \textbf{48.5} & \textbf{54.5} & \textbf{41.0} \\
\textsc{\dcvlm-Baseline} (ours) & 38.9 & 38.2 & 26.3 & 43.0 & 49.1 & 39.1 \\
\midrule
\rowcolor{gray!8} \multicolumn{7}{l}{\texttt{medium} scale (2B model, 25B tokens)} \\
\textsc{LLaVA-OneVision-1.5} & 28.9 & 19.4 & 25.7 & 31.9 & 46.2 & 30.4 \\
\textsc{Nemotron-VL-2} & 41.3 & 33.1 & \textbf{33.3} & 51.4 & 46.2 & 41.1 \\
\textsc{FineVision} & 47.1 & 39.6 & 31.0 & \textbf{59.0} & 52.1 & 45.8 \\
\textsc{\dcvlm-Baseline} (ours) & \textbf{51.3} & \textbf{41.6} & 30.2 & 55.7 & \textbf{58.0} & \textbf{47.4} \\
\midrule
\rowcolor{gray!8} \multicolumn{7}{l}{\texttt{large} scale (4B model, 100B tokens)} \\
\textsc{Nemotron-VL-2} & 35.4 & 37.6 & 25.4 & 44.1 & 50.2 & 38.5 \\
\textsc{FineVision} & 41.5 & 16.0 & \textbf{40.5} & 58.8 & 38.5 & 39.1 \\
\textsc{\dcvlm-Baseline} (ours) & \textbf{56.9} & \textbf{46.0} & 36.9 & \textbf{68.3} & \textbf{77.6} & \textbf{57.1} \\
\midrule
\rowcolor{gray!8} \multicolumn{7}{l}{\texttt{x-large} scale (8B model, 200B tokens)} \\
\textsc{FineVision} & 44.4 & 35.6 & 41.5 & 58.7 & 67.9 & 49.6 \\
\textsc{\dcvlm-Baseline} (ours) & \textbf{59.6} & \textbf{50.7} & \textbf{47.6} & \textbf{75.2} & \textbf{84.2} & \textbf{63.5} \\
\bottomrule
\end{tabular}
}
\end{table}

\begin{table}[ht]
\centering
\caption{\textbf{Per-dataset Core results: Multilingual (MTL).} Per-dataset breakdown of the \textbf{MTL} column of \cref{tab:main-results}. The final \textbf{MTL} column is the row mean over these 3 datasets and reproduces \cref{tab:main-results}. \textbf{Bold} = best per scale.}
\label{tab:per-dataset-multilingual}
\resizebox{0.8\linewidth}{!}{
\begin{tabular}{lcccc}
\toprule
\textbf{Method} & \textbf{MTVQA} & \textbf{MMMB} & \textbf{MTL-MMB} & \textbf{MTL} \\
\midrule
\rowcolor{gray!8} \multicolumn{5}{l}{\texttt{small} scale (1B model, 6.25B tokens)} \\
\textsc{LLaVA-OneVision-1.5} & 2.8 & 32.2 & 5.7 & 13.6 \\
\textsc{Nemotron-VL-2} & 1.0 & 41.6 & 5.8 & 16.1 \\
\textsc{FineVision} & \textbf{10.3} & \textbf{47.7} & \textbf{26.5} & \textbf{28.2} \\
\textsc{\dcvlm-Baseline} (ours) & 7.5 & 46.7 & 22.1 & 25.4 \\
\midrule
\rowcolor{gray!8} \multicolumn{5}{l}{\texttt{medium} scale (2B model, 25B tokens)} \\
\textsc{LLaVA-OneVision-1.5} & 13.5 & 35.3 & 15.7 & 21.5 \\
\textsc{Nemotron-VL-2} & 16.3 & 60.9 & 33.0 & 36.7 \\
\textsc{FineVision} & \textbf{18.6} & 61.2 & 42.1 & 40.6 \\
\textsc{\dcvlm-Baseline} (ours) & 16.4 & \textbf{63.8} & \textbf{52.3} & \textbf{44.2} \\
\midrule
\rowcolor{gray!8} \multicolumn{5}{l}{\texttt{large} scale (4B model, 100B tokens)} \\
\textsc{Nemotron-VL-2} & 16.4 & 49.5 & 16.6 & 27.5 \\
\textsc{FineVision} & \textbf{21.0} & 65.5 & 48.7 & 45.1 \\
\textsc{\dcvlm-Baseline} (ours) & 20.2 & \textbf{71.6} & \textbf{60.9} & \textbf{50.9} \\
\midrule
\rowcolor{gray!8} \multicolumn{5}{l}{\texttt{x-large} scale (8B model, 200B tokens)} \\
\textsc{FineVision} & \textbf{23.1} & 71.0 & 51.0 & 48.4 \\
\textsc{\dcvlm-Baseline} (ours) & 22.8 & \textbf{76.8} & \textbf{68.5} & \textbf{56.0} \\
\bottomrule
\end{tabular}
}
\end{table}

\begin{table}[ht]
\centering
\caption{\textbf{Per-dataset Core results: Text-Only Understanding (Text).} Per-dataset breakdown of the \textbf{Text} column of \cref{tab:main-results}. The final \textbf{Text} column is the row mean over these 8 datasets and reproduces \cref{tab:main-results}. \textbf{Bold} = best per scale.}
\label{tab:per-dataset-text}
\resizebox{\linewidth}{!}{
\begin{tabular}{lccccccccc}
\toprule
\textbf{Method} & \textbf{C-Eval} & \textbf{CMMLU} & \textbf{GAOKAO} & \textbf{MMLU} & \textbf{NQ} & \textbf{RACE} & \textbf{TriviaQA} & \textbf{WinoG} & \textbf{Text} \\
\midrule
\rowcolor{gray!8} \multicolumn{10}{l}{\texttt{small} scale (1B model, 6.25B tokens)} \\
\textsc{LLaVA-OneVision-1.5} & 1.5 & 0.0 & 6.3 & 13.3 & 2.9 & 12.9 & 16.9 & 1.0 & 6.8 \\
\textsc{Nemotron-VL-2} & 17.0 & 14.2 & 19.6 & 32.3 & 0.6 & 44.1 & 5.8 & 32.3 & 20.7 \\
\textsc{FineVision} & 21.9 & 20.2 & 17.9 & 37.4 & \textbf{9.7} & 55.6 & \textbf{23.0} & 45.3 & 28.9 \\
\textsc{\dcvlm-Baseline} (ours) & \textbf{26.5} & \textbf{49.3} & \textbf{21.6} & \textbf{46.2} & 7.0 & \textbf{58.8} & 18.3 & \textbf{50.3} & \textbf{34.8} \\
\midrule
\rowcolor{gray!8} \multicolumn{10}{l}{\texttt{medium} scale (2B model, 25B tokens)} \\
\textsc{LLaVA-OneVision-1.5} & 10.7 & 0.5 & 23.0 & 24.7 & 8.3 & 24.3 & 33.0 & 3.6 & 16.0 \\
\textsc{Nemotron-VL-2} & 22.8 & 29.9 & 25.2 & 43.5 & 0.3 & 59.2 & 5.8 & 41.8 & 28.6 \\
\textsc{FineVision} & 31.9 & 61.1 & 38.3 & 59.7 & \textbf{15.5} & 75.6 & \textbf{37.1} & 51.3 & 46.3 \\
\textsc{\dcvlm-Baseline} (ours) & \textbf{32.8} & \textbf{64.5} & \textbf{39.7} & \textbf{61.6} & 13.6 & \textbf{80.0} & 35.3 & \textbf{55.2} & \textbf{47.8} \\
\midrule
\rowcolor{gray!8} \multicolumn{10}{l}{\texttt{large} scale (4B model, 100B tokens)} \\
\textsc{Nemotron-VL-2} & 30.7 & 55.5 & 35.2 & 48.9 & 4.0 & 57.2 & 9.9 & 49.6 & 36.4 \\
\textsc{FineVision} & 33.0 & 66.8 & 37.4 & 65.1 & 18.4 & 82.0 & \textbf{46.1} & 60.8 & 51.2 \\
\textsc{\dcvlm-Baseline} (ours) & \textbf{35.5} & \textbf{69.5} & \textbf{41.4} & \textbf{66.2} & \textbf{19.5} & \textbf{86.8} & 43.8 & \textbf{67.8} & \textbf{53.8} \\
\midrule
\rowcolor{gray!8} \multicolumn{10}{l}{\texttt{x-large} scale (8B model, 200B tokens)} \\
\textsc{FineVision} & 36.0 & 72.5 & 46.3 & 70.4 & 21.5 & 85.0 & 51.4 & 62.2 & 55.7 \\
\textsc{\dcvlm-Baseline} (ours) & \textbf{38.2} & \textbf{75.4} & \textbf{51.9} & \textbf{71.5} & \textbf{26.9} & \textbf{90.7} & \textbf{55.5} & \textbf{78.9} & \textbf{61.1} \\
\bottomrule
\end{tabular}
}
\end{table}

\clearpage

\section{Limitations and Future Directions}\label{app:future-directions}
Here, we highlight some limitations and open questions of our work which may be useful for exciting future research. 
\subsection{Limitations}\label{appsub:limitations}
\begin{itemize}
    \item \textbf{Scaling to longer token budgets.} In this work, we see clear scaling trends in downstream performance across our model and token budget scales from \texttt{small} (1B, 6.25B tokens) to \texttt{x-large} (8B, 200B tokens), however we remain compute-limited to  a 200B token pretraining budget. Competitive open-weight alternatives such as Qwen2-VL and Qwen3-VL train on 1.4T and 2.2T tokens during pretraining respectively.  
    \item \textbf{Scaling to larger model sizes.} In addition to scaling token budgets during the pretraining stage, stress-testing our our data pool and data mixture conclusions at model sizes $\geq$20B is an important research question not empirically addressed in this work.
    \item \textbf{Testing DCVLM data on native multimodal pretraining.} The pretraining strategy adopted in this work starts with already pretrained vision encoders and language models. However, recent works have advocated for native pretraining paradigms to enable training VLMs from scratch, which is not something we have implemented in this work. Effectively the dominance of instruction-heavy data mixtures may not exist when the inductive biases of separately pretrained encoders are absent, and is an important research question worth addressing.
    \item \textbf{More principled scaling ladder. } Our four-scale ladder is not tied to the joint optimization of model size and data token budgets, currently set up as practical defaults instead of theoretical optimal estimates. Designing a more principled scaling ladder, especially for vision-language models that use pretrained components, is a research problem of its own and we consider this out-of-scope in this work. 
    \item \textbf{Capability Coverage.} While \dcvlm{} spans a broad range of capabilities in both the extended and core set of evaluations, certain tasks like \emph{grounding and detection} are absent. While we observe issues with these benchmarks and tasks and highlighted them in \cref{app:grounding-detection-issues}, incorporating these tasks in a faithful evaluation setup is a natural and critical followup.
    \item \textbf{Poor OCR performance.} We conclude that our OCR performance lags behind \fv{} due to their better coverage and priority given to high-resolution scanned documents and pdf-derived sources. Whether simply scaling up OCR data or curating extremely high-quality and information-dense samples can bridge this gap is not adequately resolved in this work.
\end{itemize}

\subsection{Future Work}

\begin{itemize}
    \item \textbf{Towards instructional multimodal documents.} Our results suggest that instruction-heavy mixtures scale more favorably with compute, raising an important question: how can we better exploit and transform the vast supply of raw multimodal documents? 
    Such transformations may preserve the breadth and diversity of naturally occurring multimodal data while providing a denser and more targeted learning signal for instruction-tuned multimodal models.
    \item \textbf{Fine-grained mixture optimization.} Our mixing experiments only involved dense line-search sweeps over the proportion of image-caption pairs and instruction-tuning data in the mixture. A direction worth exploring is to use swarm-based methods~\citep{liu2024regmix,xie2023doremi,chen2026olmix} using proxy models and verify if they discover similar findings of instruction-heavy mixes being more scalable.
    \item \textbf{Token-level mixing.} Our mixing experiments operationalized data mixing at the \textit{sample level}, i.e. we fix the data type ratios based on the number of samples in each dataset. However, a good direction to explore would be to operationalize mixing at the token-level. As seen from~\cref{app:pool-perdataset}, the distribution of token-counts and sample-counts per-dataset can vary, and hence it would be an interesting avenue to compare and contrast different mixing strategies across.  
    \item \textbf{Quality-aware upsampling.} In our work, we only used all the quality-scores for data filtering, i.e. to discard data samples below a certain threshold. This is exclusionary by nature, and might induce harmful biases into models~\citep{hong2024s,dodge2021documenting}. Another potential use for such quality-scores which does not involved discarding data is using those scores for upsampling data samples rather than filtering. 
\end{itemize}
\clearpage

\newpage
\section*{NeurIPS Paper Checklist}

\begin{enumerate}

\item {\bf Claims}
    \item[] Question: Do the main claims made in the abstract and introduction accurately reflect the paper's contributions and scope?
    \item[] Answer: \answerYes{} %
    \item[] Justification: This work proposes a benchmark for controlled data-centric experiments to improve VLM training. Additionally, we show that data mixing is the key to assembling high-quality training data. Finally, the baseline dataset (DCVLM-Baseline) obtained from our results enables training models to achieve an accuracy improvement of +5.4pp. 
    The benchmark is described in \cref{sec:methodology}, results on data mixing in \cref{sec:mixing}, and improvements over existing pretraining datasets in \cref{sec:results}.
    \item[] Guidelines:
    \begin{itemize}
        \item The answer \answerNA{} means that the abstract and introduction do not include the claims made in the paper.
        \item The abstract and/or introduction should clearly state the claims made, including the contributions made in the paper and important assumptions and limitations. A \answerNo{} or \answerNA{} answer to this question will not be perceived well by the reviewers. 
        \item The claims made should match theoretical and experimental results, and reflect how much the results can be expected to generalize to other settings. 
        \item It is fine to include aspirational goals as motivation as long as it is clear that these goals are not attained by the paper. 
    \end{itemize}

\item {\bf Limitations}
    \item[] Question: Does the paper discuss the limitations of the work performed by the authors?
    \item[] Answer: \answerYes{} %
    \item[] Justification: The limitations are discussed in the ``Limitations and Future Directions'' section (\cref{appsub:limitations}).
    \item[] Guidelines:
    \begin{itemize}
        \item The answer \answerNA{} means that the paper has no limitation while the answer \answerNo{} means that the paper has limitations, but those are not discussed in the paper. 
        \item The authors are encouraged to create a separate ``Limitations'' section in their paper.
        \item The paper should point out any strong assumptions and how robust the results are to violations of these assumptions (e.g., independence assumptions, noiseless settings, model well-specification, asymptotic approximations only holding locally). The authors should reflect on how these assumptions might be violated in practice and what the implications would be.
        \item The authors should reflect on the scope of the claims made, e.g., if the approach was only tested on a few datasets or with a few runs. In general, empirical results often depend on implicit assumptions, which should be articulated.
        \item The authors should reflect on the factors that influence the performance of the approach. For example, a facial recognition algorithm may perform poorly when image resolution is low or images are taken in low lighting. Or a speech-to-text system might not be used reliably to provide closed captions for online lectures because it fails to handle technical jargon.
        \item The authors should discuss the computational efficiency of the proposed algorithms and how they scale with dataset size.
        \item If applicable, the authors should discuss possible limitations of their approach to address problems of privacy and fairness.
        \item While the authors might fear that complete honesty about limitations might be used by reviewers as grounds for rejection, a worse outcome might be that reviewers discover limitations that aren't acknowledged in the paper. The authors should use their best judgment and recognize that individual actions in favor of transparency play an important role in developing norms that preserve the integrity of the community. Reviewers will be specifically instructed to not penalize honesty concerning limitations.
    \end{itemize}

\item {\bf Theory assumptions and proofs}
    \item[] Question: For each theoretical result, does the paper provide the full set of assumptions and a complete (and correct) proof?
    \item[] Answer: \answerNA{} %
    \item[] Justification: The paper does not include theoretical results and proofs.
    \item[] Guidelines:
    \begin{itemize}
        \item The answer \answerNA{} means that the paper does not include theoretical results. 
        \item All the theorems, formulas, and proofs in the paper should be numbered and cross-referenced.
        \item All assumptions should be clearly stated or referenced in the statement of any theorems.
        \item The proofs can either appear in the main paper or the supplemental material, but if they appear in the supplemental material, the authors are encouraged to provide a short proof sketch to provide intuition. 
        \item Inversely, any informal proof provided in the core of the paper should be complemented by formal proofs provided in appendix or supplemental material.
        \item Theorems and Lemmas that the proof relies upon should be properly referenced. 
    \end{itemize}

    \item {\bf Experimental result reproducibility}
    \item[] Question: Does the paper fully disclose all the information needed to reproduce the main experimental results of the paper to the extent that it affects the main claims and/or conclusions of the paper (regardless of whether the code and data are provided or not)?
    \item[] Answer: \answerYes{} %
    \item[] Justification: While general details regarding data collection, curation, and training are provided in the main paper, the Appendix fully complements these, reporting trained model architecture details (\cref{app:architecture}), the training hyperparameters used (\cref{app:hparams}), the annotation infrastructure (\cref{app:annotation}), train-test decontamination (\cref{app:decontamination}), DCVLM data pool details (\cref{app:dataset_pool}), our evaluation suite (\cref{app:eval_details}), and safety preprocessing details (\cref{app:safety}).
    \item[] Guidelines:
    \begin{itemize}
        \item The answer \answerNA{} means that the paper does not include experiments.
        \item If the paper includes experiments, a \answerNo{} answer to this question will not be perceived well by the reviewers: Making the paper reproducible is important, regardless of whether the code and data are provided or not.
        \item If the contribution is a dataset and\slash or model, the authors should describe the steps taken to make their results reproducible or verifiable. 
        \item Depending on the contribution, reproducibility can be accomplished in various ways. For example, if the contribution is a novel architecture, describing the architecture fully might suffice, or if the contribution is a specific model and empirical evaluation, it may be necessary to either make it possible for others to replicate the model with the same dataset, or provide access to the model. In general. releasing code and data is often one good way to accomplish this, but reproducibility can also be provided via detailed instructions for how to replicate the results, access to a hosted model (e.g., in the case of a large language model), releasing of a model checkpoint, or other means that are appropriate to the research performed.
        \item While NeurIPS does not require releasing code, the conference does require all submissions to provide some reasonable avenue for reproducibility, which may depend on the nature of the contribution. For example
        \begin{enumerate}
            \item If the contribution is primarily a new algorithm, the paper should make it clear how to reproduce that algorithm.
            \item If the contribution is primarily a new model architecture, the paper should describe the architecture clearly and fully.
            \item If the contribution is a new model (e.g., a large language model), then there should either be a way to access this model for reproducing the results or a way to reproduce the model (e.g., with an open-source dataset or instructions for how to construct the dataset).
            \item We recognize that reproducibility may be tricky in some cases, in which case authors are welcome to describe the particular way they provide for reproducibility. In the case of closed-source models, it may be that access to the model is limited in some way (e.g., to registered users), but it should be possible for other researchers to have some path to reproducing or verifying the results.
        \end{enumerate}
    \end{itemize}

\item {\bf Open access to data and code}
    \item[] Question: Does the paper provide open access to the data and code, with sufficient instructions to faithfully reproduce the main experimental results, as described in supplemental material?
    \item[] Answer: \answerYes{} %
    \item[] Justification: The submission comes with a hyperlink to where the data is stored, while the Appendix fully reports the necessary details to reproduce our results.
    \item[] Guidelines:
    \begin{itemize}
        \item The answer \answerNA{} means that paper does not include experiments requiring code.
        \item Please see the NeurIPS code and data submission guidelines (\url{https://neurips.cc/public/guides/CodeSubmissionPolicy}) for more details.
        \item While we encourage the release of code and data, we understand that this might not be possible, so \answerNo{} is an acceptable answer. Papers cannot be rejected simply for not including code, unless this is central to the contribution (e.g., for a new open-source benchmark).
        \item The instructions should contain the exact command and environment needed to run to reproduce the results. See the NeurIPS code and data submission guidelines (\url{https://neurips.cc/public/guides/CodeSubmissionPolicy}) for more details.
        \item The authors should provide instructions on data access and preparation, including how to access the raw data, preprocessed data, intermediate data, and generated data, etc.
        \item The authors should provide scripts to reproduce all experimental results for the new proposed method and baselines. If only a subset of experiments are reproducible, they should state which ones are omitted from the script and why.
        \item At submission time, to preserve anonymity, the authors should release anonymized versions (if applicable).
        \item Providing as much information as possible in supplemental material (appended to the paper) is recommended, but including URLs to data and code is permitted.
    \end{itemize}

\item {\bf Experimental setting/details}
    \item[] Question: Does the paper specify all the training and test details (e.g., data splits, hyperparameters, how they were chosen, type of optimizer) necessary to understand the results?
    \item[] Answer: \answerYes{} %
    \item[] Justification: The main experimental procedure is described in \cref{sec:methodology,sec:experiments,sec:results}, and more details are provided in the Appendix (\cref{app:architecture,app:hparams,app:decontamination,app:dataset_pool,app:eval_details}).
    \item[] Guidelines:
    \begin{itemize}
        \item The answer \answerNA{} means that the paper does not include experiments.
        \item The experimental setting should be presented in the core of the paper to a level of detail that is necessary to appreciate the results and make sense of them.
        \item The full details can be provided either with the code, in appendix, or as supplemental material.
    \end{itemize}

\item {\bf Experiment statistical significance}
    \item[] Question: Does the paper report error bars suitably and correctly defined or other appropriate information about the statistical significance of the experiments?
    \item[] Answer: \answerNo{} %
    \item[] Justification: Error bars are not reported as they are too computationally expensive to compute.
    \item[] Guidelines:
    \begin{itemize}
        \item The answer \answerNA{} means that the paper does not include experiments.
        \item The authors should answer \answerYes{} if the results are accompanied by error bars, confidence intervals, or statistical significance tests, at least for the experiments that support the main claims of the paper.
        \item The factors of variability that the error bars are capturing should be clearly stated (for example, train/test split, initialization, random drawing of some parameter, or overall run with given experimental conditions).
        \item The method for calculating the error bars should be explained (closed form formula, call to a library function, bootstrap, etc.)
        \item The assumptions made should be given (e.g., Normally distributed errors).
        \item It should be clear whether the error bar is the standard deviation or the standard error of the mean.
        \item It is OK to report 1-sigma error bars, but one should state it. The authors should preferably report a 2-sigma error bar than state that they have a 96\% CI, if the hypothesis of Normality of errors is not verified.
        \item For asymmetric distributions, the authors should be careful not to show in tables or figures symmetric error bars that would yield results that are out of range (e.g., negative error rates).
        \item If error bars are reported in tables or plots, the authors should explain in the text how they were calculated and reference the corresponding figures or tables in the text.
    \end{itemize}

\item {\bf Experiments compute resources}
    \item[] Question: For each experiment, does the paper provide sufficient information on the computer resources (type of compute workers, memory, time of execution) needed to reproduce the experiments?
    \item[] Answer: \answerYes{} %
    \item[] Justification: For each experiment ``scale'' (\texttt{small}, \texttt{medium}, \texttt{large}, \texttt{x-large}), \cref{tab:scales} reports the number of `H100 hrs' required to reproduce our results.
    \item[] Guidelines:
    \begin{itemize}
        \item The answer \answerNA{} means that the paper does not include experiments.
        \item The paper should indicate the type of compute workers CPU or GPU, internal cluster, or cloud provider, including relevant memory and storage.
        \item The paper should provide the amount of compute required for each of the individual experimental runs as well as estimate the total compute. 
        \item The paper should disclose whether the full research project required more compute than the experiments reported in the paper (e.g., preliminary or failed experiments that didn't make it into the paper). 
    \end{itemize}
    
\item {\bf Code of ethics}
    \item[] Question: Does the research conducted in the paper conform, in every respect, with the NeurIPS Code of Ethics \url{https://neurips.cc/public/EthicsGuidelines}?
    \item[] Answer: \answerYes{} %
    \item[] Justification: We carefully reviewed the NeurIPS Code of Ethics and believe our work complies with it.
    \item[] Guidelines:
    \begin{itemize}
        \item The answer \answerNA{} means that the authors have not reviewed the NeurIPS Code of Ethics.
        \item If the authors answer \answerNo, they should explain the special circumstances that require a deviation from the Code of Ethics.
        \item The authors should make sure to preserve anonymity (e.g., if there is a special consideration due to laws or regulations in their jurisdiction).
    \end{itemize}

\item {\bf Broader impacts}
    \item[] Question: Does the paper discuss both potential positive societal impacts and negative societal impacts of the work performed?
    \item[] Answer: \answerNA{} %
    \item[] Justification: Since DCVLM is a benchmark for controlled data-centric experiments with the goal of improving VLM training, our work fits as fundamental research and, therefore, it is not tied to particular applications.
    \item[] Guidelines:
    \begin{itemize}
        \item The answer \answerNA{} means that there is no societal impact of the work performed.
        \item If the authors answer \answerNA{} or \answerNo, they should explain why their work has no societal impact or why the paper does not address societal impact.
        \item Examples of negative societal impacts include potential malicious or unintended uses (e.g., disinformation, generating fake profiles, surveillance), fairness considerations (e.g., deployment of technologies that could make decisions that unfairly impact specific groups), privacy considerations, and security considerations.
        \item The conference expects that many papers will be foundational research and not tied to particular applications, let alone deployments. However, if there is a direct path to any negative applications, the authors should point it out. For example, it is legitimate to point out that an improvement in the quality of generative models could be used to generate Deepfakes for disinformation. On the other hand, it is not needed to point out that a generic algorithm for optimizing neural networks could enable people to train models that generate Deepfakes faster.
        \item The authors should consider possible harms that could arise when the technology is being used as intended and functioning correctly, harms that could arise when the technology is being used as intended but gives incorrect results, and harms following from (intentional or unintentional) misuse of the technology.
        \item If there are negative societal impacts, the authors could also discuss possible mitigation strategies (e.g., gated release of models, providing defenses in addition to attacks, mechanisms for monitoring misuse, mechanisms to monitor how a system learns from feedback over time, improving the efficiency and accessibility of ML).
    \end{itemize}
    
\item {\bf Safeguards}
    \item[] Question: Does the paper describe safeguards that have been put in place for responsible release of data or models that have a high risk for misuse (e.g., pre-trained language models, image generators, or scraped datasets)?
    \item[] Answer: \answerYes{} %
    \item[] Justification: \Cref{app:safety} reports the safety processing details we used to reduce as much as possible harmful, toxic, and biased content.
    \item[] Guidelines:
    \begin{itemize}
        \item The answer \answerNA{} means that the paper poses no such risks.
        \item Released models that have a high risk for misuse or dual-use should be released with necessary safeguards to allow for controlled use of the model, for example by requiring that users adhere to usage guidelines or restrictions to access the model or implementing safety filters. 
        \item Datasets that have been scraped from the Internet could pose safety risks. The authors should describe how they avoided releasing unsafe images.
        \item We recognize that providing effective safeguards is challenging, and many papers do not require this, but we encourage authors to take this into account and make a best faith effort.
    \end{itemize}

\item {\bf Licenses for existing assets}
    \item[] Question: Are the creators or original owners of assets (e.g., code, data, models), used in the paper, properly credited and are the license and terms of use explicitly mentioned and properly respected?
    \item[] Answer: \answerYes{} %
    \item[] Justification: \Cref{appsub:licensing,tab:dcvlm-datamix-licenses} report links pointing to the datasets we used in the DCVLM pool, with all available licenses associated. Furthermore, we properly credited authors by citing their work.
    \item[] Guidelines:
    \begin{itemize}
        \item The answer \answerNA{} means that the paper does not use existing assets.
        \item The authors should cite the original paper that produced the code package or dataset.
        \item The authors should state which version of the asset is used and, if possible, include a URL.
        \item The name of the license (e.g., CC-BY 4.0) should be included for each asset.
        \item For scraped data from a particular source (e.g., website), the copyright and terms of service of that source should be provided.
        \item If assets are released, the license, copyright information, and terms of use in the package should be provided. For popular datasets, \url{paperswithcode.com/datasets} has curated licenses for some datasets. Their licensing guide can help determine the license of a dataset.
        \item For existing datasets that are re-packaged, both the original license and the license of the derived asset (if it has changed) should be provided.
        \item If this information is not available online, the authors are encouraged to reach out to the asset's creators.
    \end{itemize}

\item {\bf New assets}
    \item[] Question: Are new assets introduced in the paper well documented and is the documentation provided alongside the assets?
    \item[] Answer: \answerYes{} %
    \item[] Justification: The link to the data attached to this submission contains a README file explaining how the data is organized and how to navigate it.
    \item[] Guidelines:
    \begin{itemize}
        \item The answer \answerNA{} means that the paper does not release new assets.
        \item Researchers should communicate the details of the dataset\slash code\slash model as part of their submissions via structured templates. This includes details about training, license, limitations, etc. 
        \item The paper should discuss whether and how consent was obtained from people whose asset is used.
        \item At submission time, remember to anonymize your assets (if applicable). You can either create an anonymized URL or include an anonymized zip file.
    \end{itemize}

\item {\bf Crowdsourcing and research with human subjects}
    \item[] Question: For crowdsourcing experiments and research with human subjects, does the paper include the full text of instructions given to participants and screenshots, if applicable, as well as details about compensation (if any)? 
    \item[] Answer: \answerNA{} %
    \item[] Justification: Our work does not involve crowdsourcing nor research with human subjects.
    \item[] Guidelines:
    \begin{itemize}
        \item The answer \answerNA{} means that the paper does not involve crowdsourcing nor research with human subjects.
        \item Including this information in the supplemental material is fine, but if the main contribution of the paper involves human subjects, then as much detail as possible should be included in the main paper. 
        \item According to the NeurIPS Code of Ethics, workers involved in data collection, curation, or other labor should be paid at least the minimum wage in the country of the data collector. 
    \end{itemize}

\item {\bf Institutional review board (IRB) approvals or equivalent for research with human subjects}
    \item[] Question: Does the paper describe potential risks incurred by study participants, whether such risks were disclosed to the subjects, and whether Institutional Review Board (IRB) approvals (or an equivalent approval/review based on the requirements of your country or institution) were obtained?
    \item[] Answer: \answerNA{} %
    \item[] Justification: Our work does not involve crowdsourcing nor research with human subjects.
    \item[] Guidelines:
    \begin{itemize}
        \item The answer \answerNA{} means that the paper does not involve crowdsourcing nor research with human subjects.
        \item Depending on the country in which research is conducted, IRB approval (or equivalent) may be required for any human subjects research. If you obtained IRB approval, you should clearly state this in the paper. 
        \item We recognize that the procedures for this may vary significantly between institutions and locations, and we expect authors to adhere to the NeurIPS Code of Ethics and the guidelines for their institution. 
        \item For initial submissions, do not include any information that would break anonymity (if applicable), such as the institution conducting the review.
    \end{itemize}

\item {\bf Declaration of LLM usage}
    \item[] Question: Does the paper describe the usage of LLMs if it is an important, original, or non-standard component of the core methods in this research? Note that if the LLM is used only for writing, editing, or formatting purposes and does \emph{not} impact the core methodology, scientific rigor, or originality of the research, declaration is not required.
    \item[] Answer: \answerYes{} %
    \item[] Justification: The paper is about training large vision-language models (rather than using them as components). We describe the architecture and the pretrained LLM we use in Section \ref{modeltrainingrecipe} and Appendix \ref{app:architecture}. 
    \item[] Guidelines:
    \begin{itemize}
        \item The answer \answerNA{} means that the core method development in this research does not involve LLMs as any important, original, or non-standard components.
        \item Please refer to our LLM policy in the NeurIPS handbook for what should or should not be described.
    \end{itemize}

\end{enumerate}

\end{document}